# Navigating Uncertainties in Machine Learning for Structural Dynamics: A Comprehensive Survey of Probabilistic and Non-Probabilistic Approaches in Forward and Inverse Problems

**Wang-Ji Yan[1,2], Lin-Feng Mei[1*], Jiang Mo[1], Costas Papadimitriou[3], Ka-Veng Yuen[1,2], Michael Beer[4,5,6]**

*[1]State Key Laboratory of Internet of Things for Smart City and Department of Civil and Environmental Engineering, University of Macau, China*
*[2]Guangdong-Hong Kong-Macau Joint Laboratory for Smart Cities, China*
*[3]Department of Mechanical Engineering, University of Thessaly, Volos, Greece*
*[4]Leibniz Universität Hannover, Institute for Risk and Reliability, Hannover, Germany*
*[5]Department of Civil and Environmental Engineering, University of Liverpool, Liverpool L69 3BX, United Kingdom*
*[6]International Joint Research Center for Resilient Infrastructure & International Joint Research Center for Engineering Reliability and Stochastic Mechanics, Tongji University, Shanghai 200092, PR China*

**Abstract:** In the era of big data, machine learning (ML) has become a powerful tool in various scientific and engineering fields, notably impacting structural dynamics. ML algorithms offer advantages by modeling physical phenomena based on data, even in the absence of underlying mechanisms. However, uncertainties such as measurement noise and modeling errors can compromise the reliability of ML predictions, highlighting the need for effective uncertainty awareness to enhance prediction robustness. This paper presents a comprehensive literature survey on navigating uncertainties in ML, categorizing uncertainty-aware approaches into probabilistic methods (including Bayesian and frequentist perspectives) and non-probabilistic methods (such as interval learning and fuzzy learning). Bayesian neural networks, known for their uncertainty quantification and nonlinear mapping capabilities, are emphasized for their superior performance and potential. The survey covers various techniques and methodologies for addressing uncertainties in ML, discussing fundamentals and implementation procedures of each method. While providing a concise overview of fundamental concepts, the paper refrains from in-depth critical explanations. Strengths and limitations of each approach are examined, along with their applications in uncertainty-aware structural dynamic forward problems like response prediction, sensitivity assessment, and reliability analysis, and inverse problems like system identification, model updating, and damage identification. Additionally, the survey identifies research gaps and suggests future directions for investigations, aiming to provide comprehensive insights to the research community. By offering an extensive overview of both probabilistic and non-probabilistic approaches, this survey aims to assist researchers and practitioners in making informed decisions when utilizing ML techniques to address uncertainties in structural dynamic problems.

**Key words:** Structural dynamics; Machine learning; Uncertainty-aware learning; Uncertainty quantification and propagation; Bayesian deep learning.

*Corresponding author.
E-mail address: wangjiyan@um.edu.mo (W.J. Yan); yc17409@um.edu.mo (L.F. Mei)

**List of Abbreviations**

| | |
|---|---|
| AD | anomaly detection |
| AE | autoencoder |
| BAL-GP | Bayesian active learning-augmented Gaussian process |
| BBL | Bayesian broad learning |
| BDL | Bayesian deep learning |
| BN | Bayesian network |
| BNN | Bayesian neural network |
| CDF | cumulative density function |
| CNN | convolutional neural network |
| DGP | deep Gaussian process |
| DINN | deep interval neural network |
| DPMM | Dirichlet process mixture model |
| DNN | deep neural network |
| DTL | deep transfer learning |
| ELBO | evidence lower bound |
| EVT | extreme value theory |
| FERCM | fuzzy equivalence relation-based clustering method |
| FNN | feedforward neural network |
| GMM | Gaussian mixture model |
| GP | Gaussian process |
| GRU | Gated recurrent unit |
| HMM | hidden Markov model |
| HNN | Hamiltonian neural network |
| KL | Kullback-Leibler |
| LUBE | lower upper bound estimation |
| LSTM | long-short term memory |
| MAP | maximum a posterior |
| MCMC | Markov chain Monte Carlo |
| ML | machine learning |
| MLE | maximum likelihood estimation |
| MLP | multilayer perceptron |
| NN | neural network |
| OOD | out-of-distribution |
| PCA | principal component analysis |
| PD | probabilistic distance |
| PDF | probability density function |
| PIML | physics-informed machine learning |
| PMF | probability mass function |
| PML | probabilistic machine learning |
| RF | random forest |
| RNN | recurrent neural network |
| RVM | relevance vector machine |
| SVM | support vector machines |
| TL | transfer learning |
| UQ | uncertainty quantification |
| UT | uncertainty treatment |
| VAE | variational autoencoder |
| VI | variational inference |

## 1 Introduction

Structural dynamics holds considerable significance in the fields of civil, mechanical, and aerospace engineering, for its role in the management and operation of engineering structures. Typically, the basic elements of structural dynamics include the input, the dynamical system, and the output (response) [1], while structural dynamic problems can be broadly categorized as forward problems and inverse problems according to provided information and objectives [2], as illustrated in Fig. 1. Forward problems focus on simulating the response of a given system, primarily involving response prediction, sensitivity analysis, and reliability analysis. In contrast, inverse problems aim to deduce underlying information about the system based on input-output or output-only measurements [2], encompassing tasks such as structural system identification, model updating, and structural damage identification.

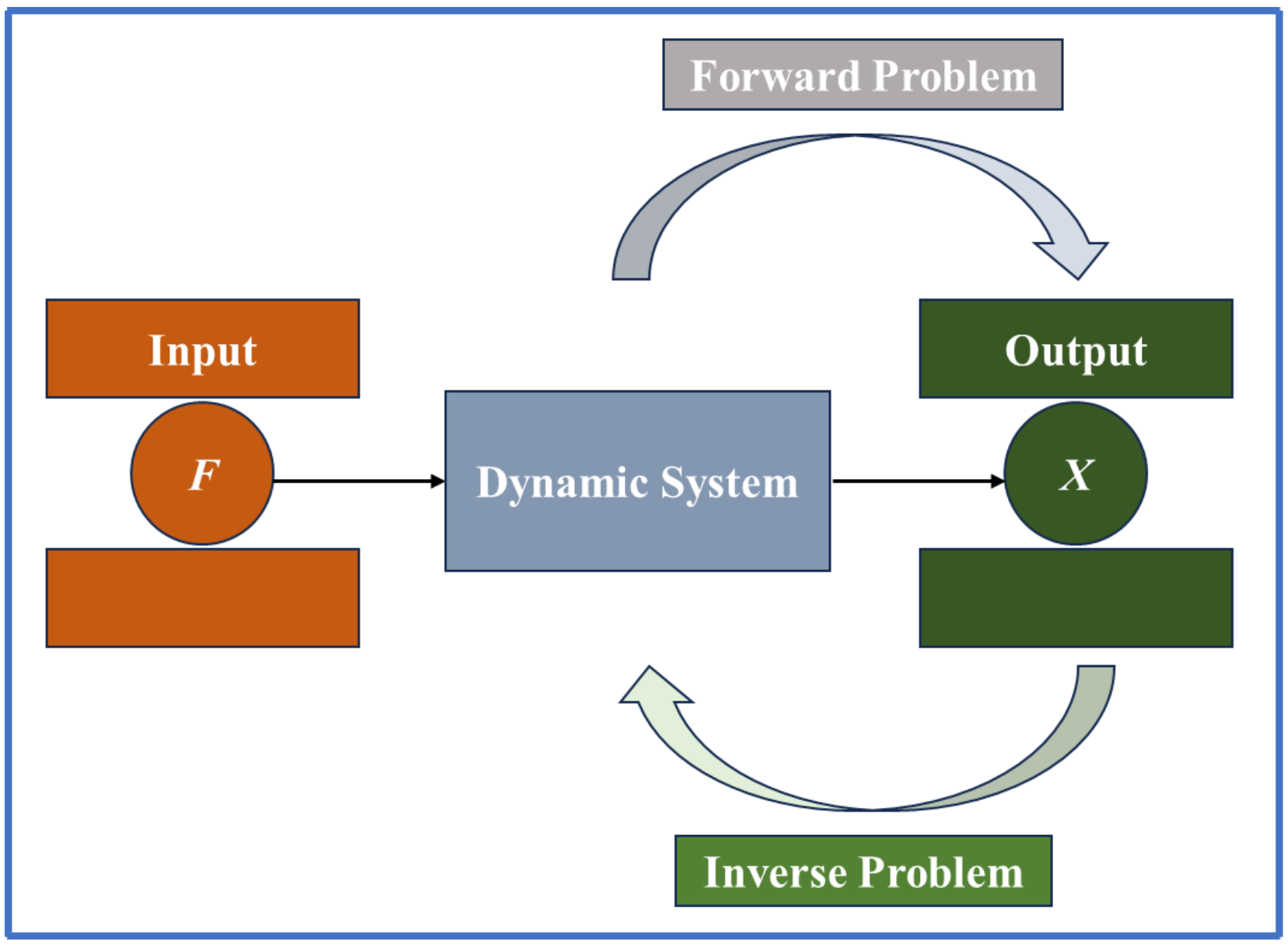

Figure 1. Forward and inverse problems in structural dynamics.

In terms of the approaches to structural dynamic analysis, they can be classified as model-driven approaches and data-driven approaches [3]. The former rely on a physics-based or law-based model (typically a finite element model) of the investigated structure for dynamic analysis, but difficulties could arise from ensuring the accuracy of the physics-based model as well as the efficiency of simulation on this model, especially for large-scale civil infrastructures such as bridges. Alternatively, data-driven approaches directly construct statistical models of the measured data, and then use pattern recognition methods for dynamic analysis. In the past few decades, the rapid development of sensor systems, computing resources, and algorithmic improvements, has led to an explosive growth in both the quantity and quality of available data, fostering the prevalence of data-driven approaches in structural dynamics. Meanwhile, machine learning (ML) has emerged as a pivotal analytical technique within these methodologies. Over the years, a huge number of ML techniques, including supervised, unsupervised, semi-supervised, and reinforcement learning, have been developed and extensively applied in academia and industry to analyze voluminous and intricate datasets to uncover hidden patterns and reach incisive insights, which has infiltrated almost every field of science and become a crucial part of various real-world applications [4, 5]. In the realm of structural dynamics, the adoption of ML-enabled data-driven approaches has also witnessed a surge and has been proved successful for various tasks, which have been comprehensively reviewed in the literature [3, 4, 6-15].

Despite these achievements of ML in structural dynamic analysis, it is also well-recognized that ML-based methods sometimes make unexpected, incorrect, but overconfident predictions, especially in a complex real-world environment [16-18]. On the other hand, structural dynamics is a high-stakes field, in which false predictions could lead to serious consequences. As a result, it is non-trivial for the ML-based methods to be aware of the level of uncertainty in their predictions to avoid overconfident results. Generally, ML involves the procedures of collecting data, pre-processing data, choosing a model to learn from the data, selecting a learning algorithm to train the selected model and inferring results from the learned model. Each of these steps inherently involves uncertainties stemming from two reasons, namely, data uncertainty (aleatoric uncertainty) and model uncertainty (epistemic uncertainty) [19, 20]. Aleatoric uncertainty captures noise inherent in the data, which could result from the nature of random vibration, measurement error, material variations, environmental, operational, and manufacturing variabilities, etc. in the context of structural dynamics, and cannot be reduced given more training data. Epistemic uncertainty, on the other hand, captures the lack of knowledge about which model generated the collected data, which can be explained away given enough data. However, despite the clear conceptual distinction between these two types of uncertainties, explicitly separating them in the predictions of ML models remains challenging. This difficulty arises because the definitions of aleatoric and epistemic uncertainties are often relative and can be influenced by the modeling process. In different application scenarios, the type of uncertainties may even be interchangeable [20]. Therefore, although some approaches have been proposed to separately identify and quantify aleatoric and epistemic uncertainties [19, 21], a more intuitive and commonly used strategy is to estimate the overall uncertainty in ML models' predictions [22, 23], which comprises both aleatoric and epistemic components and is commonly referred to as predictive uncertainty. It is worth noting that here and throughout this work, the uncertainty, unless otherwise specified, refers to this predictive uncertainty.

Given the ubiquitous presence of uncertainty, the involvement of highly inhomogeneous data sources, and the scarcity of labeled data in the field of structural dynamics, it is essential for ML-based methods to include uncertainties and provide uncertainty estimates to uncover beneficial information for a better decision-making process. To this end, the development and application of novel uncertainty treatment (UT) methods in tandem with different ML-enhanced techniques are crucial to yield useful information and amplify the interpretability and reliability of the results. Generally, the problems in UT for ML can be classified into two categories as illustrated in Fig. 2:

- **The forward propagation of uncertainty:** This type of UT investigates the uncertainty propagation from the inputs and parameters of the dynamic system throughout the ML model to systematically assess the effect of the uncertainty on model prediction, which requires representing the uncertainties via prior knowledge, assumptions, and experimental estimation for uncertainty analysis.
- **The inverse assessment of uncertainty:** This type of UT involves estimating uncertainties in the investigated parameters or characteristics of the dynamic system based on the measured responses, which is a "soft" decision making process [24] that essentially reverses the typical flow of forward uncertainty analysis.

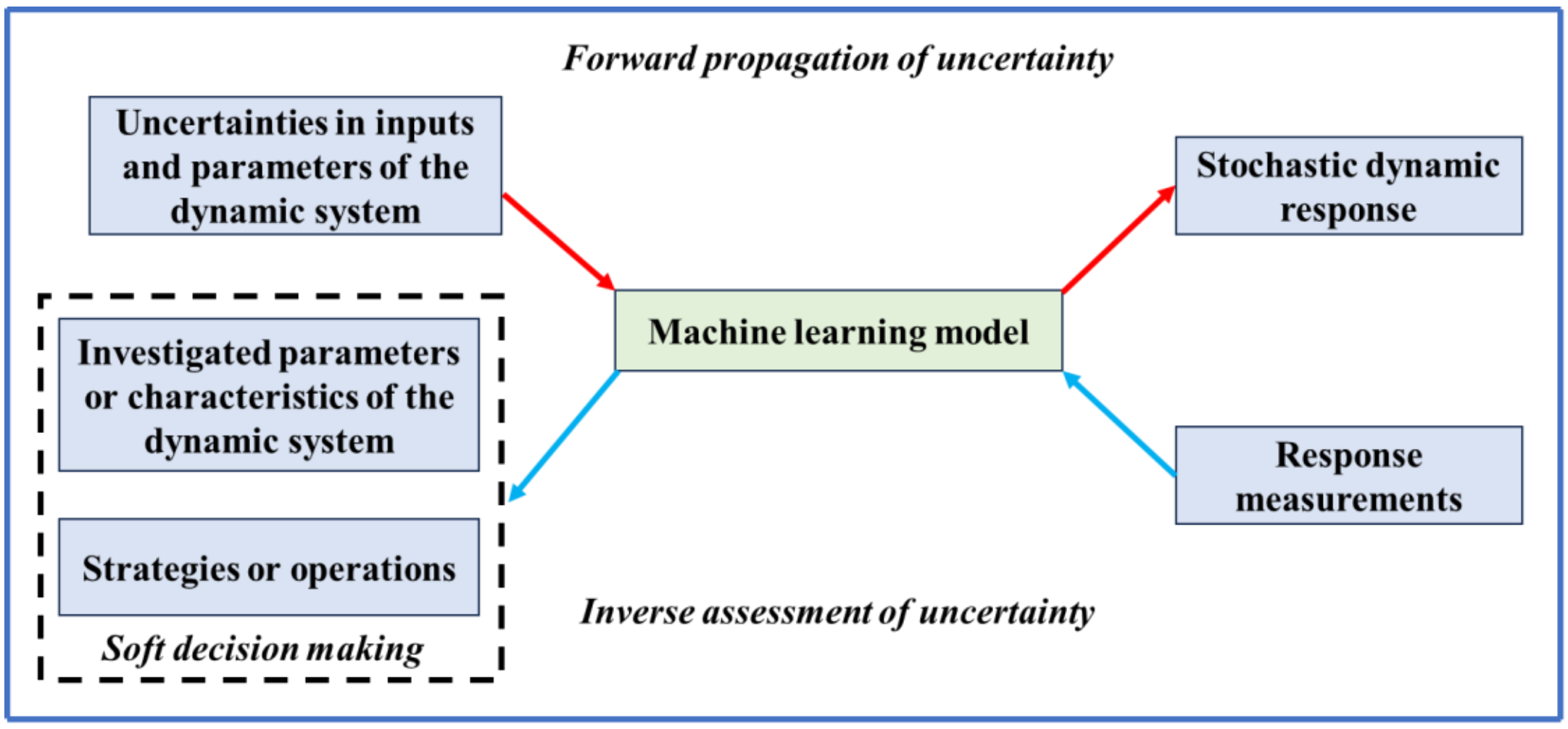

Figure 2. Schematic diagram of forward and inverse assessment of uncertainty in ML-based

methods for structural dynamics (reproduced from [24]).

For the UT approaches, the current mainstream falls to a probabilistic perspective, where probability distributions are used to represent the uncertainty of parameters or predictions of ML models [17, 23, 25]. This type of UT techniques can be further divided into Non-Bayesian and Bayesian methods, corresponding to the frequentist and Bayesian probabilities, respectively. In addition to probabilistic methods, non-probabilistic approaches such as prediction intervals and fuzzy logic also play crucial roles in UT for ML, which capture uncertainty through various criteria including the interval, imprecision, or vagueness of predictions of ML models. One obvious advantage of UT for ML models is its ability to assess the trustworthiness of model predictions and indicate when extra caution is required for decision making based on these predictions [26]. For instance, in a binary classification problem, an ML model with uncertainty estimate might predict class 1 with probability of 0.55 and class 2 with probability of 0.45 conditioned on a certain input. This result reflects a low confidence level in the model prediction, signaling that more data should be collected before relying on this model. In contrast, without UT, this model only predicts class 1 for the same input, exhibiting overconfidence and probably leading to incorrect decisions. As a result, the integration of UT in ML has a viable potential to address a central research concern faced by the ML community, namely the robustness of ML models [26]. In fact, the absence of essential performance characteristics to estimate the reliability of ML models has emerged as one of the fundamental obstacles that limits the application scope of ML in high-stake, high-reward decision environments, such as healthcare, power grid system, and structural dynamics, which underscores the significant value of UT for ML models.

With these challenges and research gaps in mind, more recently, there has been a noticeable surge in research focusing on UT for ML-enabled methods within the field of structural dynamics. On the contrary, whilst substantial works have been proposed to systematically review the applications of ML in structural dynamics, to the authors' best knowledge, few of these works have examined ML-based approaches from a perspective of UT. Moreover, the majority of current research on uncertainty in ML concentrates on probabilistic UT techniques for deep neural networks (DNNs) [22, 23, 27, 28], while UT for traditional ML models, as well as non-probabilistic UT methods, receive significantly less attention despite their importance. To fill these gaps, a comprehensive literature survey is provided in this work, focusing on typical UT techniques for ML-based methods, as well as their application in the field of structural dynamics. Specifically, this survey categorizes the approaches for UT in ML into probabilistic methods, including both Non-Bayesian and Bayesian approaches, and non-probabilistic methods, including interval learning and fuzzy learning. Bayesian neural networks (BNNs), known for their proficiency in handling uncertainty quantification (UQ) and nonlinear mapping [28], are emphasized because of their superior performance. Based on this categorization, this survey encompasses a diverse array of techniques and methodologies developed to tackle uncertainties. While not delving into an exhaustive and critical exploration of their details, the advantages and drawbacks of each approach are discussed in detail concerning their applications in forward and inverse problems of structural dynamics. Furthermore, this work outlines some research gaps and potential future directions to facilitate a better understanding of the applications of UT for ML models within the context of structural dynamics.

The remaining parts of this work are organized as follows: Section 2 outlines the fundamentals of uncertainty in ML within the context of structural dynamics, while also defining the scope of this work. Section 3 and Section 4 introduce prevalent probabilistic ML methods considering uncertainties from the frequentist and Bayesian perspective, respectively. Section 5 examines uncertainty-aware ML methods within a non-probabilistic framework. Some typical applications of these UT techniques for ML-based approaches on the forward and inverse problems of structural dynamics are summarized in Section 6, while Section 7 highlights the current research gaps in this field and suggests some promising future directions. Finally, Section 8 presents some concise conclusions drawn from this survey to facilitate understanding.

## 2 Overview of Uncertainty in Machine Learning

In this section, we provide a fundamental overview of ML, as well as some background of two major types of uncertainty, namely aleatoric uncertainty, and epistemic uncertainty. Specifically, we summarize the categorization of ML methods and analyze the potential sources of each type of uncertainty in ML within the context of structural dynamics. Additionally, we discuss how uncertainty is represented from both probabilistic and non-probabilistic perspectives.

### *2.1 Fundamentals of ML*

Although ML is an extensive concept that covers a wide range of models and can be categorized via a variety of criteria from different perspectives (e.g. parametric model versus nonparametric model, discriminative model versus generative model, shallow learning versus deep learning, etc.), the most fundamental and commonly applied classification for ML is from the perspective of the learning approach, which divides ML methods into four categories, namely supervised learning, unsupervised learning, semi-supervised learning, and reinforcement learning (It is worth noting that the application of reinforcement learning in structural dynamics concerning UT is relatively limited [14], thus it is not covered in this work.):

- **Supervised learning**: Supervised learning is the ML task to learn a function that maps the input space to the output space using a set of labeled training data comprising input-output pairs, which involves classification and regression tasks. In classification, the prediction of the ML model is a categorical or nominal variable denoting the class index, where the class of each testing instance is determined based on common patterns of each class captured from the training data. On the other hand, regression involves discovering a continuous function that maps the input to the output based on the training data to predict numerical values rather than class labels. Supervised learning is the most widely used ML technique due to its superior learning capacity [29], but its application in structural dynamics could be impeded by the high cost of acquiring well-annotated training data [3].
- **Unsupervised learning**: Unsupervised learning is the ML task to unveil the underlying patterns and latent structure of the dataset using only unlabeled training data. The most common unsupervised learning tasks include density estimation (finding the data distribution), clustering (partitioning the dataset into groups with maximum similarity), and dimensionality reduction (discovering a lower-dimensional space of latent variables that capture essential information in the data). Unsupervised models are generally more flexible than their supervised counterparts since labels are not required during training, but they often exhibit reduced inference capability as a trade-off, leading to their predominant use for fundamental tasks in structural dynamics, such as data preprocessing and damage detection.

- **Semi-supervised learning**: To strike the balance between the inherent advantages of supervised methods and the limitations posed by insufficient well-annotated data, there is a growing interest in the intermediate realm between supervised and unsupervised learning, referred to as semi-supervised learning. This approach seeks to utilize a small set of labeled data in conjunction with a larger pool of unlabeled data to achieve optimal model performance with minimal labeling cost, which mainly involves semi-supervised classification and semi-supervised clustering [30]. Semi-supervised classification is similar to its supervised counterpart but identifies data patterns from unlabeled training data to reduce the reliance on labeled training data, while semi-supervised clustering employs a small amount of labeled data with side information in clustering as additional constraints to help identify data patterns. These semi-supervised approaches, while not that prevalent as supervised and unsupervised methods in structural dynamics, hold potential in this field due to their distinct advantages in reducing labeled training data and improving model capacity.

As asserted by the ‘‘no free lunch’’ theorem, no single learning approach consistently outperforms others across all domains [31], underscoring the significance of acknowledging uncertainty when selecting an appropriate ML model. Furthermore, other uncertainties are ubiquitous throughout the ML literature, associated with both the observed data and the ML model itself. Consequently, accurately recognizing and representing these uncertainties is crucial for enhancing the performance and reliability of ML-based methods.

### *2.2 Sources of uncertainty in ML*

While uncertainty in ML can be resulted from a variety of different sources with complex mechanisms, it can generally be categorized as aleatoric uncertainty and epistemic uncertainty. This categorization, which originates from characterizing uncertainties in engineering modeling for risk and reliability analysis [20], can be extended to all fields within the model universe including ML [19]. The definitions and sources of these two types of uncertainty can be summarized as follows:

- **Aleatoric uncertainty:** This type of uncertainty, also known as data uncertainty [28], arises from the inherent randomness or noise of data, which is irreducible even if more data are collected to train the ML model and can be further divided into homoscedastic uncertainty and heteroscedastic uncertainty [19]. The former remains constant for different inputs, while the latter depends on the inputs as some inputs potentially having more noisy outputs than others. In ML, aleatoric uncertainty embodies the inherent stochasticity within the modeling process, which can arise from various factors, including the inputs, the outputs, or the dependency between them. In some cases, aleatoric uncertainty may also originate from the intrinsic stochastic behavior of the modeled physical system, which is often captured and represented using stochastic simulators [32]. In the context of structural dynamics, this type of uncertainty can be associated with the entire process of data acquisition and processing. Example sources of aleatoric uncertainty include variability of material properties and external loads, measurement noise of excitation and dynamic response, environmental and operational variabilities (EOVs), inappropriate data pre-processing methods, etc. Moreover, data uncertainty can also accumulate from multiple sources and propagate into the ML model.

- **Epistemic uncertainty:** This type of uncertainty, also referred to as model uncertainty [28], stems from a lack of knowledge about which ML model is best suited to describe the given data, which can be reduced by collecting additional training data to refine the ML model or acquiring more information about the investigated problem to guide model selection. The sources of epistemic uncertainty in ML are similar to those in engineering problems, which can be categorized as uncertainty from model parameters and uncertainty from model architecture [20, 33]. The former is associated with the precision in estimating parameters of ML models during the training process, which is mainly caused by a lack of training data or inherent limitations of the selected training algorithms. On the other hand, uncertainty in model architecture is attributed to the choice of ML model itself for interpreting different sets of observed data, which could result from simplification and approximation procedures involved in constructing the ML model. For example, neural networks (NNs) with different number of hidden layers and activation functions may exhibit different performance on the same dataset.

These two types of uncertainty are ubiquitous in areas that utilize ML-based methods, and some common causes of them in the context of structural dynamics are summarized in Table 1. Depending on the application scenario, the predominant type of uncertainty may vary, and thus specific methods are required to address them. For example, it is typically more effective to quantify aleatoric uncertainty in many big data regimes [19], as it is irreducible with the increase of training data. However, in cases where the test data falls outside the distribution of the training data, which is known as the out-of-distribution (OOD) samples, the ML model is prone to make predictions with high level of epistemic uncertainty since ML models are typically perform better in interpolation than in extrapolation, which underscores the significance of addressing epistemic uncertainty to correctly identify OOD samples. Conditioned on the distinct characteristics of these two types of uncertainty, many efforts have been devoted to explicitly distinguishing aleatoric and epistemic uncertainties in ML models. For example, the variance decomposition method has been extensively investigated in global sensitivity analysis [26, 34] to separate aleatoric and epistemic uncertainties. For regression tasks using Monte Carlo (MC) dropout-based BNNs, epistemic uncertainty can be estimated by marginalizing over the variational posterior distribution of the weights, while the aleatoric uncertainty can be interpreted as learned loss attenuation [19].

Nevertheless, in many real-world engineering problems, aleatoric and epistemic uncertainties coexist with complicated interactions, which usually prevents the explicit separation of these two types of uncertainties. Moreover, the distinction between aleatoric and epistemic uncertainties is influenced by modeling choices, making them interchangeable sometimes [20]. As a result, practical UT techniques for ML models mainly care about the overall uncertainty involved in model predictions, which is referred to as predictive uncertainty and results from the combination of aleatoric and epistemic uncertainties. For example, in structural damage quantification problems, an ML model with UT would generate a probability distribution of the predicted structure stiffness to represent the model's confidence in different predictions.

Table 1. Typical causes of aleatoric and epistemic uncertainty in ML-based methods for structural dynamic analysis

| Type | Source | Typical causes |
|---|---|---|
| Aleatoric uncertainty | Inherent randomness of data | Measurement noise, environmental and operational variabilities, variability in material properties, manufacturing variabilities and tolerances, etc. |
| Epistemic uncertainty | Lack of knowledge about the underlying model parameters or model architecture | Limited training data, local optimum of learning algorithm, inappropriate modeling technique, etc. |

### *2.3 Overview of the uncertainty treatment methods for ML models covered in this survey*

In this work, the UT techniques for ML models are divided into two categories, namely probabilistic and non-probabilistic methods.

#### *2.3.1 Probabilistic methods for uncertainty treatment*

Probabilistic methods employ a probability distribution of model predictions to measure predictive uncertainty. These methods provide an intuitive and comprehensive perspective for interpreting and capturing uncertainty, but they are subjective to the assumptions used to specify the distribution form and can be computationally expensive when employing complex assumptions. Conditioned on the two perspectives of probability, probabilistic UT methods can be further categorized into Non-Bayesian and Bayesian approaches. Typically, non-Bayesian probabilistic approaches emphasize data uncertainty through utilizing prespecified probability distributions to describe the model predictions, while the model parameters are typically treated as point estimates (e.g., maximum likelihood estimates). Meanwhile, some non-Bayesian methods can also address model uncertainty through specific approaches. For example, ensemble of NNs combines multiple NNs to formulate the predictive uncertainty and provide a mechanism to address model-related uncertainties [35, 36]. In contrast, Bayesian approaches treat the parameters of ML models as uncertain variables. The uncertainty in a variable is assigned using a probability distribution to quantify how plausible is each possible value of the variable. Then calculus of probability is used to establish a comprehensive framework to represent and quantify predictive uncertainty through the (posterior) predictive distribution. In this framework, probability distributions, including prior, likelihood, and posterior, are used to describe the uncertainty about these variables.

Despite the clear definition of probabilistic methods for UT, the formulation of predictive distribution varies across tasks depending on their specific objectives and modeling assumptions:

- In classification, the objective is to learn a parameterized ML model from training data $\mathcal{D}=\{x_i, y_i\}_{i=1}^{N}$, which captures the conditional distribution of output labels $y^*$ given the input data $x^*$. This distribution is typically a categorical distribution and is denoted as $p\left(y^* \middle| x^*, \theta\right)$. Non-Bayesian methods usually assume fixed (often maximum likelihood estimated) model parameters $\hat{\theta}$, i.e., $\hat{\theta} = \arg\max_{\theta} \prod_{i=1}^{N} p\left(y_i \middle| x_i, \theta\right)$, and use a predetermined distribution to model predictions given by $p\left(y^* \middle| x^*, \hat{\theta}\right)$. This formulation primarily captures aleatoric uncertainty from the inherent noise of the input. In contrast, Bayesian methods treat model parameters as uncertain variables governed by a posterior distribution $p\left(\theta \middle| \mathcal{D}\right)$, which can be estimated using Bayes' rule given a prior distribution $p(\theta)$ and the training data. The predictive distribution is then obtained by marginalizing over the posterior:

$$p\left(y^* \middle| x^*, \mathcal{D}\right) = \int_{\theta} p\left(y^* \middle| x^*, \theta\right) p\left(\theta \middle| \mathcal{D}\right) d\theta \tag{1}$$

  This Bayesian formulation provides a more comprehensive representation of uncertainty by accounting for both aleatoric and epistemic components. Based on the predictive distribution, the predictive uncertainty is typically measured using the entropy of this distribution [37], given by $\mathcal{H}\left[p\left(y^*\middle|x^*,\mathcal{D}\right)\right]=-\sum_k p\left(y^*=k\middle|x^*,\mathcal{D}\right)\log p\left(y^*=k\middle|x^*,\mathcal{D}\right)$.

- In regression, a noise variable $\varepsilon$ is introduced to capture uncertainty, which is typically assumed to follow a Gaussian distribution, i.e., $\varepsilon\sim\mathcal{N}\left(0,\sigma^2\boldsymbol{I}\right)$. For non-Bayesian methods, the model's prediction can be expressed as $y^*=f\left(x^*;\hat{\theta}\right)+\varepsilon$, resulting in a predictive distribution of the form $p\left(y^*\middle|x^*,\hat{\theta}\right)=\mathcal{N}\left(y^*\middle|f\left(x^*;\hat{\theta}\right),\sigma^2\boldsymbol{I}\right)$. For Bayesian methods, the predictive distribution is obtained from Eq. (1) and becomes a mixture of Gaussians, which is often approximated using a Gaussian distribution [28, 36]:

$$\begin{aligned}
&p\left(y^*\middle|x^*,\mathcal{D}\right)\approx\mathcal{N}\left(y^*\middle|\bar{\mu},\bar{\sigma}^2\right)\\
&\bar{\mu}=\mathbb{E}_{p(\theta|\mathcal{D})}\left[f\left(x^*;\theta\right)\right]\\
&\bar{\sigma}^2=\mathbb{E}_{p(\theta|\mathcal{D})}\left[\mathrm{Var}\left(y^*\middle|x^*,\theta\right)\right]+\mathrm{Var}_{p(\theta|\mathcal{D})}\left[\mathbb{E}\left(y^*\middle|x^*,\theta\right)\right]=\sigma^2+\mathrm{Var}_{p(\theta|\mathcal{D})}\left[f\left(y^*\middle|x^*,\theta\right)\right]
\end{aligned}\tag{2}$$

  Given the predictive distribution, the predictive uncertainty is typically measured using its variance, i.e., $\bar{\sigma}^2$. Additionally, Eq. (2) enables an explicit decomposition of predictive uncertainty into aleatoric and epistemic components through a variance decomposition approach [26]. Specifically, aleatoric uncertainty can be measured using the variance of the noise variable, i.e., $\mathbb{E}_{p(\theta|\mathcal{D})}\left[\mathrm{Var}\left(y^*\middle|x^*,\theta\right)\right]$, while epistemic uncertainty arising from model parameters can be quantified using the variance of model predictions across different parameters, i.e., $\mathrm{Var}_{p(\theta|\mathcal{D})}\left[\mathbb{E}\left(y^*\middle|x^*,\theta\right)\right]$. This decomposition provides a more interpretable view of the sources of uncertainty in regression tasks and is crucial for high-stakes applications, such as structural dynamics, as it enables the identification of the reducible (epistemic) parts of predictive uncertainty to facilitate developing more robust and trustworthy ML models.

- In unsupervised learning, the formulation of the predictive distribution is generally more challenging and task-specific. Here a common task in structural dynamics, namely anomaly detection, is used as an illustrative example. In this context, a baseline distribution $p(x|\theta)$ is typically established from the training set $\mathcal{D}=\{x_i\}_{i=1}^{N}$, with the model parameters $\theta$ estimated using maximum likelihood estimation (MLE). To detect anomalies, an anomaly score $s(x^*, p(x|\theta))$, is computed to measure the deviation of a test sample $x^*$ from the learned baseline distribution. A threshold is then constructed (often based on the training data) to determine whether $x^*$ is anomalous. This framework captures aleatoric uncertainty stemming from inherent data noise through probabilistic modeling of the baseline condition. To further account for epistemic uncertainty associated with model parameters, $\theta$ can be treated as uncertain variables with a prior distribution $p(\theta)$. The posterior distribution $p(\theta|\mathcal{D})$ can then be derived using Bayes' rule, and the predictive distribution of the anomaly score becomes:

$$p\left(s\left(x^*, p(x|\theta)\right)\middle|\mathcal{D}\right) = \int s\left(x^*, p(x|\theta)\right) p(\theta|\mathcal{D}) d\theta \tag{3}$$

Based on this predictive distribution, the empirical variance of the anomaly score can be calculated to quantify the predictive uncertainty [38], which is given by $\mathrm{Var}\left[s\left(x^*, p(x|\theta)\right)\right] = \frac{1}{T}\sum_{i=1}^{T}\left(s\left(x^*, p(x|\theta_i)\right) - \mathbb{E}_{p(\theta|\mathcal{D})}\left[s\left(x^*, p(x|\theta)\right)\right]\right)^2\Big|_{\theta_i \sim p(\theta|\mathcal{D})}$.

### *2.3.2 Non-probabilistic methods for uncertainty treatment*

This type of UT technique, including interval learning and fuzzy learning, adopts alternative perspectives to probabilistic approaches in addressing predictive uncertainty. Interval learning quantifies the predictive uncertainty through intervals (upper and lower bounds) of model predictions, which can be estimated by propagating the intervals of inputs and model parameters through the ML model. On the other hand, fuzzy learning addresses uncertainty by assigning degrees of membership to data points across various classes or clusters, which reflects the uncertainty or ambiguity in classification or clustering results. These non-probabilistic methods, although not that prevalent compared to probabilistic methods, also hold significant importance in the context of UT for ML models because of their unique approaches to capturing uncertainty. In this survey, particular emphasis is placed on the use of variable intervals (i.e., bounds) for uncertainty quantification and propagation. The formulation of variable intervals varies across tasks, depending on specific objectives:

- In unsupervised learning, a common task is to quantify uncertain variables $x$ using intervals derived from a given set of samples $\{x_i\}_{i=1}^{N}$. A basic approach to determining the interval of the variables is to identify the maximum and minimum values among the samples, i.e.,

$$\begin{aligned}\overline{x} &= \max\{x \in \mathcal{X}\} \\ \underline{x} &= \min\{x \in \mathcal{X}\}\end{aligned} \tag{4}$$

  However, this method may lead to an overestimation of uncertainty, particularly when the data exhibit multiple clusters. To address this, unsupervised learning techniques can be employed to partition the original samples into multiple groups $\{x_i\}_{i=1}^{N} = \{x_{i_1}\}_{i_1=1}^{N_1} \cup \{x_{i_2}\}_{i_2=1}^{N_2} \cup \cdots \cup \{x_{i_n}\}_{i_n=1}^{N_n}$. Subsequently, the intervals can be computed by determining the bounds of each group:

$$\begin{aligned}\overline{x}_{i_j} &= \max\left\{x \in \{x_{i_j}\}_{i_j=1}^{N_n}\right\} \\ \underline{x}_{i_j} &= \min\left\{x \in \{x_{i_j}\}_{i_j=1}^{N_j}\right\}\end{aligned} \tag{5}$$

  By leveraging data structure, unsupervised learning methods enhance the accuracy of non-probabilistic uncertainty quantification.

- In supervised learning, given a set of sample pairs $\{x_i, y_i\}_{i=1}^{N}$ of system inputs $x$ and

outputs $y$, the objective is to predict the intervals of the outputs:

$$\begin{aligned} \bar{y} &= \max\left\{ y = y(x) \mid x \in [\underline{x}, \bar{x}] \right\} \\ \underline{y} &= \min\left\{ y = y(x) \mid x \in [\underline{x}, \bar{x}] \right\} \end{aligned} \tag{6}$$

To achieve this goal, both inputs and outputs are represented as interval numbers, and machine learning models such as neural networks are employed to establish the mapping from inputs to outputs. Denoting the models as $M_p$, the intervals of outputs can be obtained by:

$$\left[\underline{y}, \bar{y}\right] = M_p(x),\ x \in \left[\underline{x}, \bar{x}\right] \tag{7}$$

The prediction methods depend on the models $M_p$. Typical approaches include interval neural networks and the Taylor expansion method, which will be discussed in detail in Section 5.2.

### *2.3.3 Scope of this work and included uncertainty treatment methods*

Based on the categorization of UT techniques, this work offers a systematic and comprehensive survey of representative probabilistic and non-probabilistic ML methods that incorporate UT, along with their applications in both forward and inverse problems of structural dynamics. It is important to note that while ML models inherently rely on empirical data and are grounded in mathematical statistics, and thus often operate within a probabilistic framework, this survey focuses on "uncertainty-aware" ML methods that are able to explicitly represent, accommodate, or quantify uncertainties in input data, model parameters, or both of them, as illustrated in Sections 2.3.1 and 2.3.2. Therefore, to ensure a focused and relevant discussion, critical selection criteria are adopted for the inclusion of ML methods in this work. For more comprehensive discussions on the application of ML models in structural dynamics, one can refer to some existing reviews [9, 14, 15]. In terms of the selection criteria for this work, within

probabilistic methods:

- For shallow learning models, only those that explicitly operate within a probabilistic framework and express their predictions using well-defined probabilistic formulations are included. Models that are typically considered deterministic, such as support vector machines and k-nearest neighbors [39], are excluded as they do not explicitly account for uncertainty.
- For deep learning models, two broad categories that have been widely adopted for uncertainty quantification are included, i.e., ensembles of neural networks and Bayesian neural networks. Other deep learning approaches, such as conventional deep classifiers that use softmax outputs to form predictive distributions, are excluded as these formulations are often empirical and lack a rigorous probabilistic foundation.
- A particular emphasis is placed on ensemble and Bayesian methods due to their effectiveness in representing and interpreting epistemic uncertainty.

For non-probabilistic methods, two categories of methods are included. The first category encompasses non-probabilistic ML models where the model inputs, outputs and/or parameters are quantified in a non-probabilistic manner. The second category pertains to the application of conventional ML models for non-probabilistic uncertainty quantification and propagation. Specifically, techniques such as principal component analysis, clustering methods, and generative models are employed to effectively quantify and propagate non-probabilistic uncertainties.

Additionally, as this survey emphasizes the application domain of structural dynamics, only ML methods that have demonstrated substantial practical use or show strong potential for application in this field are included. This domain-driven focus ensures that the discussion remains relevant to both ML researchers interested in UT and practitioners in structural dynamics seeking reliable and interpretable ML solutions. Consequently, reinforcement learning methods are not included due to their limited applications to UT in structural dynamics.

With these criteria in mind, an overview of the methods covered in this work is presented in Fig. 3, with the aim of enhancing the clarity of this survey. Meanwhile, concise introductions to these methods are provided in Sections 3-5, focusing on their approaches to capturing and/or quantifying the predictive uncertainty. Without delving into the detailed mechanisms of each method, this survey strives to assist researchers and practitioners in making well-informed decisions when utilizing ML techniques to address uncertainties in the field of structural dynamics.

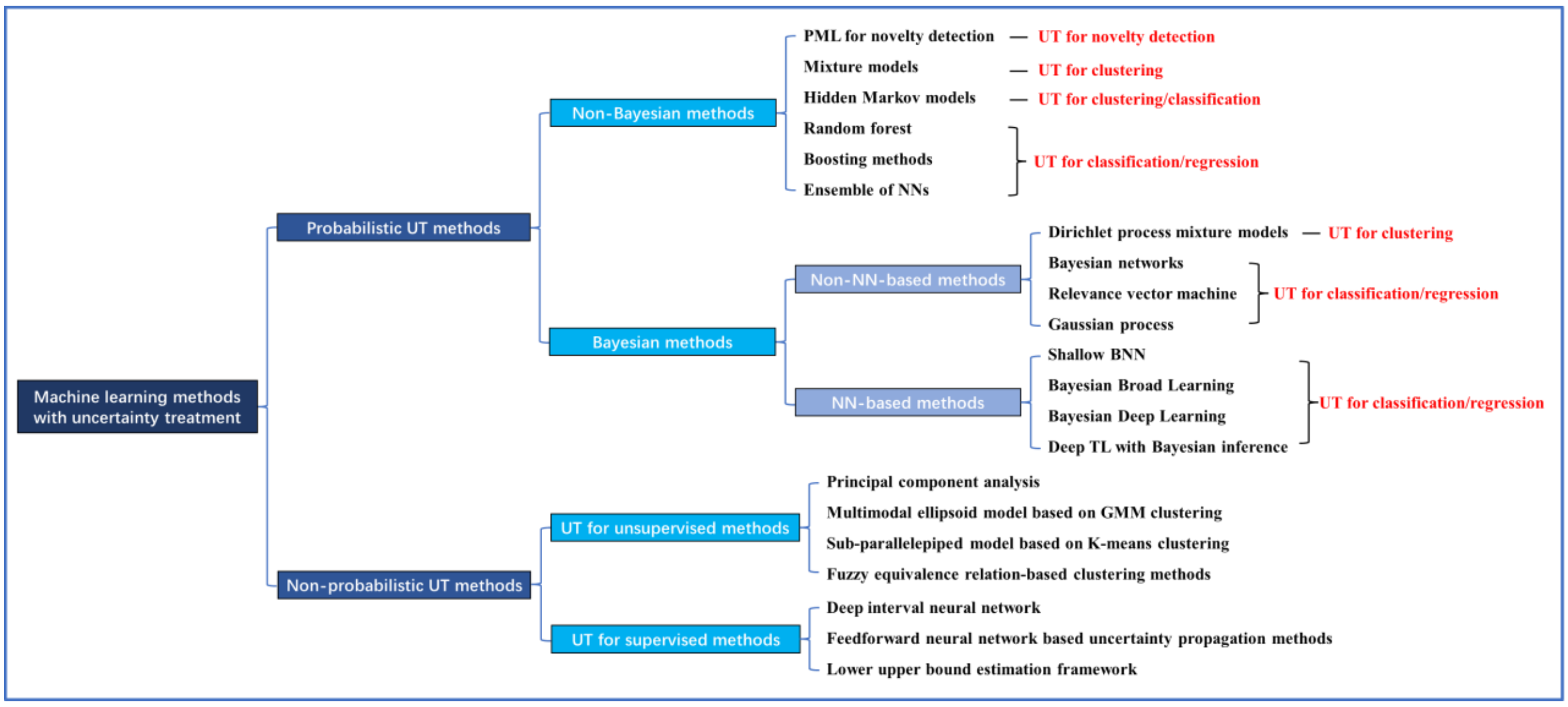

Figure 3. Overview of the ML methods with uncertainty treatment elucidated in this study.

## 3 Exploring Non-Bayesian Techniques for Probabilistic Machine Learning

Currently, probabilistic UT methods have fallen into the mainstream of UT for ML models [17, 22, 23, 27, 28, 40], which can be further divided into two branches corresponding to the two perspectives of probability, i.e., the frequentist (non-Bayesian) and Bayesian perspectives. Compared to their Bayesian counterparts, non-Bayesian approaches are more flexible and computationally efficient, but many of them struggle with capturing epistemic uncertainty as they rely on point estimates of model parameters. This section provides succinct introductions to some state-of-the-art ML models employing non-Bayesian UT techniques, accompanied by detailed, tutorial-style analysis on the way of each method to capturing and estimating the predictive uncertainty. Meanwhile, some representative applications of these UT methods in structural dynamics are summarized in Table 2, along with discussions on their distinct advantages and limitations.

Table 2. Selection of literature on the applications of non-Bayesian ML methods with UT in structural dynamics.

<table>
<tr><th colspan="3">UT method</th><th>Advantages</th><th>Disadvantages</th><th>Ref.</th></tr>
<tr><td rowspan="4">Non-Bayesian</td><td rowspan="2">PML for anomaly detection</td><td>PD</td><td>flexibility; efficiency</td><td>require assumptions on underlying data distribution</td><td>[41-46]</td></tr>
<tr><td>EVT</td><td>interpretability; efficiency</td><td>limited to univariate distributions; sensitive to model choice</td><td>[47-52]</td></tr>
<tr><td colspan="2">Mixture Models</td><td>flexibility; generative modeling</td><td>difficult to determine appropriate model complexity; sensitive to initialization</td><td>[53-56]</td></tr>
<tr><td colspan="2">HMM</td><td>capturing uncertainty propagation</td><td>strong assumptions for state space; sensitive to initialization</td><td>[57-59]</td></tr>
</table>

| | | | | |
|---|---|---|---|---|
| | RF | flexibility; outlier robustness | sensitive to imbalanced and noisy data | [60-64] |
| | Boosting | providing the importance of samples to indicate the driving factors behind uncertainty | computational expensive due to limited parallelization; sensitive to biased data | [61, 65-69] |
| | Ensemble of NNs | potential of uncertainty quantification; robust to OOD samples; efficiency | lack of interpretability; difficult to tune hyperparameters | [70-73] |

### *3.1 Probabilistic ML methods for anomaly detection with UT*

Anomaly detection (AD), also referred to as novelty or outlier detection, aims to construct a baseline using training data, and any deviation of testing data from this baseline is defined as an anomaly. In structural dynamics, AD is commonly used for structural damage detection by attributing the detected anomalies to the effect of damage [3]. UT in this context mainly relies on probabilistic modeling of the baseline, aiming to mitigate the effect of uncertainty arising from data noise and enable more robust structural condition assessment. Generally, there are two probabilistic machine learning (PML) methods for AD considering uncertainties:

- **Probabilistic distance (PD)**: PD-based methods account for aleatoric uncertainty by measuring differences between distributions instead of individual samples. A common example is the Mahalanobis squared distance (MSD), which assumes the baseline is a multivariate Gaussian distribution [3] and quantifies the deviation of a test point from this

baseline as:

$$\mathrm{MSD} = \left(x-\bar{x}\right)^{T} \Sigma^{-1}\left(x-\bar{x}\right) \tag{8}$$

where $\bar{x}$ and $\Sigma$ are the mean vector and covariance matrix estimated from the training data. In this case, the threshold can be constructed through an MC sampling approach [45]. In addition to assessing the discordancy of individual data points, the difference between the distribution of test data and the baseline can also be measured using PDs such as Kullback-Leibler (KL) divergence and Bhattacharyya distance [43, 44, 74]. Owing to their foundation in statistical modeling of the baseline, PD-based methods generally exhibit robustness to various types of noise [3, 43-45].

- **Extreme value theory (EVT)**: EVT-based methods address uncertainty by modeling the statistical behavior of extreme values in the training set, rather than modeling the full data distribution. As a branch of probability theory, EVT focuses on the tails of distributions and models the extreme values (i.e. maximum or minimum samples) using specific distribution forms [3, 51]. Given a set of independent and identically distributed (i.i.d.) samples, EVT provides an asymptotic approximation of their cumulative density function (CDF) when there exist constants $a_n > 0$ and $b_n \in \Re$ such that the following limit converges to a non-degenerate distribution:

$$\lim_{n\to\infty} \Pr\left(\frac{M_n - b_n}{a_n} \le z\right) = G(z) \tag{9}$$

where $M_n$ is the maximum of the samples. On this basis, the limiting distribution $G(z)$ belongs to the generalized extreme value distribution (GEVD):

$$G(z) \propto \exp\left\{-\left(1+\varsigma z\right)^{-\frac{1}{\varsigma}}\right\} \tag{10}$$

where $\varsigma$ depends on the tail shape of the distribution and when normalized, $G(z)$ belongs to Weibull, Gumbel or Fréchet distributions. For more details about EVT, one can refer to [75]. Compared to other AD methods, a key advantage of EVT-based methods is that the extreme value distribution is restricted to three categories, thereby reducing the uncertainty associated with selecting a baseline distribution.

In summary, these PML methods for AD address aleatoric uncertainty through probabilistic modeling of the baseline, leading to more robust detection results. However, their performance can be sensitive to epistemic uncertainty arising from limited knowledge or insufficient training data. For example, selecting an inappropriate distribution to model the training data or choosing unsuitable extreme values can significantly compromise the effectiveness of these methods.

### *3.2 Mixture models for UT in clustering*

Clustering aims to partition observed data into groups based on their intrinsic similarities, which can be flexibly applied in unsupervised, supervised, and semi-supervised learning scenarios [76]. Among various clustering methods, mixture models are prominent probabilistic approaches that naturally account for uncertainty by modeling the dataset using a combination of probability distributions. Specifically, each sample $x_i$ is assumed to be drawn from a mixture of $K$ components, with each representing a specific distribution $p_k\left(x \mid \theta_k\right)$. The most widely used mixture model is the Gaussian mixture model (GMM), where each component is a Gaussian distribution, i.e., $p_k\left(x \mid \theta\right) = \mathcal{N}\left(x \mid \mu_k, \mathbf{\Sigma}_k\right)$. The resulting mixture distribution is given by:

$$p\left(x \middle| \theta\right) = \sum_{k=1}^{K} \pi_k \mathcal{N}\left(x \mid \mu_k, \mathbf{\Sigma}_k\right) \tag{11}$$

where $\pi_k$ denotes the mixing proportion of the $k$ th component, and it has $\sum_{k=1}^{K} \pi_k = 1$.

In mixture model-based clustering, each component corresponds to a distinct cluster. To determine the clustering result, a latent variable $z_i$ is introduced to denote the cluster assignment for $x_i$, i.e., $z_i = k$ indicates that $x_i$ belongs to the $k$ th component. The expectation-maximization (EM) algorithm is commonly used to estimate model parameters and latent variables through MLE. In the E-step, the responsibility $r_{i,k}$ is computed using current parameters $\pi_k, \mu_k, \boldsymbol{\Sigma}_k$, which represents the probability that component $k$ takes for explaining the observation $x_i$ [77]. In the M-step, $\pi_k, \mu_k, \boldsymbol{\Sigma}_k$ are updated using the responsibilities. These two steps are iterated until the convergence of either the parameters or the likelihood. For more details about GMMs and the EM algorithm, one can refer to [29, 39, 77]. In this framework, the responsibilities of each sample form a categorical distribution, i.e. $p(z_i) = \mathrm{Cat}(K, r_i)$ with $r_i = \{r_{i,1}, r_{i,2}, \ldots, r_{i,K}\}$, which captures the aleatoric uncertainty associated with cluster assignments, as shown in Fig. 4. However, a major limitation of mixture models is the challenge of determining the optimal number of components $K$. This difficulty arises from the frequently inadequate prior knowledge about the underlying data generation process in real-world applications, which underscores the weakness of traditional mixture models in addressing epistemic uncertainty.

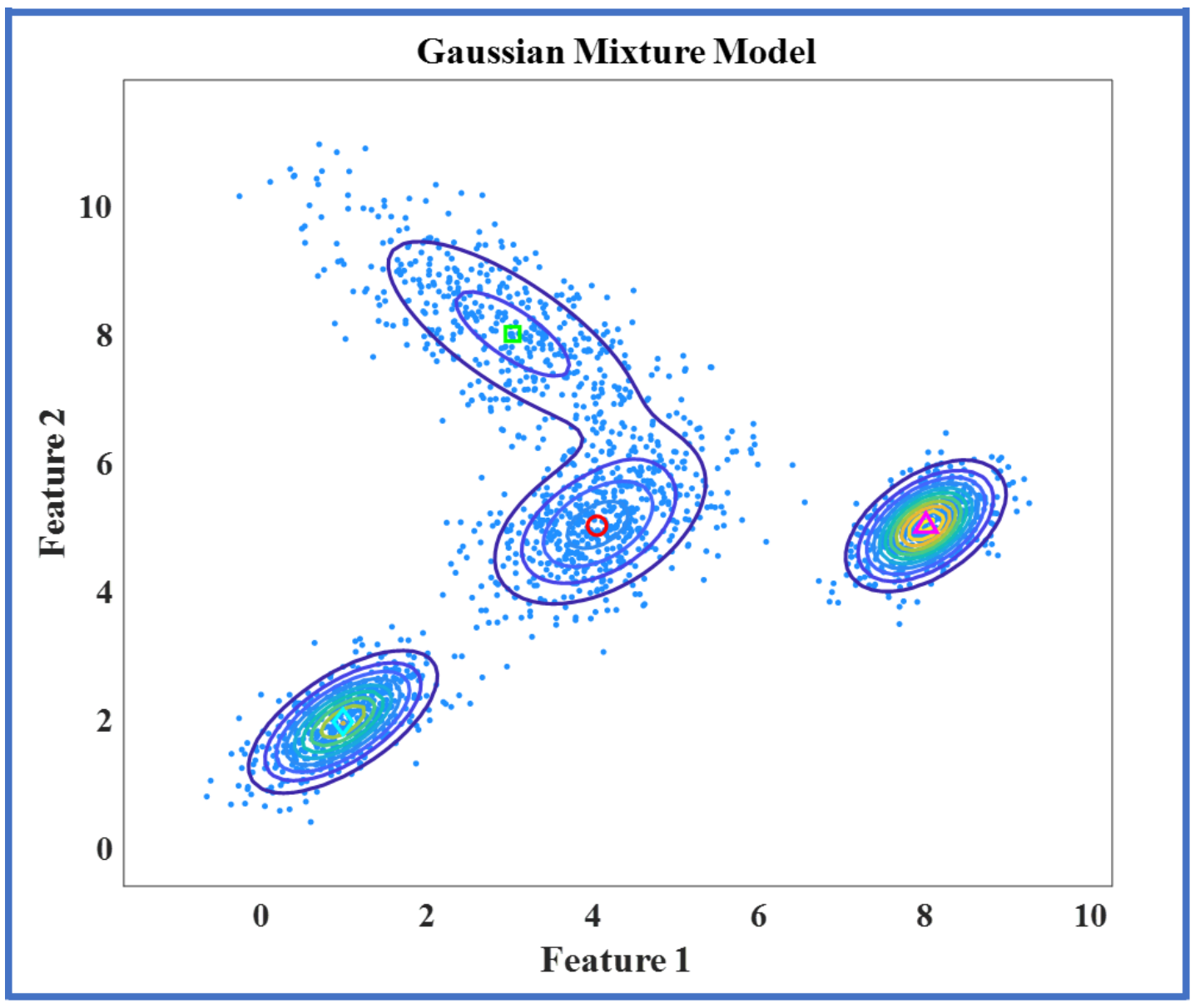

Figure 4. Schematic diagram of clustering result of a GMM, with the contours reflecting uncertainty associated with cluster assignments.

### *3.3 Hidden Markov models for UT in clustering and classification*

Hidden Markov models (HMMs) are probabilistic graphical models designed to model sequential data, which assume that the sequence of observed data is generated from a sequence of discrete hidden states [78], as shown in Fig. 5. An HMM can be represented by the hidden states $\{z_1, z_2, \ldots z_T\}$ and the observation models $p(x_t|z_t)$, with their joint distribution given by:

$$p(x_{1:T}, z_{1:T}) = p(z_{1:T})\, p(x_{1:T}|z_{1:T}) = p(z_1)\prod_{t=2}^{T} p(z_t|z_{t-1})\prod_{t=1}^{T} p(x_t|z_t) \tag{12}$$

As the hidden states are discrete, HMMs are also widely used for clustering and classification, with each hidden state representing a cluster or class [79]. The uncertainty from state transition in each step is captured through a conditional distribution $p(z_t|z_{t-1})$, which reveals the propagation of uncertainty from the initial state to the final state. Moreover, the observation model $p(x_t|z_t)$ addresses the uncertainty arising from data noise, which is usually assumed to be a Gaussian distribution for continuous $x_t$, i.e., $p(x_t|z_t=k,\theta)=\mathcal{N}(x_t|\mu_k,\boldsymbol{\Sigma}_k)$, or a categorical distribution for discrete $x_t$, i.e., $p(x_t|z_t=k,\theta)=\text{Cat}(x_t|\theta_k)$ [29]. Given the state transition probability and the observation model, the predictive distributions $p(z_{T+1}|x_{1:T})$ and $p(x_{T+1}|x_{1:T})$ can be derived using forward or forward-backward algorithms [29], enabling explicit quantification of predictive uncertainty. However, HMMs can suffer from epistemic uncertainty due to the large number of parameters required for defining the transition matrices and observation models, which can lead to increased model complexity and the risk of overfitting. Furthermore, model selection for HMMs is challenging, as the relationship between states and observations is often unconstrained and may take on arbitrary forms or even involve high nonlinearity. To overcome this limitation, HMMs are sometimes combined with Bayesian methods to address epistemic uncertainty. For more comprehensive information about HMMs, one can refer to [39, 80].

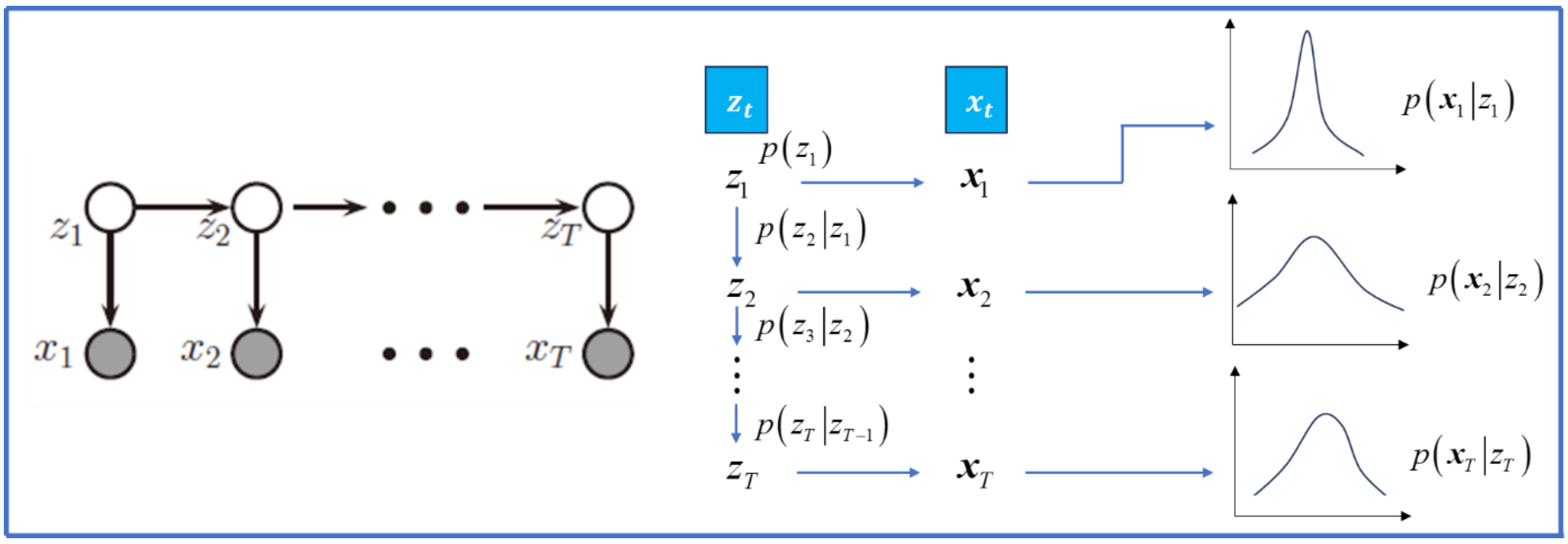

Figure 5. Schematic illustration of the structure and UT method of the HMM.

### *3.4 Random forests for UT in classification and regression*

Random forests (RFs) are ensembles of decision trees designed to mitigate the instability and high variance associated with predictions of individual trees, which can be applied to both classification and regression tasks. In an RF, individual decision trees are trained concurrently on different subsets of data, with each subset constructed by randomly sampling from the entire dataset with replacement (bootstrapped sampling) [29]. A schematic illustration of an RF is shown in Fig. 6, and its output can be expressed as:

$$f(x) = \frac{1}{M}\sum_{i=1}^{M} f_i(x|\theta_i) \tag{13}$$

where $f_i(x|\theta_i)$ denotes the prediction of the $i$ th decision tree. By aggregating predictions from multiple trees trained on different datasets, RFs normally exhibit improved predictive performance and alleviate overfitting [29] due to capturing the uncertainty arising from the randomness in the training phase. In addition, RFs can also address epistemic uncertainty by combining decision trees with different structures. The predictive uncertainty in regression tasks can be quantified by the empirical variance of predictions across individual decision trees [81], as illustrated in Eq. (2). In classification tasks, it is typically measured using the entropy [82], as described in Section 2.3.1.

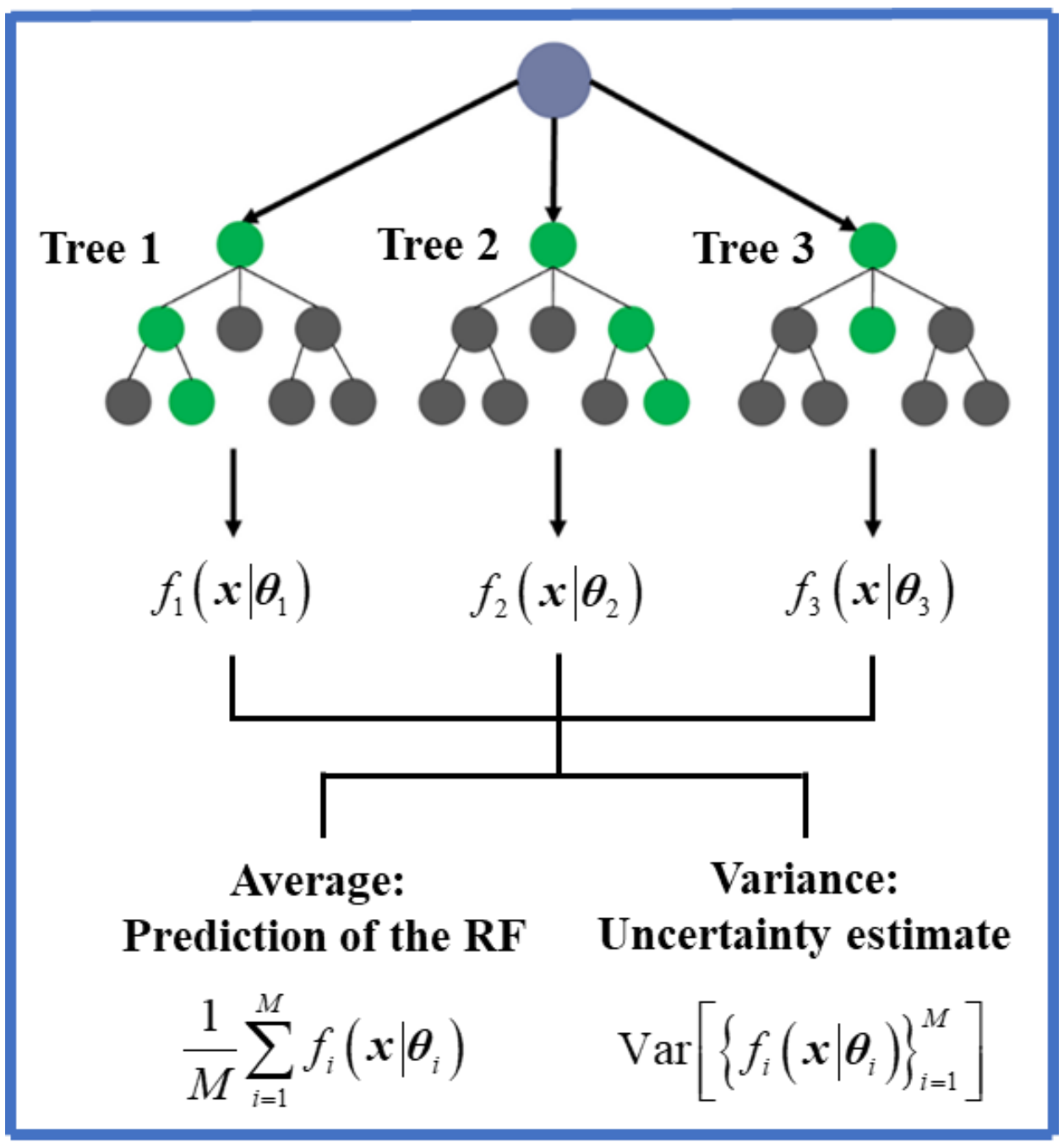

Figure 6. Schematic overview of a random forest for UT in regression.

### *3.5 Boosting for UT in classification and regression*

Boosting is an ensemble method that combines a series of base learners trained sequentially, as illustrated in Fig. 7. Before training each base learner, the training data (or gradients) are reweighted to assign greater importance to samples that were misclassified or poorly predicted by previous models [29, 83]. The final prediction of the boosted model is a weighted sum of predictions of individual base learners, which is given by:

$$f(x) = \sum_{i=1}^{M} w_i f_i(x|\theta_i) \tag{14}$$

where $w_i$ denotes the weights of the $i$th base learner $f_i(x|\theta_i)$ and $\sum_{i=1}^{M} w_i = 1$. Compared to RFs, which focus on reducing variance, boosting mainly targets reducing bias in predictions [83]. However, boosting is susceptible to noisy data, as its weighting mechanism can amplify the impact of outliers and potentially lead to overfitting [83], highlighting its limitation in handling heteroscedastic data uncertainty. Moreover, while boosting methods can address uncertainties, especially epistemic uncertainties arising from model parameters, by aggregating independent models, they typically lack explicit mechanisms for interpreting such uncertainties. Nevertheless, boosting remains one of the most powerful techniques for enhancing predictive performance with many successful applications [29]. Over the years, various booting methods have been proposed to enhance both predictive performance and the ability to handle uncertainty, with the two most renowned ones being *AdaBoost* and *XGBoost*.

In the field of structural dynamics, these boosting methods have been applied to tasks such as damage identification [66, 68, 84] and response prediction [85, 86]. However, most of these applications mainly focus on improving the detection or prediction accuracy by combining the outputs of different ML models, with few explicitly leveraging boosting techniques to address uncertainties inherent in structural dynamics. This highlights a promising direction for further research. A pioneering work in this area was proposed by Wen et al. [87], where a light gradient boosting machine, combined with a bootstrap resampling scheme, was developed to decompose and quantify both aleatoric and epistemic uncertainties in ground motion models.

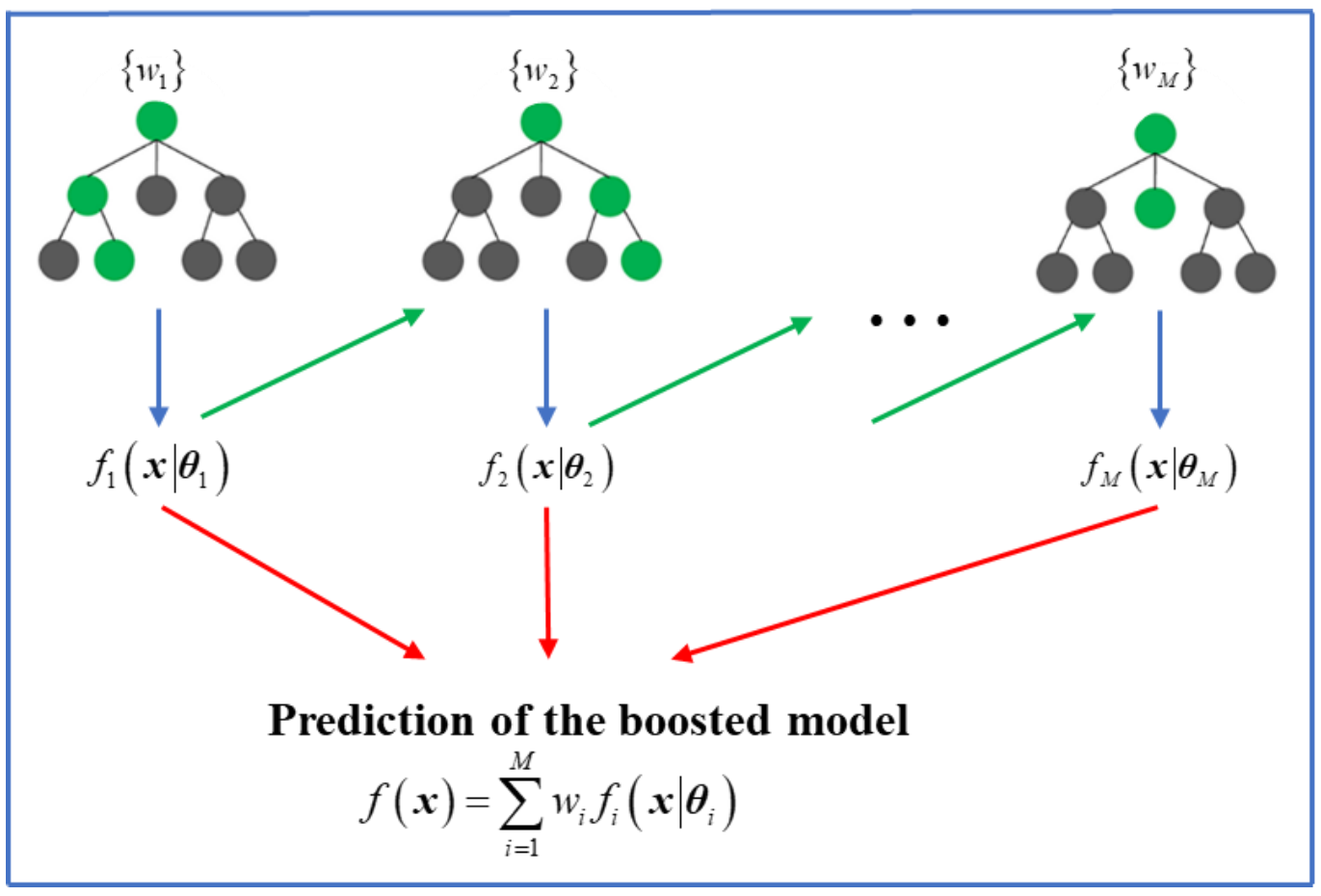

Figure 7. Schematic illustration of a boosted model.

### *3.5.1 AdaBoost*

AdaBoost, short for adaptive boosting, is a boosting method mainly used for binary classification [29]. Conditioned on a binary classification task with the training set $\{x_1, x_2, ..., x_N\}$ and corresponding target labels $\{y_1, y_2, ..., y_N\}$, $y_i \in \{-1, +1\}$, AdaBoost initializes by assigning equal weights $\alpha_n^{(1)} = \frac{1}{N}$ to each data point. Then, the algorithm trains each base learner $f_i(x|\theta_i)$, $i = 1, 2, ..., M$ sequentially by minimizing a weighted loss function:

$$\mathcal{L}_i = \sum_{n=1}^{N} \alpha_n^{(i)} I\left[ f_i\left(x_n|\theta_i\right) \neq y_n \right] \tag{15}$$

where the superscript or subscript $i$ denotes the $i$th base learner; $I\left[ f_i\left(x_n|\theta_i\right) \neq y_n \right]$ is an indicator function, which equals 1 when $f_i\left(x_n|\theta_i\right) = y_n$ and 0 otherwise. For a new base learner, the weights $\alpha_n$ are adjusted according to the weighted measures of the error rates of the previously trained learners. After the training of all base learners, the prediction of the AdaBoost model can be derived according to Eq. (14). By aggregating individual base learners, AdaBoost effectively converts their deterministic classification outputs into a Bernoulli-distributed prediction, thereby enabling the estimation of predictive uncertainty. In addition to

binary classification, AdaBoost can also be extended to multiclass classification and regression by employing different loss functions and target variables [77].

### *3.5.2 XGBoost*

XGBoost, short for extreme gradient boosting, is a scalable and highly accurate variant of gradient boosting. Gradient boosting is a boosting technique that trains a set of base learners sequentially through gradient descent optimization, with each base learner trained using a specified loss function with respect to the false residuals (the difference between the output of previous base learners and the target) [83]. XGBoost employs the gradient boosting framework but introduces a regularization term into the loss function to penalize the complexity of the boosted model [88], which prevents overfitting and enhances the robustness of predictions of the boosted model. By combining the predictions of base learners trained through regularized gradient descent optimization, XGBoost exhibits competitive predictive performance in many real-world scenarios [83]. Compared with AdaBoost, which sequentially trains each base learner by reweighting the training data, XGBoost introduces an approximate algorithm that finds splitting points for its base learners based on weighted percentiles of feature distribution. This approximation algorithm, combined with a sparse-aware split finding scheme, enables more efficient training of each base learner. Additionally, XGBoost supports parallel and distributed learning through a column subsampling strategy to handle the training data [88]. These attributes make XGBoost more efficient, scalable, and robust to outliers compared to AdaBoost. However, due to the use of approximations and various optimization techniques, XGBoost is generally less interpretable, making AdaBoost might be a better choice for certain smaller-scale engineering problems where interpretability is a primary concern.

### *3.6 Ensemble of NNs for UT in classification and regression*

NNs form the backbone of deep learning (DL), which dominates modern ML due to its strong feature extraction and nonlinear modeling capabilities [89]. However, a well-recognized limitation of NN-based models is that they are prone to generate unexpected, incorrect, but overconfident predictions [16-18], which highlights the significance of UT in NNs. While Bayesian approaches are commonly used to address uncertainty associated with NNs due to their advantages in interpreting and estimating predictive uncertainty [17, 22, 23, 27, 28, 36, 90], ensemble methods provide a promising alternative from a frequentist perspective [27, 36, 91, 92]. Like other ensemble methods, ensemble of NNs has long been recognized as a powerful tool to improve the predictive performance of NNs, but investigation on its usefulness for estimating predictive uncertainty has a much shorter history. The pioneering work on this topic was proposed by Lakshminarayanan et al. [36], where they introduced a simple and scalable framework for predictive uncertainty quantification based on ensemble of NNs. In this framework, a set of NNs is trained in parallel via bootstrapping and adversarial training, and the ensemble is treated as a uniformly-weighted mixture model for prediction and uncertainty estimation. For classification, this corresponds to averaging the predicted softmax probabilities of each NN, while the negative log-likelihood serves as an estimate of predictive uncertainty [36]. For regression, each NN outputs both the mean and variance based on a Gaussian assumption [93], resulting in an ensemble predictive distribution modeled as a mixture of Gaussians. This predictive distribution is then approximated using a Gaussian distribution to facilitate uncertainty estimation [36], as described in Section 2.3.1. A schematic overview of this approach is shown in Fig. 8. Compared to Bayesian methods, ensemble of NNs provides a more computationally efficient alternative by training a set of simple NNs in parallel. In addition, it also offers a unique approach to UT for NNs from the frequentist perspective, which has been reported to produce well-calibrated uncertainty estimates [36]. However, in the field of structural dynamics, limited research has explored the application of this method for UT, resulting in a research gap that warrants further investigation.

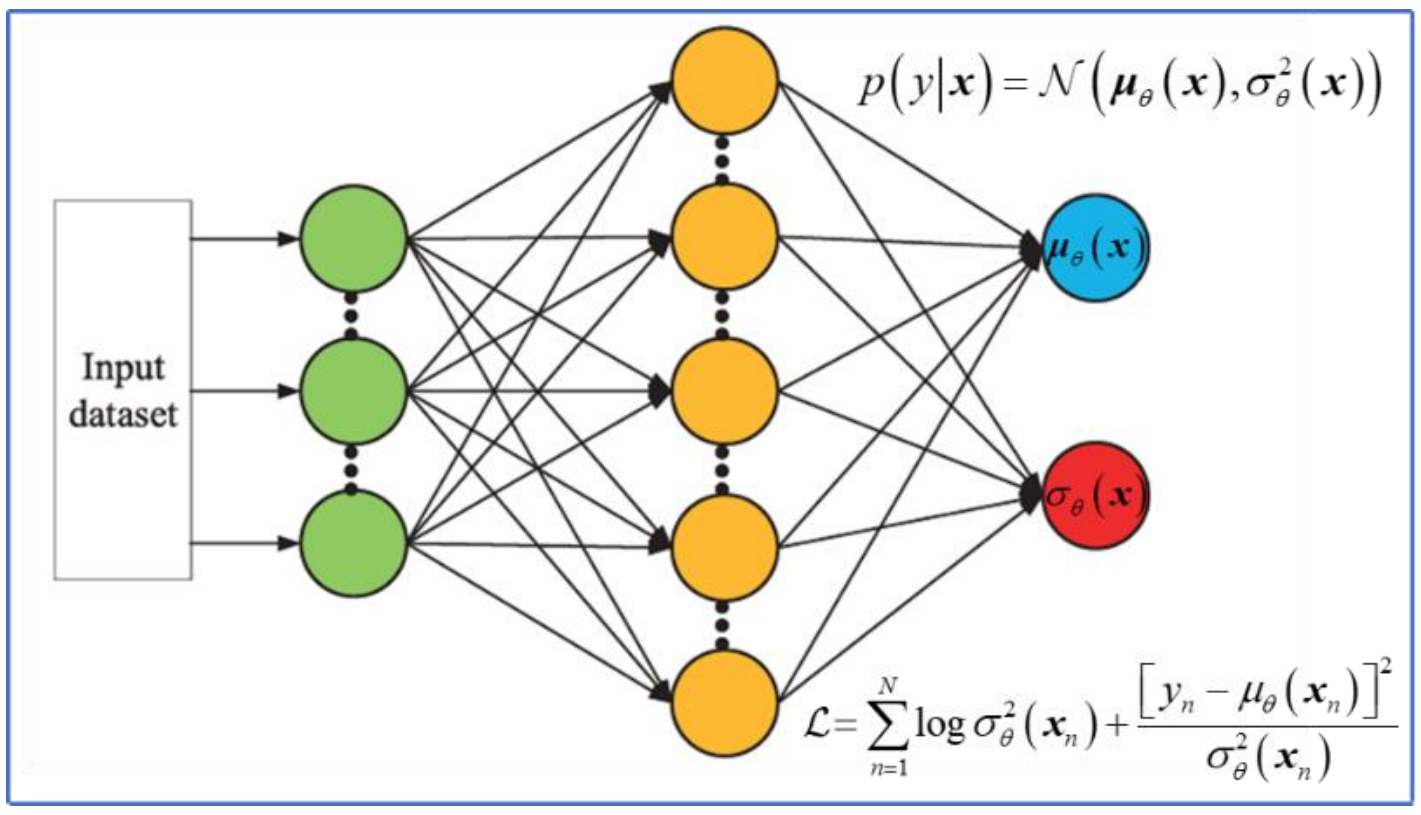

(a)

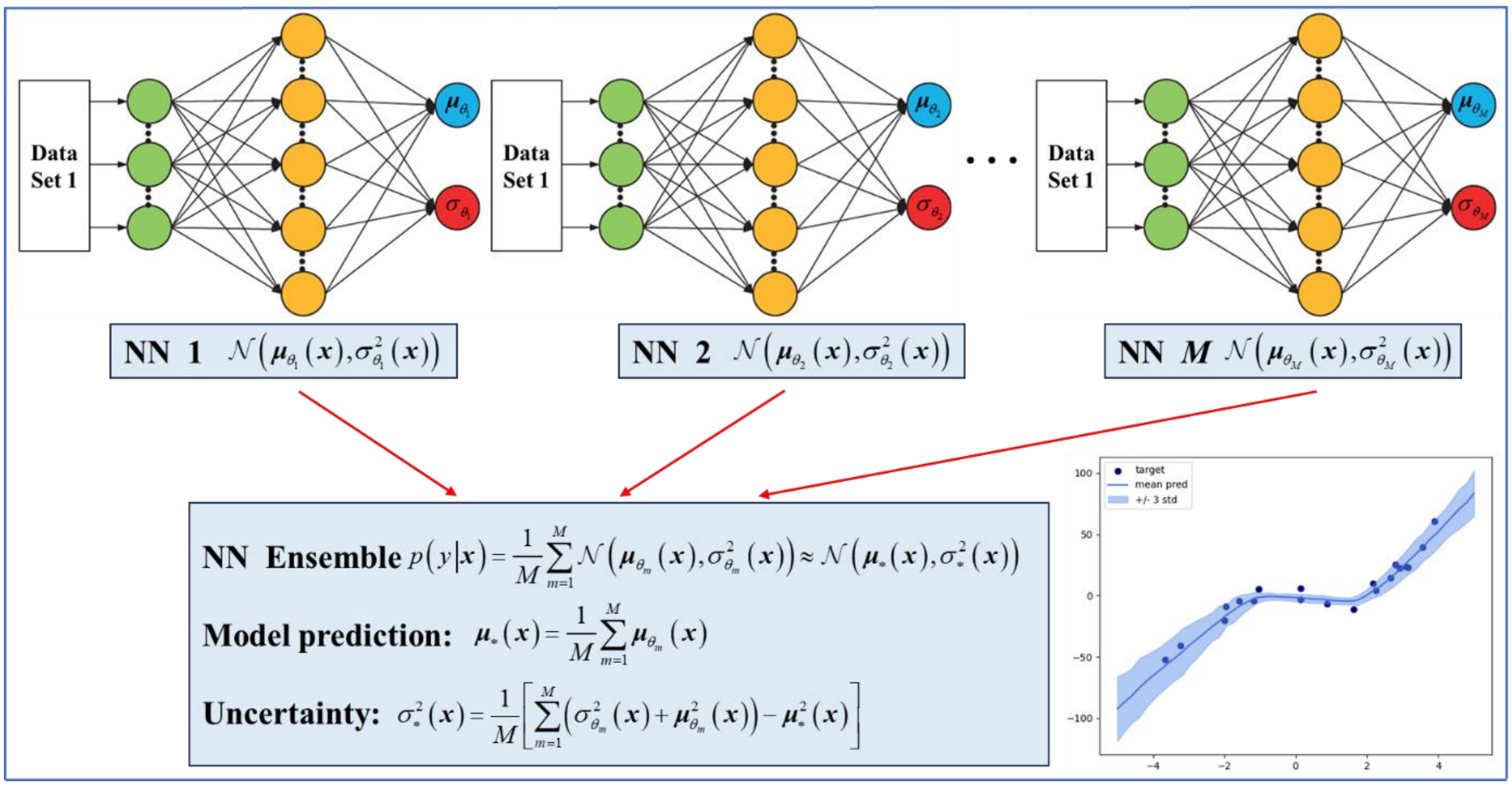

(b)

Figure 8. Schematic illustration of (a) mean-variance estimation NN and (b) uncertainty estimation using an ensemble of mean-variance estimation NNs in regression tasks.

# 4 Exploring Bayesian Approaches for Probabilistic Machine Learning

Bayesian methods provide a brand-new perspective for interpreting and estimating uncertainty in ML models by treating model parameters as uncertain variables rather than point estimates. This fundamental distinction from frequentist methods has led to the widespread acknowledgment of the advantages of Bayesian methods in UQ [17, 23, 28, 86]. However, it is also worth noting that Bayesian approaches are generally more computationally expensive and involve more assumptions in the probabilistic modeling process. This work places significant emphasis on the UT methods for ML models within the Bayesian framework, along with their applications in the field of structural dynamics. A particular focus is given to BNNs due to the dominance of NN-based models in modern ML techniques, as well as the commonly recognized importance of UT for NNs to prevent their overconfident predictions. Table 3 outlines some representative applications of these Bayesian UT methods in the field of structural dynamics, each of which will be elaborated in subsequent sections.

Table 3. Selection of literature on the applications of Bayesian UT methods for ML models in structural dynamic problems.

| UT method | | | Advantages | Disadvantages | Ref. |
|---|---|---|---|---|---|
| Bayesian | Non-NN methods | DPMM | flexibility; automatic choice of model complexity due to the nonparametric nature | computationally expensive; sensitive to prior choices | [38, 94-98] |
| | | BN | causal inference; interpretability; efficiency | requiring expert knowledge to define the conditional probabilities; limited scalability | [99-104] |
| | | RVM | capabilities of sparse model representation and nonlinear mapping | sensitive to choices of hyperparameters and noisy data | [105-107] |

| | | | | |
|---|---|---|---|---|
| | GP (BAL-GP, DGP) | flexibility; nonlinear mapping capability | limited efficiency for high-dimensional data; sensitive to choices of kernels and hyperparameters | [108-120] |
| NN-based methods | VAE | capabilities of feature extraction and generative modeling | ignoring the uncertainty from NNs | [121-126] |
| | Shallow BNN | flexibility; nonlinear mapping capability | limited capability of representation learning | [25, 127-129] |
| | BBL | flexibility; efficiency in network training | limited capability of representation learning | [130, 131] |
| | BDL | better capabilities of representation learning, nonlinear mapping, and epistemic uncertainty modeling than other methods | highest computational cost; limited interpretability; gradient vanishing or exploding problems | [132-148] |

### *4.1 Fundamental of Bayesian ML*

In general, Bayesian ML methods rely on the posterior distribution of model parameters to capture model uncertainty and formulate the (posterior) predictive distribution for uncertainty estimate. The posterior distribution of model parameters, $\theta$, is derived through the application of Bayes' theorem, as illustrated in Eq. (16). Meanwhile, the (posterior) predictive distribution can be obtained by marginalizing out the parameters given their posterior distribution, which is usually approximated using MC integration when the closed form solution is intractable, as shown in Eq. (17).

$$p(\theta|\mathcal{D})=\frac{p(\theta)p(\mathcal{D}|\theta)}{\int p(\mathcal{D}|\theta)p(\theta)d\theta}=\frac{p(\theta)p(\mathcal{D}|\theta)}{p(\mathcal{D})} \tag{16}$$

$$p(y \mid x,\mathcal{D}) = \int p(y \mid x,\theta)\, p(\theta \mid \mathcal{D}) d\theta \approx \frac{1}{N}\sum_{i=1}^{N} p(y|x,\theta_i)\Big|_{\theta_i \sim p(\theta|\mathcal{D})} \tag{17}$$

where $p(\theta)$ incorporates prior knowledge about the model parameters and is independent to the observed data $\mathcal{D}$; $p(\mathcal{D} \mid \theta)$ is the likelihood that denotes the probability of observing $\mathcal{D}$ given a fixed value of $\theta$; the marginal likelihood $p(\mathcal{D})$ is known as the evidence, which is actually the probability density function (PDF) of observing data $\mathcal{D}$ irrespective of any specific hypothesis; $p(y \mid x,\mathcal{D})$ is the model predictive distribution given a new observation $x$, with its empirical mean denoting the optimal prediction and empirical variance serving as an estimate of predictive uncertainty. Compared to non-Bayesian methods, Bayesian methods incorporate prior knowledge naturally, offering a more intrinsic approach to describing uncertainty in model parameters or indeed in model selection [77]. Moreover, Bayesian methods systematically mitigate the issue of bias, which lies at the root of the problem of overfitting [77]. This issue is frequently encountered in MLE, which is a widely used approach in non-Bayesian ML methods. However, despite the advantages of Bayesian methods in UT and model robustness enhancement, the posterior distribution is often analytically intractable in real applications due to the marginalization. This limitation has hindered the widespread use of Bayesian ML methods until the development of approximation techniques for estimating the posterior, with Laplace approximation, variational inference (VI), and Markov chain Monte Carlo (MCMC) sampling being among the most commonly used methods [23].

**(1) *Laplace approximation***

The Laplace approximation aims to find a Gaussian approximation of the true posterior through the maximum a posterior (MAP) estimation of model parameters, which is achieved based on a Taylor expansion of the negative of the logarithm of the posterior distribution [43,

77, 149]. Through Laplace approximation, the posterior distribution can be expressed as:

$$p(\theta|\mathcal{D}) \approx p(\hat{\theta}|\mathcal{D})\exp\left\{-\frac{1}{2}(\theta-\hat{\theta})H(\hat{\theta})(\theta-\hat{\theta})\right\} \propto \exp\left\{-\frac{1}{2}(\theta-\hat{\theta})H(\hat{\theta})(\theta-\hat{\theta})\right\} \tag{18}$$

which means that the posterior can be approximated via the Gaussian distribution $\mathcal{N}(\hat{\theta}, H^{-1})$, with $\hat{\theta}$ being the MAP estimate of $\theta$ and $H(\hat{\theta})$ denoting the Hessian matrix whose elements are second derivatives of the negative logarithm of the posterior distribution, i.e., $H_{ij} = -\left.\frac{\partial \log p(\theta|\mathcal{D})}{\partial\theta_i\partial\theta_j}\right|_{\theta=\hat{\theta}}$. Despite the computational simplicity of Laplace approximation, a critical issue of this approach is that the approximated posterior distribution is a Gaussian centered at a local mode of the true posterior [77], making it inadequate for cases where the true posterior is a multi-modal distribution. Moreover, for large scale ML models such as DNNs, computing the inverse of the Hessian matrix in Eq. (18) remains challenging or even infeasible [17, 150], which constrains the widespread application of Laplace approximation in deriving the posterior distribution and addressing uncertainty in ML models.

**(2)** ***Variational inference***

Different from the Laplace approximation, which relies on a Gaussian approximation to a local mode of the true posterior, VI seeks a more tractable approximation to the posterior using more global criteria, and thus has been broadly applied [77]. Specifically, assume the approximation is denoted by $q_\phi(\theta)$ with a set of free parameters $\phi$, VI aims to minimize the KL divergence between $q_\phi(\theta)$ and the true posterior by tuning the variational parameters $\phi$:

$$D_{KL}\left[q_\phi(\theta)\|p(\theta|\mathcal{D})\right] = \int q_\phi(\theta)\ln\left\{\frac{q_\phi(\theta)}{p(\theta|\mathcal{D})}\right\}d\theta \tag{19}$$

To minimize the KL divergence, Eq. (19) are normally decomposed as follows [151]:

$$
\begin{aligned}
D_{KL}\left[q_\phi(\theta)\|p(\theta|\mathcal{D})\right] &= \log p(\mathcal{D}) - \int q_\phi(\theta)\ln\left\{\frac{p(\mathcal{D},\theta)}{q_\phi(\theta)}\right\}d\theta \\
&= \log p(\mathcal{D}) - \mathbb{E}_{\theta\sim q_\phi(\theta)}\left[\log p(\mathcal{D},\theta)\right] + \mathbb{E}_{\theta\sim q_\phi(\theta)}\left[\log q_\phi(\theta)\right] \\
&= \log p(\mathcal{D}) - \mathcal{L}(q)
\end{aligned} \tag{20}
$$

where $\mathcal{L}(q) = \mathbb{E}_{\theta\sim q_\phi(\theta)}\left[\log p(\mathcal{D},\theta)\right] - \mathbb{E}_{\theta\sim q_\phi(\theta)}\left[\log q_\phi(\theta)\right]$ is referred to as the evidence lower bound (ELBO). As $\log p(\mathcal{D})$ is independent to $\phi$, minimizing $D_{KL}\left[q_\phi(\theta)\|p(\theta|\mathcal{D})\right]$ is equivalent to maximizing the ELBO. Therefore, VI effectively transforms inferring the posterior into an optimization problem, which is actually to find a tradeoff between prediction accuracy and model complexity [17]. Originally, the optimization relies on a fully factorized variational distribution (mean-field VI), namely $q_\phi(\theta) = \prod_{i=1}^{M} q_{\phi_i}(\theta_i)$. Based on this assumption, the optimal distribution of each factor can be derived analytically as follows [77]:

$$
q_{\phi_i}^*(\theta_i) = \frac{\exp\left\{\mathbb{E}_{i\neq j}\left[\ln p(\mathcal{D},\theta)\right]\right\}}{\int \exp\left\{\mathbb{E}_{i\neq j}\left[\ln p(\mathcal{D},\theta)\right]\right\}d\theta_j} \tag{21}
$$

while the optimal distribution $q_\phi^*(\theta)$ can be approximated by updating each factorized distribution $q_{\phi_i}^*(\theta_i)$ iteratively, which is referred to as coordinate ascent algorithm [152]. In addition to coordinate ascent, another method for estimating the ELBO is stochastic VI based on stochastic gradient descent optimization [153]. Some remarkable extensions to stochastic VI include the natural gradient [153], the Bayes by Backprop [154] and the local reparameterization trick [155]. Conventionally, these VI approaches employ the mean-field assumption with factorized variational distribution, which leads to a straightforward lower bound for optimization, but the approximation capability is limited due to the ignorance of posterior correlations among variational parameters. To address this issue, some approaches adopt a Gaussian distributions with full covariance matrix [156, 157] as the variational distribution, but incorporating correlations inevitably introduces substantial memory and computational cost due to the increase of tunable parameters.

In summary, VI has become one of the most prevalent methods to infer the posterior distribution in Bayesian methods for UQ in ML due to its flexibility, scalability, and computational efficiency. However, due to the simplifications introduced to facilitate the optimization process, VI inevitably encounters some practical issues. For example, the variational distribution is often chosen to be a Gaussian or another unimodal distribution for computational efficiency, but such choices can limit the ability of VI to approximate multimodal posterior distributions, resulting in capturing only a single mode [77]. In contrast, using a multimodal variational distribution, such as a GMM, can alleviate this issue but typically incurs a significantly higher computational cost. Consequently, it is essential to balance the approximation capability and computational cost for VI, as the amount of variational parameters scales quadratically with the number of parameters in the model [17]. Conditioned on the variational approximation $q_\phi(\theta)$, the predictive distribution can be estimated using MC integration:

$$p(y|x,\mathcal{D}) \approx \int p(y|x,\theta) q_\phi(\theta) d\theta \approx \frac{1}{N}\sum_{i=1}^{N} p(y|x,\theta_i)\Big|_{\theta_i \sim q_\phi(\theta)} \tag{22}$$

**(3)** ***MCMC sampling***

MCMC is a general method for sampling from an intractable distribution. Specifically, a Markov chain is a stochastic process that models a sequence of samples $\theta^{(1)},\theta^{(2)},...,\theta^{(T)}$ with each state connected to the previous state by a transition probability $T_t\left(\theta^{(t)},\theta^{(t+1)}\right) \equiv p\left(\theta^{(t+1)}\middle|\theta^{(t)}\right)$. If $T_t\left(\theta^{(t)},\theta^{(t+1)}\right)$ is same for all $t$, the Markov chain is called homogeneous. Another crucial element for a Markov chain is the invariant, or stationary, distribution, which is defined as $p\left(\theta^{(t+1)}\right)=\sum_{t} T_t\left(\theta^{(t)},\theta^{(t+1)}\right)p\left(\theta^{(t)}\right)$ for a homogeneous Markov chain with transition probability $T\left(\theta^{(t)},\theta^{(t+1)}\right)$. A sufficient (but not necessary) condition to ensure a distribution $p(\theta)$ is invariant is that the transition probability satisfies the detailed balance, namely $p\left(\theta^{(t+1)}\right)T_t\left(\theta^{(t+1)},\theta^{(t)}\right)=p\left(\theta^{(t)}\right)T_t\left(\theta^{(t)},\theta^{(t+1)}\right)$. For MCMC methods, the goal is to use Markov chains to sample from a given distribution, which can be achieved by constructing a Markov chain to make the desired distribution invariant. In addition, the property of ergodicity must also be ensured, which is defined as that for $t\rightarrow\infty$, the distribution $p\left(z^{(t)}\right)$ converges to the invariant distribution $p(\theta)$, and $p(\theta)$ is called the equilibrium distribution under this condition. Then, the samples from the desired distribution can be obtained by recording states from the chain [77]. Based on this technique, the predictive distribution can be estimated using samples of model parameters drawn from the posterior distribution, as illustrated in Eq. (17).

MCMC sampling ensures that samples converge to the exact posterior distribution after a sufficient number of iterations, but determining this sufficiency poses a challenge. As a result, MCMC sampling typically demands a considerable number of samples to guarantee convergence, resulting in an excessive investment of time and computing resources, thereby limiting its application to small-scale problems [17]. Over the years, a variety of MCMC methods, such as Metropolis-Hastings sampling [158], Gibbs sampling [159], slice sampling [160], Sequential Monte Carlo sampling [161], and Hybrid Monte Carlo (also known as Hamiltonian Monte Carlo) sampling [162], have been developed to improve sampling efficiency and reduce convergence iterations. Another potential direction involves combining MCMC sampling with VI to take advantage of both methods [163]. Nevertheless, the issue of trade-off between sampling accuracy and computation cost remains a challenge for the application of MCMC sampling in UQ for ML models and requires further investigation.

In summary, most practical Bayesian ML approaches estimate the posterior distribution of model parameters using the aforementioned three methods, and then formulate the predictive distribution to provide a systematic framework for uncertainty estimate and quantification. In this section, some well-known and commonly used Bayesian methods for UT in ML will be elaborately reviewed. Conditioned on the dominance of NNs in the modern ML literature, the remaining part of this section is divided into two subsections: the first covers several prevalent Bayesian ML methods that are not based on NN architecture, while the second focuses on shallow and deep BNNs with different network architecture, as well as the application of BNNs in several state-of-the-art (SOTA) learning techniques.

### *4.2 Non-NN-based probabilistic ML within a Bayesian framework*

This section provides a comprehensive introduction to several Bayesian ML methods that, while not rooted in NNs, are crucial for interpreting and estimating predictive uncertainty and widely used in structural dynamics, which includes Dirichlet process mixture models, Bayesian networks, relevance vector machines, and Gaussian processes.

#### *4.2.1 Dirichlet process mixture models for UT in clustering*

As discussed in Section 3.2, mixture models are powerful tools for UT, but a key limitation lies in the difficulty of determining the optimal number of components, reflecting their weakness in addressing epistemic uncertainty. Dirichlet process mixture models (DPMMs) are developed to address this issue, which can be viewed as a Bayesian extension of conventional mixture models [164]. Specifically, DPMMs employ a Dirichlet process (DP) prior that encompasses both the component number and the parameters of each component, and the posteriors can then be derived according to the Bayes' theorem through MCMC sampling [95] or VI [96, 152]. A DP is a stochastic process whose sample paths are probability measures with probability one. Each sample drawn from a DP is a random distribution whose marginal distributions are Dirichlet distributions [164]. For a random distribution $G$, if for any finite measurable partition $\mathrm{A}=\{A_1,A_2,...,A_r\}$ of the domain $\mathbf{\Theta}$, the following equation holds:

$$\left(G(A_1),G(A_2),...,G(A_r)\right)\sim\mathrm{Dir}\left(\alpha G_0(A_1),\alpha G_0(A_2),...,\alpha G_0(A_r)\right) \tag{23}$$

Then, $G$ follows a DP, denoted by $G\sim\mathrm{DP}(\alpha,G_0)$, where $G_0$ is a random distribution defined over $\mathbf{\Theta}$ known as the base distribution; $\alpha$ is a positive number referred to as the concentration parameter. As $G$ is a distribution itself, a set of i.i.d. samples $\{\theta_1,\theta_2,...,\theta_N\}$, referred to as atoms, can be drawn from $G$. Conditioned on these samples, the DPMM treats $\theta_n$ as the parameters for the distribution of the observation $x_n$, i.e., $x_n\sim F\left(x_n|\theta_n\right)$, while the distinct values $\{\theta_1^*,\theta_2^*,...,\theta_K^*\}$ among these samples naturally induce a partition, with each $\theta_k^*$ corresponding to an independent component in the mixture model.

In practice, mixture components are usually modeled using Gaussian distributions, with $\theta_k^*=\{\mu_k,\mathbf{\Lambda}_k\}$ denoting the mean vector and precision matrix, leading to the formulation of a Dirichlet process Gaussian mixture model (DPGMM). The generative process of the DPGMM can be expressed using the stick-breaking construction outlined as follows [152, 165]:

$$\begin{gathered} v_k\sim\mathrm{Beta}(1,\alpha);\ \pi_k(v)=v_k\prod_{i=1}^{K-1}(1-v_i),\ K\to\infty \\ \theta_k^*=\{\mu_k,\mathbf{\Lambda}_k\}\sim G_0;\ z_n\sim\mathrm{Mult}\left(\pi_k(v)\right);\ x_n\sim\mathcal{N}\left(x\mid z_n,\theta_{z_n}^*\right) \end{gathered} \tag{24}$$

where $\pi_k$ is the mixing proportion and $z_n$ is the latent variable denoting cluster assignment. Therefore, given a set of i.i.d. samples $\mathrm{X}=\{x_1,x_2,...,x_N\}$, the likelihood function can be

expressed as [152]:

$$p(\mathrm{X}|\mathbf{\Phi})=\prod_{n=1}^{N}p(x_n|\mathbf{\Phi})=\prod_{n=1}^{N}\prod_{k=1}^{K}\left[\mathcal{N}(x_n\mid\mu_k,\mathbf{\Lambda}_k)\right]^{\mathbf{1}[z_n=k]},\ K\to\infty \tag{25}$$

where $\mathbf{1}[z_n=k]$ is an indicator function that equals 1 if $z_n=k$ and 0 otherwise; $\mathbf{\Phi}=\{v,z,\mu,\mathbf{\Lambda}\}$ denotes the parameters of the DPGMM. The posterior distribution of $\mathbf{\Phi}$ can be inferred accordingly using Bayes' theorem:

$$p(\mathbf{\Phi}|\mathrm{X})=\frac{p(\mathbf{\Phi})p(\mathrm{X}|\mathbf{\Phi})}{\int p(\mathbf{\Phi})p(\mathrm{X}|\mathbf{\Phi})d\mathbf{\Phi}}=\frac{\prod_{k=1}^{K}p(v_k)p(\mu_k,\mathbf{\Lambda}_k)\prod_{n=1}^{N}p(z_n\mid v)p(x_n\mid z_n,\mu,\mathbf{\Lambda})}{p(\mathrm{X})} \tag{26}$$

To estimate the posterior distribution, various methods have been developed, including collapsed Gibbs sampling [95], blocked Gibbs sampling [152], truncated VI [152], stochastic VI [153], streaming truncation-free VI [96]. Based on the inferred posterior, the predictive distribution of the DPGMM can be formulated for uncertainty estimation, as illustrated in Section 4.1. Compared to traditional mixture models, DPMMs explicitly address epistemic uncertainty stemming from model complexity and model parameters by introducing a nonparametric prior, while aleatoric uncertainty is captured through the soft assignment strategy. Consequently, DPMMs offer a systematic framework for elucidating and quantifying predictive uncertainty in clustering, resulting in greater robustness.

#### *4.2.2 Bayesian networks for UT in classification and regression*

Bayesian networks (BNs), also known as Bayes nets or belief networks, are probabilistic graphical models constructed by directed acyclic graphs (DAGs) [29], where the nodes represent uncertain variables and the directed links from parent nodes to child nodes indicate the dependence between variables, as shown in Fig. 9. Generally, BNs model discrete variables through probability mass functions (PMFs) over intervals [104, 166] to estimate the associated uncertainty. Assume $\mathrm{X}=\{x_1,x_2,...,x_K\}$ denote the uncertain variables in a BN, the joint PMF of the BN in Fig. 9 can be expressed as:

$$p(\mathrm{X})=\prod_{k=1}^{K} p(x_k|pa_k)=p(x_5|x_2,x_3)p(x_4|x_2)p(x_3|x_1)p(x_2|x_1)p(x_1) \tag{27}$$

where $pa_k$ denotes the set of parents of $x_k$. Conditioned on the joint distribution, BN inference algorithms propagate through all nodes to compute the conditional probabilities using Bayes' theorem when provided with evidence $e$ on one or several variables [104]:

$$p(\mathrm{X}^{-i}|x_i=e)=\frac{p(\mathrm{X}^{-i})p(x_i=e|\mathrm{X}^{-i})}{p(x_i=e)} \tag{28}$$

where $\mathrm{X}^{-i}$ denotes the set of all variables except $x_i$. Therefore, BNs provide a coherent framework to represent and reason about uncertainty propagation by explicitly encoding conditional dependencies within a network structure [167]. They also enable more efficient Bayesian updating results across all nodes compared to many other Bayesian methods [104]. However, constructing an accurate and representative BN for a given system can be challenging, as it typically requires expert knowledge to define the conditional probabilities in the BN. For more details about the BN and its variants, one can refer to [29, 39].

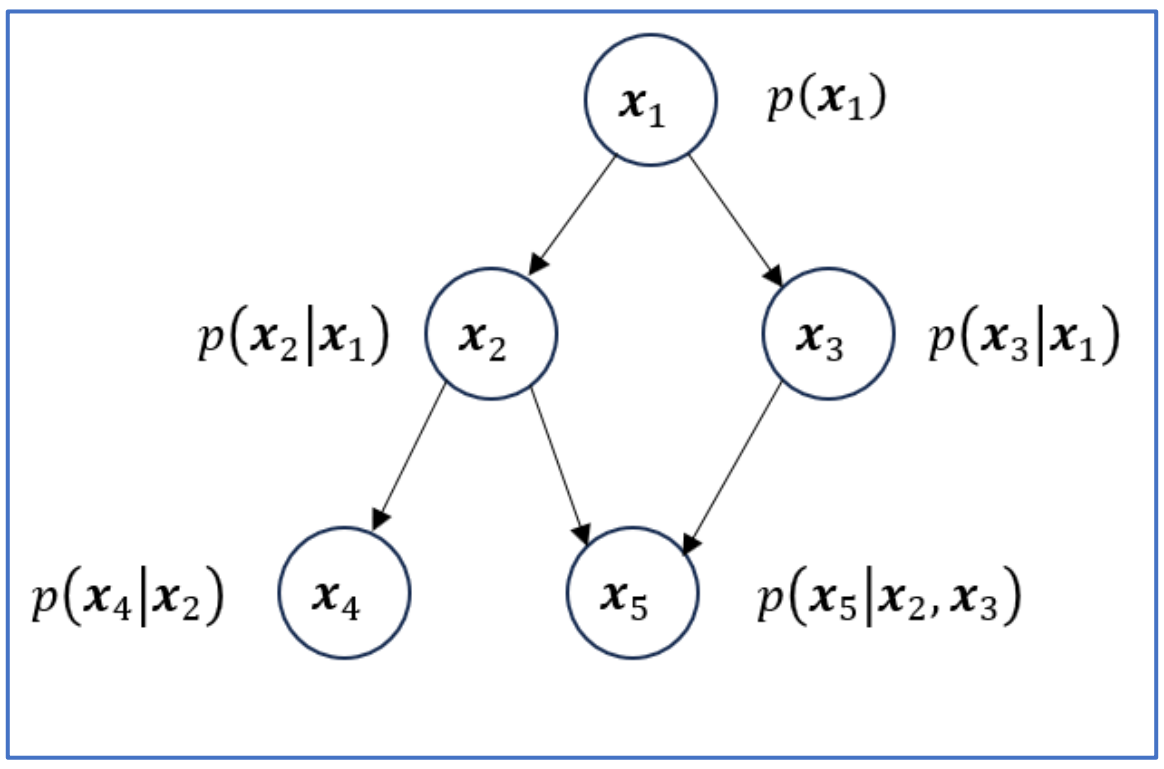

Figure 9. The schematic diagram of a Bayesian network.

#### *4.2.3 Relevance vector machines for UT in classification and regression*

In ML, raw data is typically transformed into feature vectors to train the model. However, it can be difficult to represent certain types of raw data, such as text documents, with fixed-sized feature vectors. An alternative approach to address this issue is using a kernel function, denoted by $\kappa(x,x')=\phi(x)^T\phi(x')$, to implicitly operate data in a high-dimensional feature space [29]. Over the years, various kernel methods have been developed for different ML tasks, with support vector machines (SVMs) becoming among the most widely used. SVMs aim to find a hyperplane or set of hyperplanes in a high-dimensional feature space that best generalizes the training dataset, making them applicable for both classification and regression [29]. However, despite their efficiency and effectiveness, SVMs typically lack mechanisms for UT due to their deterministic nature [29, 39]. To address this issue, relevance vector machines (RVMs) have been developed as a Bayesian extension of SVMs to address uncertainties [77].

Take a regression task for example, the RVM introduces a kernel function $\kappa(x,x_n)$ into the linear regression model to capture nonlinearities:

$$f(x)=w^T\kappa(x,x_n) \tag{29}$$

where $w$ is the weight vector. To account for aleatoric uncertainty, a noise term $\varepsilon\sim\mathcal{N}(0,\sigma^2\boldsymbol{I})$ is introduced to formulate a Gaussian distribution for the target output, as illustrated in Section 2.3.1, i.e., $p(y|x,\sigma^2)=\mathcal{N}(y|f(x),\sigma^2\boldsymbol{I})$. Meanwhile, as a Bayesian approach, the RVM addresses epistemic uncertainty in model parameters by introducing an independent zero-mean Gaussian prior over the weights $w$ [77], i.e., $p(w|\alpha)=\prod_{i=1}^{M}\mathcal{N}(w_i|0,\alpha_i^{-1})$. Based on this prior, given a set of training data $\mathcal{D}_{tr}=\{x_i,y_i\}_{i=1}^{N}$, the posterior of $w$ can be analytically derived using Bayes' rule, which is also a Gaussian distribution given by:

$$p\left(w \middle| \mathcal{D}_{tr}, \alpha, \sigma^2\right) = \frac{p\left(w \middle| \alpha\right) p\left(y \middle| x, w, \sigma^2\right)}{p\left(y \middle| x, \alpha, \sigma^2\right)} = \mathcal{N}\left(w \middle| m, \mathbf{\Sigma}\right) \tag{30}$$

where $m$ and $\mathbf{\Sigma}$ can be calculated given the hyperparameters $\alpha$ and $\sigma^2$, and these hyperparameters are commonly estimated by maximizing the evidence through derivatives or the EM algorithm [77]. Based on the posterior distribution, the predictive distribution of the RVM can be derived using Eq. (17), with its variance serving as a measure of predictive uncertainty. Similarly, RVMs can also be applied to classification by replacing real-valued targets to class labels and modeling predictions using a Bernoulli or Categorical distribution [29]. The only difference from regression is that the posterior cannot be derived analytically in classification, requiring the use of approximation methods. Fig. 10 schematically illustrates the UT of RVMs in regression and classification, which explicitly quantify the predictive uncertainty through probabilistic decision boundary or predictive variance for classification and regression tasks, respectively. For more details about RVMs, one can refer to [168].

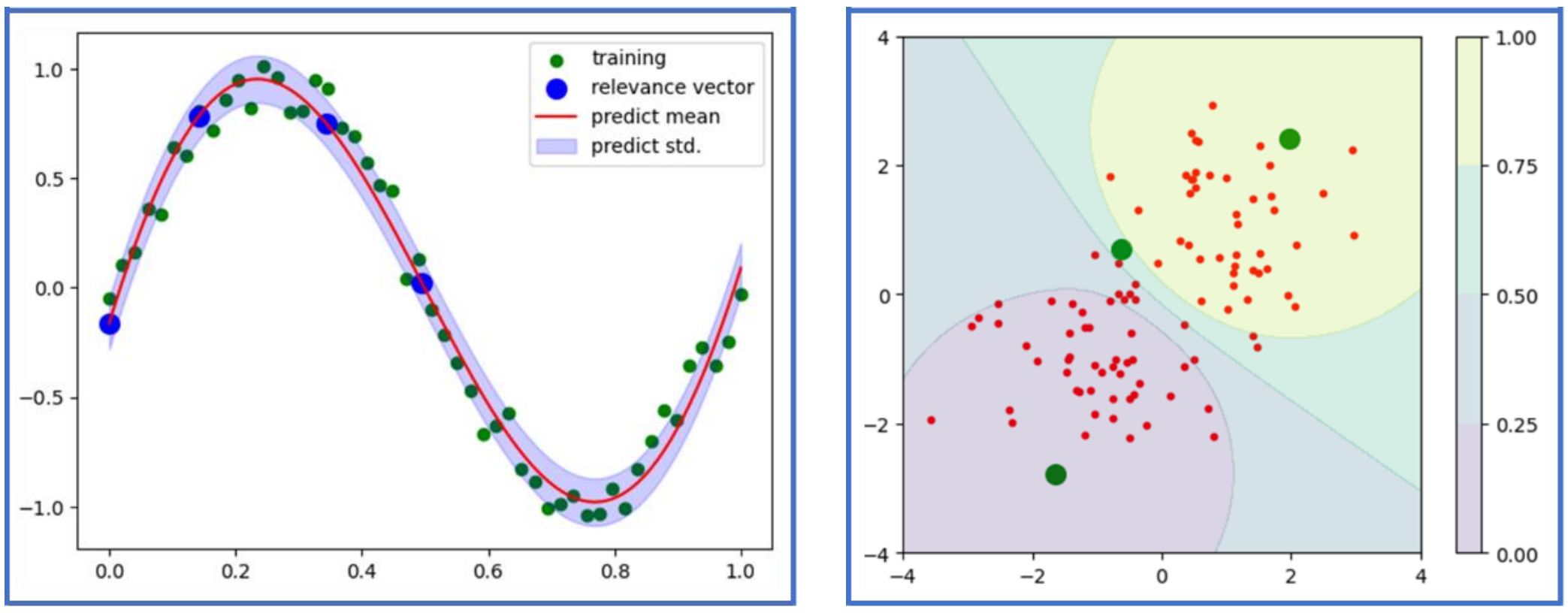

(a) (b)

Figure 10. Uncertainty treatment of RVMs for (a) regression and (b) classification.

#### *4.2.4 Gaussian processes for UT in classification and regression*

##### *4.2.4.1 Gaussian processes*

The Gaussian process (GP) is a nonparametric Bayesian kernel method that performs

Bayesian inference over the mapping functions $f$ from the inputs $x$ to target variables $y$ directly, rather than inferring parametric representations for these functions [29, 169]. Specifically, a function $f(x)$ distributed as a GP with the mean function $m(\bullet)$ and covariance function $\kappa(\bullet)$ can be denoted by:

$$f(x) \sim GP(m(x), \kappa(x, x')) \tag{31}$$

where $m(x) = \mathbb{E}[f(x)]$ and $\kappa(x, x') = \mathbb{E}\left[(f(x) - m(x))(f(x') - m(x))^T\right]$ is a positive definite kernel function. For any finite set of data points $\mathrm{X} = \{x_1, x_2, ..., x_N\}$, this process defines a joint Gaussian distribution, namely $p(f|\mathrm{X}) = \mathcal{N}(f|\mu(\mathrm{X}), \mathbf{\Sigma}(\mathrm{X}))$, with mean and covariance given by $\mu(\mathrm{X}) = \{m(x_1), m(x_2), ..., m(x_N)\}$ and $\mathbf{\Sigma}_{ij}(\mathrm{X}) = \kappa(x_i, x_j)$, which demonstrates that each sample of a GP is itself a distribution, and every finite collection of these samples has a multivariate Gaussian distribution [29]. Due to their nonparametric nature, GPs are highly flexible in modeling data with complex structure and correlations and can be used for both regression and classification.

Take a regression task for example, where a GP is used as the prior of the regression function. Suppose the training set is denoted by $\mathcal{D}_{tr} = \{\mathrm{X}_{tr}, \mathrm{Y}_{tr}\} = \{x_i, y_i\}_{i=1}^{N}$ with the targets governed by some mapping function and some independent Gaussian noise, namely $y = f(x) + \epsilon, \epsilon \sim \mathcal{N}(0, \sigma^2)$, the aim is to predict the target $\mathrm{Y}$ given new observations $\mathrm{X}$. Based on the definition of GP, it has:

$$\begin{pmatrix} f(\mathrm{X}_{tr}) \\ f(\mathrm{X}) \end{pmatrix} \sim \mathcal{N}\left(\begin{pmatrix} \mu(\mathrm{X}_{tr}) \\ \mu(\mathrm{X}) \end{pmatrix}, \begin{pmatrix} \mathbf{\Sigma}_{tr} & \mathbf{\Sigma}' \\ (\mathbf{\Sigma}')^T & \mathbf{\Sigma} \end{pmatrix}\right) \tag{32}$$

where $\mathbf{\Sigma}_{tr} = \kappa(\mathrm{X}_{tr}, \mathrm{X}_{tr})$, $\mathbf{\Sigma}' = \kappa(\mathrm{X}_{tr}, \mathrm{X})$, and $\mathbf{\Sigma} = \kappa(\mathrm{X}, \mathrm{X})$. For computational simplicity, it is common to assume the mean is zero and use the squared exponential (SE) kernel, aka Gaussian kernel or radial basis function (RBF) kernel [170], that is, $\begin{pmatrix} \mu(\mathrm{X}_{tr}) \\ \mu(\mathrm{X}) \end{pmatrix} = \mathbf{0}$ and $\kappa(x,x') = \sigma_f^2 \exp\left\{-\frac{1}{2l^2}|x-x'|^2\right\}$ . As a result, Eq. (32) can be rewritten as $\begin{pmatrix} f(\mathrm{X}_{tr}) \\ f(\mathrm{X}) \end{pmatrix} \sim \mathcal{N}\left(\mathbf{0}, \begin{pmatrix} \mathbf{\Sigma}_{tr} & \mathbf{\Sigma}' \\ (\mathbf{\Sigma}')^T & \mathbf{\Sigma} \end{pmatrix}\right)$. Since the model prediction $\mathrm{Y}$ can be represented by the sum of $f(\mathrm{X})$ and the noise $\epsilon$, the joint distribution of the model prediction and the targets in the training data is also a Gaussian, namely $\begin{pmatrix} \mathrm{Y}_{tr} \\ \mathrm{Y} \end{pmatrix} = \begin{pmatrix} f(\mathrm{X}_{tr}) \\ f(\mathrm{X}) \end{pmatrix} + \begin{pmatrix} \epsilon \\ \epsilon \end{pmatrix} \sim \mathcal{N}\left(\mathbf{0}, \begin{pmatrix} \mathbf{\Sigma}_{tr} + \sigma^2 I & \mathbf{\Sigma}' \\ (\mathbf{\Sigma}')^T & \mathbf{\Sigma} + \sigma^2 I \end{pmatrix}\right)$. Therefore, the predictive distribution can be analytically derived by marginalizing out the function $f$ [29]:

$$p(\mathrm{Y}|\mathrm{X}, \mathcal{D}_{tr}) = \mathcal{N}\left(f(\mathrm{X})|\mu^*, \mathbf{\Sigma}^*\right) \tag{33}$$

where $\mu^*$ and $\mathbf{\Sigma}^*$ can be calculated given the training data with the detailed derivation provided in [29]. For classification tasks, GPs can be employed in a similar manner to regression tasks by replacing the multinomial logit function with the multinomial probit function [29]. A schematic illustration of GP for one-dimensional regression is presented in Fig. 11, where the variance quantifies the predictive uncertainty.

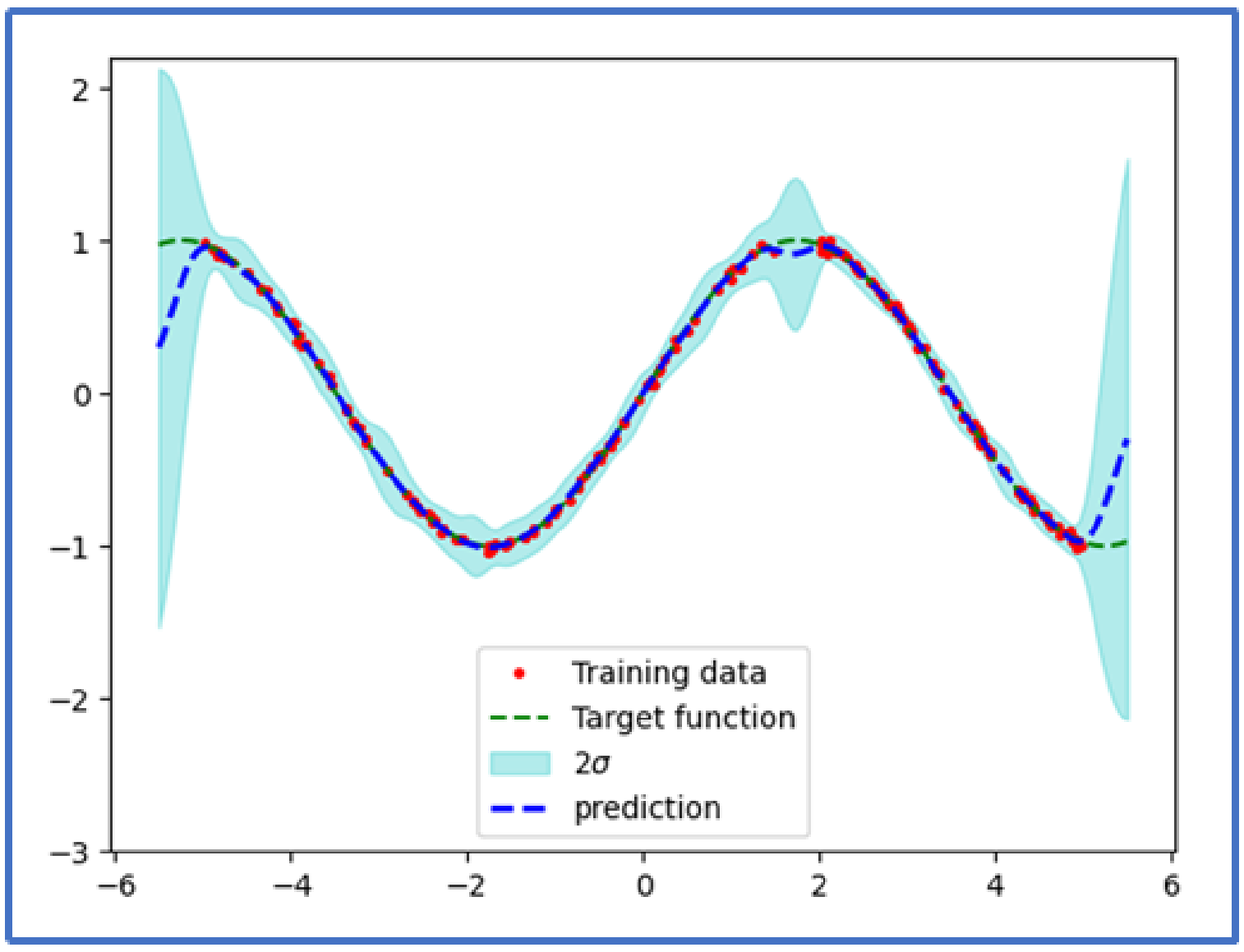

Figure 11. Schematic diagram of a GP for UT in regression.

Compared to RVM, the nonparametric nature of GP provides it with greater flexibility but incurs substantial computational and storage costs due to the dense nature of the kernel matrix [77]. Nevertheless, GPs remain a preferred Bayesian approach for UT, as their closed-form predictive distribution significantly simplifies the formulation of predictive distributions and the quantification of predictive uncertainty compared to many other Bayesian methods. However, a limitation of GP is its sensitivity to the choice of hyperparameters, such as the noise variance and the kernel parameters, which motivates researchers to optimize hyperparameters using Bayesian methods by treating them as uncertain variables [29, 108, 169]. Another approach to addressing this issue, which has been applied to structural dynamics, is the use of Bayesian (or partially Bayesian) active learning strategies to facilitate hyperparameter selection and accelerate posterior inference for GPs [117-120], which is referred to as Bayesian active learning-augmented Gaussian process (BAL-GP) in this work.

#### *4.2.4.2 Bayesian active learning-augmented Gaussian processes*

In the context of ML, active learning refers to strategies for selecting the most informative samples to annotate, with the goal of training an ML model to achieve maximum performance, which is typically achieved by optimizing an acquisition function or learning function [171]. Bayesian active learning is a branch of active learning that uses Bayesian inference to estimate and quantify an ML model's predictive uncertainty, which is then used to construct learning functions [171]. Typical learning functions in Bayesian active learning include predictive entropy and mutual information between model predictions and the posterior distribution of model parameters, mean standard deviation, and others [172]. In structural dynamics, Bayesian active learning has been applied to select informative simulated samples for updating surrogate models (typically Gaussian process regressors) in tasks such as structural model updating and reliability analysis [118, 173], aiming to facilitate hyperparameter selection and accelerate posterior inference for GPs. Accordingly, BAL-GP is introduced in this work as a variant of conventional GPs. A pioneering study on BAL-GP in the field of structural dynamics was proposed by Dang et al. [120], in which a GP was adopted to model the performance function of the investigated physical system, and its posterior mean and variance were used to estimate the expected failure probability of the system and the associated variance (uncertainty), respectively. To retain the performance of the GP while minimizing the computational cost of posterior inference, an active learning technique, which selects the most informative samples to update the GP model based on each sample's contribution to the upper bound of the posterior variance of the failure probability, was proposed, aiming to reduce the epistemic uncertainty of the failure probability. This method illustrates how uncertainty estimates from ML models can be effectively leveraged to guide structural design and reliability analysis, thus has been followed and extended in many subsequent studies [117-119].

#### *4.2.4.3 Deep Gaussian processes*

The deep Gaussian process (DGP) is a hierarchical extension of conventional GPs, which has a multiple-layer structure akin to DNNs with each layer represented by a GP [174], as depicted in Fig. 12. From this perspective, DGPs can also be viewed as deep belief networks based on GP mappings [175]. Compared to traditional GPs, DGPs overcome the limitation arising from the absence of kernel functions capable of handling structured data, which is achieved through a hierarchical feature extraction scheme that properly determines the similarity of a pair of data points based on the multilayer structure [174]. Consequently, DGPs exhibit enhanced capability in capturing complex relationships within large datasets and are less sensitive to the choice of kernels compared to traditional GPs [17]. Similar to NNs, the forward propagation and predictive distribution of the DGP in Fig. 12 can be expressed as:

$$y = f_n\left(f_{n-1}\left(\ldots f_1\left(x\right)\right)\right) \tag{34}$$

$$p\left(y, f \middle| x\right) = p\left(y \middle| f_n\right) \prod_{i=2}^{n} p\left(f_i \middle| f_{i-1}\right) p\left(f_1 \middle| x\right) \tag{35}$$

where each function $f_i(\bullet)$ is a GP model. The joint distribution of each hidden layer follows a Gaussian distribution due to the definition of GP, while the predictive distribution captures a more complex mapping from the input to the target variable through the composition of GPs. This composition also allows uncertainty to propagate from the input through each hidden layer to the output. However, despite the advantages of DGPs in nonlinear mapping and UT, a challenge arises from maximizing the data likelihood $p(y|x)$ in the training phase of DGPs, as the direct marginalization of hidden variables $f$ is intractable, which makes the training of DGPs commonly rely on VI by introducing inducing points in each hidden layer and optimizing the variational distribution [175].

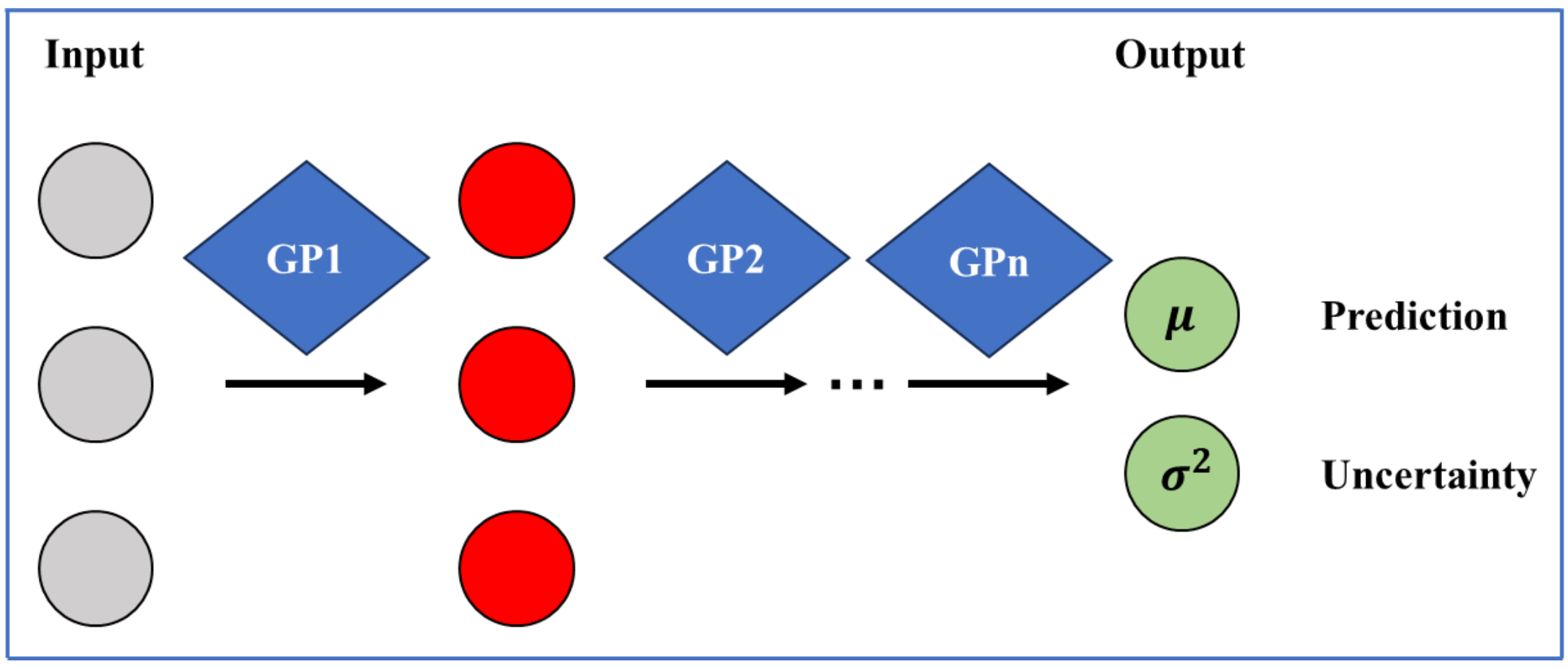

Figure 12. The schematic diagram of a deep Gaussian process.

### *4.3 NN-based ML working in a Bayesian framework*

In the context of UT for ML models, BNNs play a pivotal role by combining the advantages of NNs in representation learning and nonlinear mapping with those of Bayesian methods in addressing uncertainties. BNNs treat the parameters of NNs (weights and biases, or weights alone for efficiency) as uncertain variables with specific priors, as illustrated in Fig. 13. Subsequently, the posterior distribution of network parameters can be inferred using Bayes' rule to formulate the predictive distribution of NNs. Consequently, BNNs provide a systematic framework for UT in NNs, especially addressing epistemic uncertainty arising from model parameters, which makes them more robust against overfitting and mitigates the risk of overconfident predictions [17, 169]. Considering their unique advantages, this work places a significant emphasis on BNNs, with prevalent inference techniques in modern BNNs highlighted and their integration with modern DNN architectures exemplified.

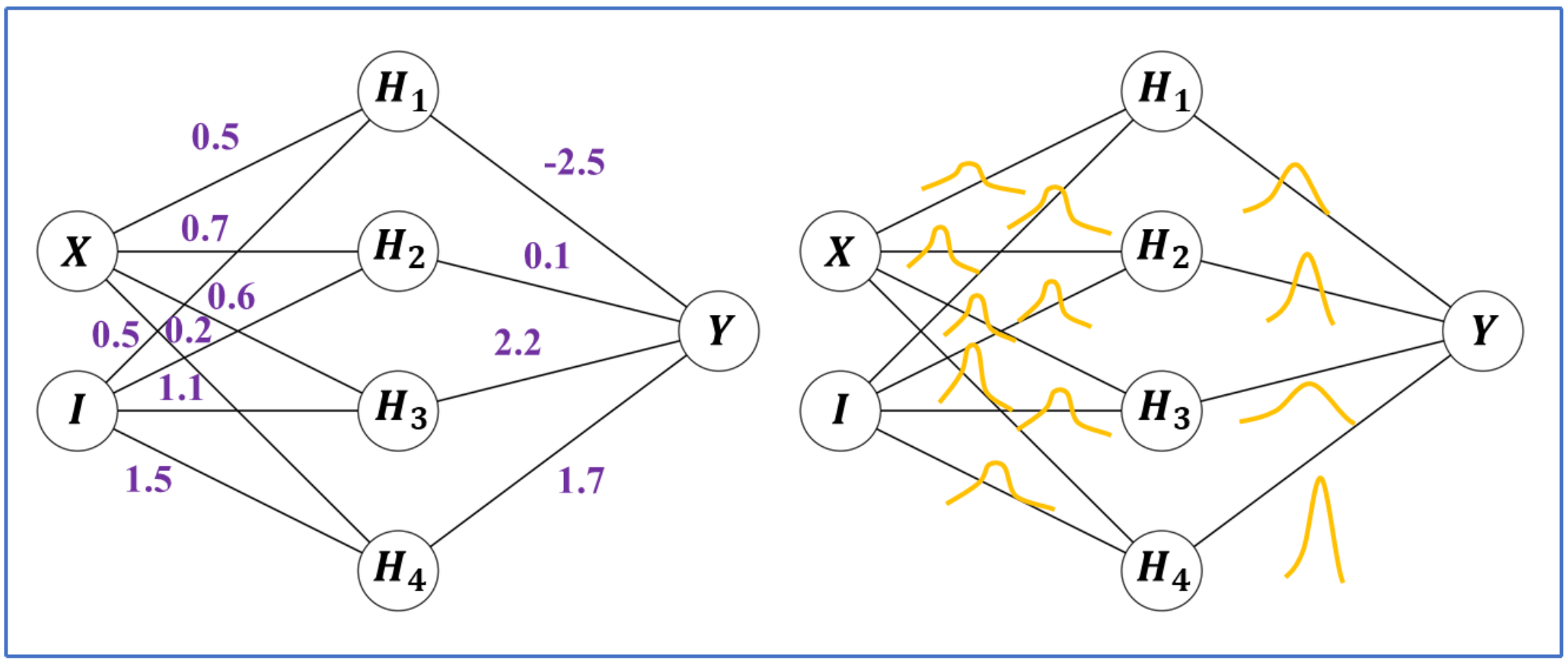

Figure 13. The difference between BNNs and traditional NNs. Left: a traditional NN with each weight being a fixed value. Right: a BNN with each weight assigned a probability distribution over possible values (reproduced from [154]).

#### *4.3.1 Fundamentals*

Similar to other Bayesian methods, the computational issue of inferring the posterior distribution also troubles BNNs, which is even worse as modern DNNs typically have a large number of parameters. Over the years, with the increasing popularity of BNNs, various methods to estimate the posterior distribution of model parameters for BNNs have been developed, among which the most notable include Laplace approximation [176], Hamiltonian Monte Carlo sampling [177], VI with full covariance matrices [156]. These approaches represent crucial initial strides towards practical BNNs, but they often suffer from prohibitive computational costs or reduced predictive performance due to simplifications, limiting their scalability to complex modern DNN architectures. To this end, a recent trend in approximation methods turns to be combining MC sampling and VI, aiming at utilizing their respective advantages to provide practical solutions to complex BNN models with satisfying predictive performance, among which the most prominent approaches include Bayes by Backprop [154] and MC dropout [33] and thus are introduced in this work.

***(1) Bayes by Backprop*:**

Bayes by Backprop is a training technique for BNNs proposed by Blundell et al. [154], which adopts a Gaussian prior over the weights $w$ and derives the posterior distribution by variational inference. Specifically, the loss function for a BNN from the VI perspective can be expressed as:

$$\mathcal{L}(\mathcal{D}_{tr},\phi)=-\mathbb{E}_{q_\phi(w)}\left[\log p\left(\mathcal{D}_{tr}|w\right)\right]+D_{KL}\left[q_\phi(w)\|p(w)\right] \tag{36}$$

The loss function is a sum of a data-dependent part $-\mathbb{E}_{q_\phi(w)}\left[\log p\left(\mathcal{D}_{tr}|w\right)\right]$ known as the likelihood loss, and a prior dependent part $D_{KL}\left[q_\phi(w)\|p(w)\right]$ known as the complexity loss. Minimizing Eq. (36) directly is computationally prohibitive, thus in Bayes by Backprop, an unbiased gradient estimate of the loss is derived based on the reparameterization trick [155].

Using MC integration, Eq. (36) can be approximated as:

$$\mathcal{L}(\mathcal{D}_{tr},\phi) \approx -\sum_{i=1}^{N} \log q_{\phi}\left(w^{(i)}\right) - \log p\left(w^{(i)}\right) - \log p\left(\mathcal{D}_{tr} \middle| w^{(i)}\right) \tag{37}$$

where $w^{(i)}$ is the $i$ th MC sample drawn from the variational posterior $q_{\phi}\left(w^{(i)}\right)$. Suppose the variational posterior is a diagonal Gaussian, then a sample of the weights $w^{(i)}$ can be obtained by employing the reparameterization trick, which can be expressed as $w = \mu + \sigma \odot \epsilon$ with $\epsilon \sim \mathcal{N}(0, I)$. Based on this construction, the gradients of the loss function in Eq. (37) with respect to the mean and variance can be analytically derived, which enables the model to be trained by the usual backpropagation algorithm. Additionally, in [154], the BNN model is further changed by assuming a mixture of Gaussians prior over each weight, liberating the algorithm from the confines of Gaussian priors and posteriors to further improve its flexibility. Case studies on some benchmark datasets demonstrate that Bayes by Backprop yields comparable performance to some SOTA techniques with an enhanced capability in interpreting and quantifying predictive uncertainty for better generalization [154]. However, even Bayes by Backprop is computationally expensive sometimes despite of the tractability of the gradients, as the use of Gaussian variational posterior doubles the number of model parameters without increasing model capacity by much [28], which could limit its application to large complex DNNs. Moreover, this method is normally implemented in fully connected layers, making it incapable of handling parameter uncertainty for more sophisticated NN architectures containing convolutional and recurrent layers. To overcome these limitations, Gal and Ghahramani [33, 178] proposed a more efficient algorithm for training BNNs, known as MC dropout, which has currently become the most popular method for training BNNs in many domains [17] due to its scalability, computational simplicity and applicability.

***(2) Monte Carlo dropout*:**

The development of MC dropout is inspired by the theoretical findings that some stochastic regularization techniques, mainly dropout, are equivalent to the approximate inference in training BNNs when the prior satisfying the KL condition [28]. Specifically, dropout omits each hidden unit from the network with a certain probability (called dropout rate) on each training case [179], as shown in Fig. 14, which prevents excessive co-tuning and constrains model complexity, thus alleviates overfitting problems in NNs. Conditioned on a NN with $L$ layers, the loss function with dropout and L2 regularization can be expressed as:

$$\mathcal{L}_{dropout} = \frac{1}{N}\sum_{i=1}^{N} L(y_i, \hat{y}_i) + \lambda \sum_{i=1}^{L}\left(\|\mathrm{W}_i\|_2^2 + \|b_i\|_2^2\right) \tag{38}$$

where $L(y_i, \hat{y}_i)$ denotes the loss function; $\mathrm{W}_\mathrm{i}$ and $b_i$ are the weight matrix and bias vector of the $i$ th layer, respectively; and $\lambda$ is the weight decay. Dropout is applied by sampling binary variables from a Bernoulli distribution with the dropout rate $p_i$ for every input point and for every network unit in each layer (apart from the last one), which can be expressed as:

$$\mathrm{W}_i = \mathrm{M}_i \cdot \mathrm{diag}\left(\left[z_{ij}\right]_{j=1}^{K_i}\right),\ z_{ij} \sim \mathrm{Bernoulli}(p_i) \text{ for } i = 1,\ldots,L, j = 1,\ldots,K_{i-1} \tag{39}$$

Consider the optimization of a BNN using VI, the objective is minimizing the KL divergence between the true posterior over the weights $p(\mathrm{W}|\mathcal{D}_{tr})$ and a variational distribution $q_\phi(\mathrm{W})$, which can be expressed as $\mathcal{L}_{VI} = -\sum_{i=1}^{N}\int q_\phi(\mathrm{W})\log p(y_n|x_n,\mathrm{W})d\mathrm{W} + D_{KL}\left(q_\phi(\mathrm{W})\|p(\mathrm{W})\right)$. According to Gal and Ghahramani [33], this objective is equivalent to that shown in Eq. (38) given appropriate variational parameters and hyperparameters, which demonstrates that the standard training techniques for NNs with dropout can be directly implemented to approximate a BNN with VI, without introducing additional computational burden. The predictive distribution can be estimated with MC integration by applying dropout during the testing phase, as illustrated in Eq. (40), thus this method is referred to as MC dropout:

$$p\left(y \mid x, \mathcal{D}_{tr}\right) = \int p\left(y \mid x, \mathrm{W}\right) q_{\phi}\left(\mathrm{W}\right) d\mathrm{W} \approx \frac{1}{N} \sum_{i=1}^{N} p\left(y \mid x, \hat{\mathrm{W}}_i\right) \Big|_{\hat{\mathrm{W}}_i \sim q_{\phi}(\mathrm{W})} \tag{40}$$

Compared to other methods, MC dropout significantly alleviates the computational burden that often prohibits the application of BNNs. It also captures the correlations among network weights by factorizing the distribution for each weight row in each weight matrix, instead of factorizing over each weight scalar. Additionally, some variants of MC dropout can be implemented in CNNs and RNNs for UT [28]. Furthermore, MC dropout-based BNNs can also differentiate the aleatoric and epistemic parts of the predictive uncertainty [19]. These characteristics enable MC dropout to be integrated into large scale DNNs for UQ and performance improvement, making it become the most prevalent method for training BNNs currently [17]. Despite its efficiency and flexibility, MC dropout also has some limitations [28]. Firstly, in MC dropout, dropout is applied in both training and testing phases (unlike traditional dropout, which is used only during training), leading to additional computational costs in the testing phase to calculate sample mean and variance. Secondly, the uncertainty estimated by MC dropout may be less calibrated compared to other methods [28, 180], resulting in increased predictive uncertainty for large magnitude data points or variations in scale across datasets. Thirdly, MC dropout tends to underestimate predictive uncertainty due to the mean-field assumption and Gaussian variational distribution that are commonly used. In this scenario, VI tends to penalize the variational distribution more for placing probability mass where the true posterior has no mass, but less for missing areas where the true posterior does have mass [28]. This can lead to a narrower approximation of the posterior distribution, resulting in lower variance estimates. With these constraints, in practice MC dropout is often combined with some other programming techniques such as distributed computing for a better model performance and uncertainty estimate.

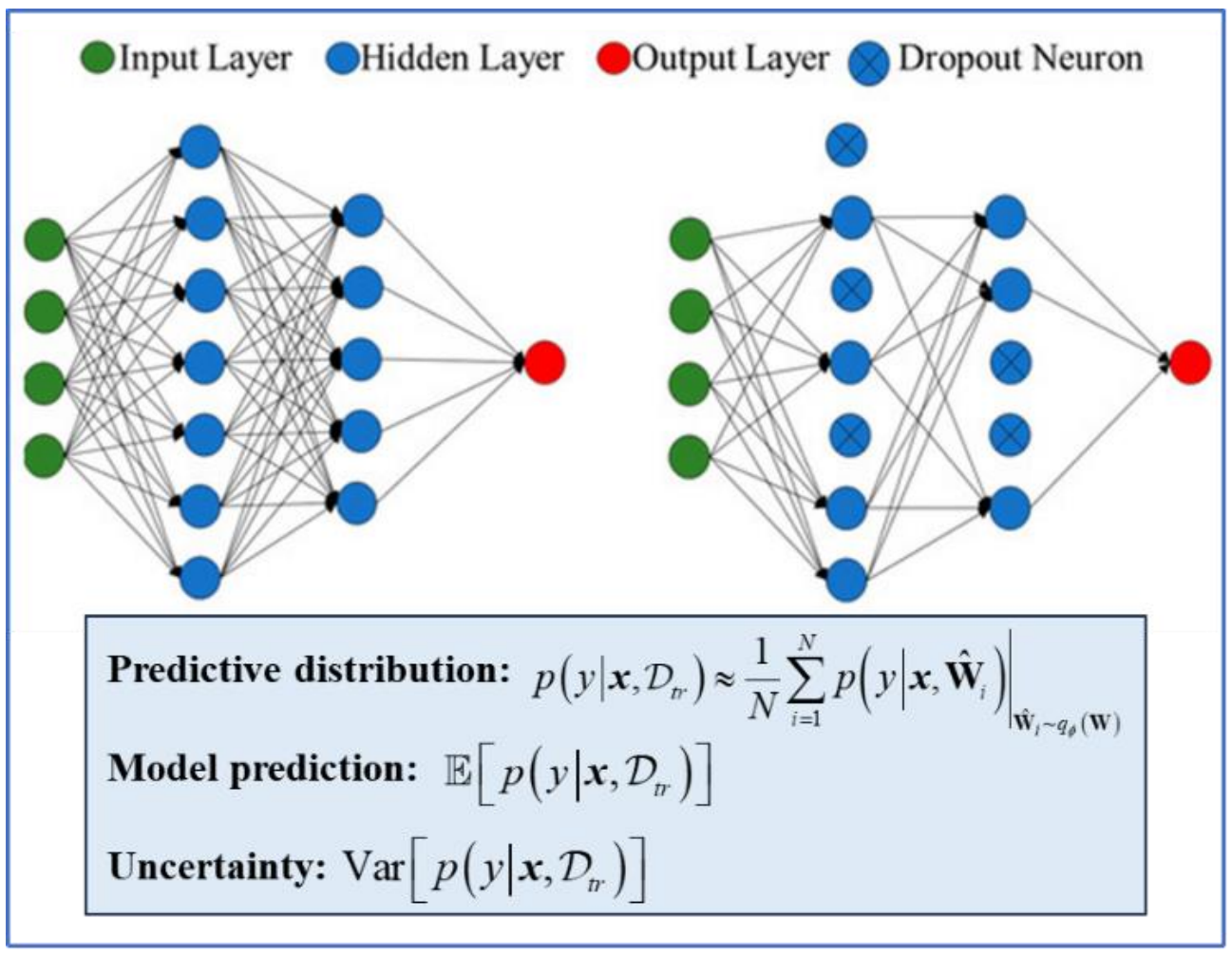

Figure 14. A fully connected neural network (left) and the same network with dropout in a particular training step (right).

#### *4.3.2 Shallow Bayesian neural networks*

Shallow NNs are simple NN architectures that only contain three layers: the input layer, the hidden layer, and the output layer, which are more interpretable and less computationally expensive than DNNs with multiple hidden layers but have a limited capacity to capture complex hierarchical features and representations. Due to their simplicity, original approximation methods for BNNs are mainly implemented on shallow NNs to investigate their applicability and UQ capability [28], and these models are referred to as shallow BNNs in this work. A remarkable advantage of shallow BNNs is that the optimization objective is analytical tractable using many approximation methods without strong assumptions [28]. Additionally, it has been proved that shallow BNNs share a close theoretical relationship with GPs, as a shallow BNN with an infinitely wide hidden layer converges to a GP when its hidden-to-output weights and output biases are i.i.d. with zero mean and finite variance [177]. This connection provides shallow BNNs with a comprehensive framework for uncertainty quantification by leveraging GP's machinery, while also enhancing the interpretability of their uncertainty estimates. Nevertheless, shallow BNNs have not become mainstream in uncertainty estimate for modern NNs due to the limited capacity inherently associated with shallow NNs. In structural dynamics, the demand for shallow BNNs in uncertainty estimate also focuses on small scale problems that require model interpretability, such as fatigue analysis [25, 128] and damage diagnosis [129]. These applications demonstrate the advantages of shallow BNNs in interpreting and quantifying predictive uncertainty, offering robust guidance for decision making.

#### *4.3.3 Bayesian broad learning*

Broad learning system (BLS) [181] is a computationally efficient alternative to traditional DNNs, which, similar to shallow NNs, only comprises of three layers. As illustrated in Fig. 15, the hidden layer of the BLS combines the feature nodes, which represent the extracted features from raw inputs, with the enhancement nodes designed to improve the generalization property. Compared to DNNs, BLS is more efficient in training and adapting reconfiguration of network architecture by utilizing an incremental learning strategy [181]. However, BLS is sensitive to the number of feature nodes and thus often requires a large network during training, which highlights its limited ability to capture epistemic uncertainty and makes it prone to overfitting [130, 182]. Bayesian broad learning (BBL) addresses this issue by treating the connecting weights of the last layer as uncertain variables (Fig. 15) with a zero-mean Gaussian [182] or uniform prior [130]. The posterior distribution of these weights is then derived using Bayes' theorem to formulate the predictive distribution and estimate predictive uncertainty according to the framework described in Section 2.3.1. Compared to shallow BNNs, BBL offers higher capacity due to its flat structure, while it is more efficient than BDL models owing to its simpler architecture. However, BBL only accounts for epistemic uncertainty from the weights of the last layer while ignoring the uncertainty in feature extraction and enhancement, making it less prevalent for UT in NNs than BDL. For more details about BBL, one can refer to [130, 182].

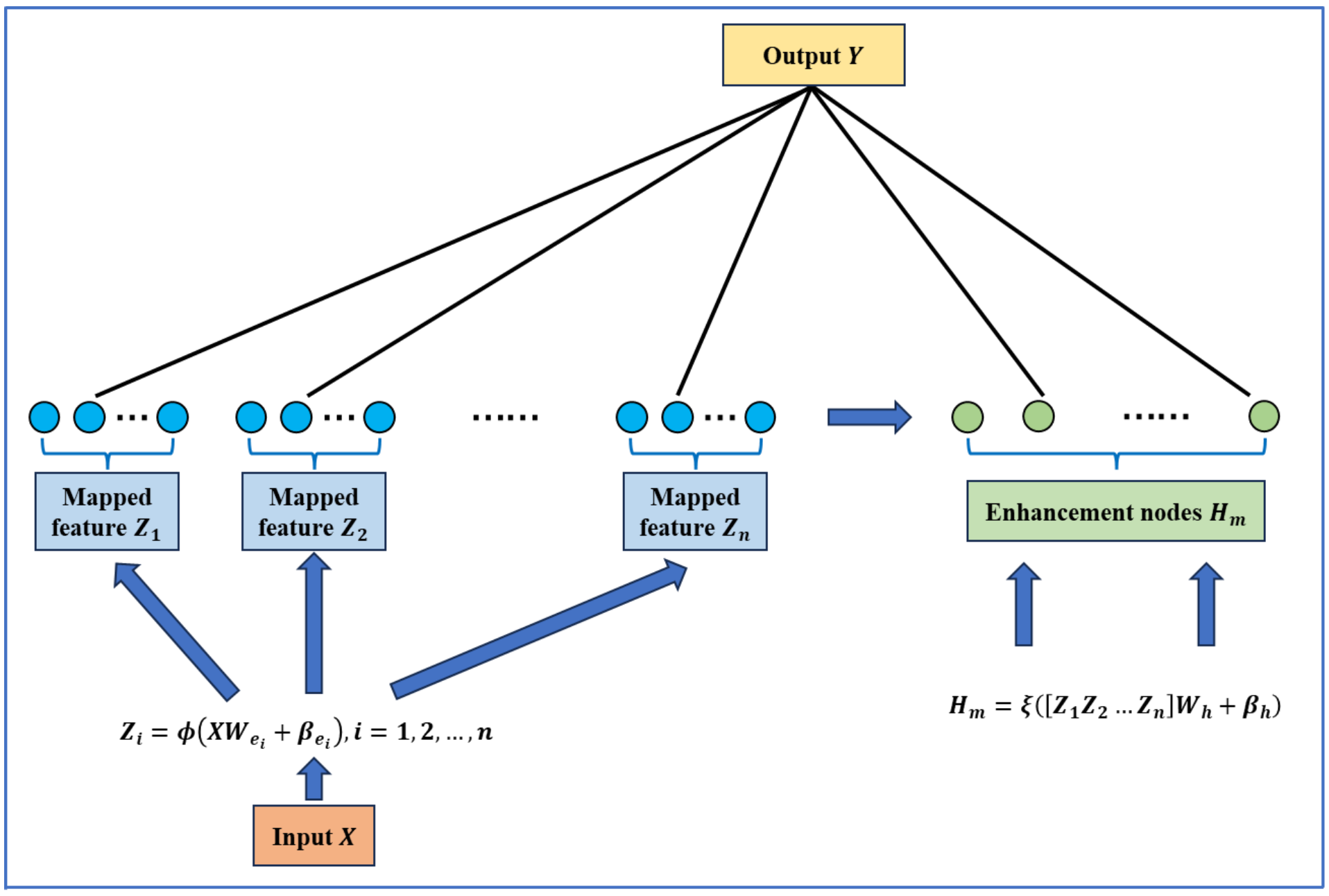

Figure 15. The architecture of a Bayesian broad learning system.

### *4.3.4 Bayesian deep learning*

DNNs dominate modern ML due to their notable advantages such as feature extraction, representation learning, nonlinear mapping, and scalability. In addition to the fundamental multilayer perceptrons (MLPs) composed of multiple fully connected layers, various DNN architectures, including convolutional neural networks (CNNs), recurrent neural networks (RNNs), and autoencoders (AEs), have been developed over the years to address diverse real-world problems. However, traditional DNNs are prone to overfitting, especially when trained on limited or noisy data or when using overly complex network architectures, which is largely attributed to their ignorance of uncertainty in their predictions.

To meet the demands for both model capacity and robustness in challenging real-world tasks, BNNs are required to be effectively extended to these complex NN architectures using modern approximate inference techniques, which is the most inherent reason of the prevalence for MC dropout in Bayesian deep learning (BDL) due to its efficiency and scalability. The most straightforward implementation of BDL treats the weights of MLPs as uncertain variables to formulate the predictive distribution and estimate predictive uncertainty, as illustrated in Section 4.3.1. This approach has been widely applied in structural dynamics [100, 132, 138, 139, 145, 183] to improve model robustness, which is particularly valuable for engineering practice where training data are commonly noisy or limited. Additionally, predictive uncertainty can guide subsequent decisions, with high uncertainty suggesting the need for more data or a more appropriate network architecture. However, while effective for UT and performance enhancement, BNNs based solely on fully connected networks are often inadequate for advanced tasks such as image processing, highlighting the need to extend BNN frameworks to more complex modern DL models with higher capacities, which thus becomes the focus of this section.

#### *4.3.4.1 Bayesian convolutional neural networks*

CNNs use a mathematical operation called convolution to replace general matrix multiplication in at least one of the layers for processing data that has a known, grid-like topology such as images composed of pixel intensities in the three-color channels [89], which are composed of five types of layers, namely the input layer, convolutional layers, pooling layers, fully connected layers, and the output layer. A convolution layer convolves the input matrix with a convolution kernel at every spatial position, resulting in a series of dot products that preserve spatial information in the input, followed by which is a pooling layer that simply takes the output of the convolution layer and reduces its dimensionality. Through a stacked combination of convolution and pooling layers, CNNs effectively transform the raw input matrix into a feature map and extract important features automatically [89], but they are prone to overfitting when dealing with small datasets due to the complex architecture, highlighting the significance of UQ for them.

Typically, BNNs are merely implemented on the fully connected layers for uncertainty estimate in CNNs, as shown in Fig. 16, which is equivalent to applying a finite deterministic transformation to the raw inputs before feeding them into BNNs [28]. For more comprehensive uncertainty estimation, Shridhar et al. [184] extended Bayes by Backprop to convolutional layers by approximating the true posterior of kernel weights using a Gaussian distribution with coupled mean and variance through VI. During training, two sequential convolutional operations are performed to update the mean and variance of the variational distribution using the reparameterization trick. The optimized variational distribution is then used to formulate the predictive distribution of convolutional layers. This method provides a fully Bayesian treatment of CNNs within the Bayes by Backprop framework, allowing for capturing epistemic uncertainty in both fully connected and convolutional layers. Alternatively, for Monte Carlo dropout, Gal et al. [28, 178] proposed a Bayesian CNN by placing a prior distribution over each kernel and approximately integrating each kernels-patch pair with Bernoulli variational distributions, which is equivalent to applying dropout after each convolution layer before pooling. Consequently, this method accounts for epistemic uncertainty arising from not only the fully connected layers but also the convolution layers in feature extraction. Subsequently, the predictive distribution can be approximated using MC dropout. Compared to using Bayesian inference only on fully connected layers, this method offer higher interpretability of uncertainty in CNNs without introducing much computational cost, thus it has been widely applied in engineering fields including structural dynamics [142, 144].

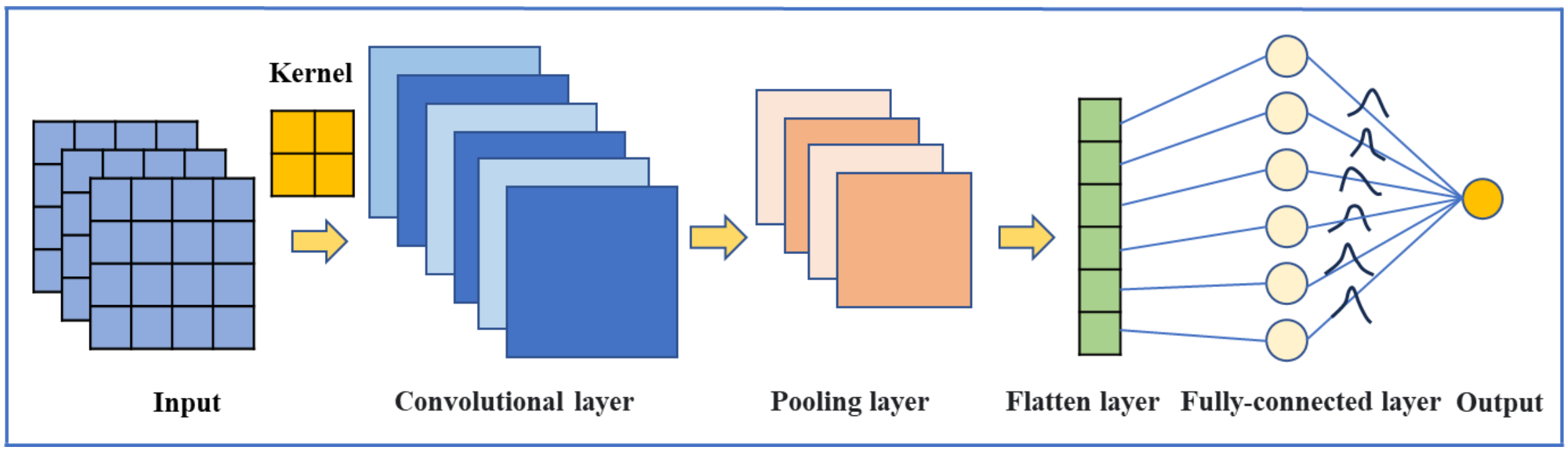

Figure 16. The architecture of a typical Bayesian convolutional neural network.

#### *4.3.4.2 Bayesian recurrent neural networks*

RNNs are designed for modeling and predicting sequential data such as time series dynamic responses, which has internal memory to preserve information from previous inputs in the network internal state and thereby influencing the network output [89], as depicted in Fig. 17. To implement Bayes by Backprop in RNNs and their variants, Fortunato et al. [185] proposed a framework in which the sequences in each minibatch of training data are truncated, with the initial state of a sequence set to be the final state of its preceding sequence. This setup allows standard Bayes by Backprop to be applied by sampling network parameters for each minibatch of training data, while keeping the parameters of the variational posterior fixed throughout each sequence within the current minibatch. Owing to its simplicity and scalability, this framework has found applications in structural dynamics to address uncertainty in sequential modeling tasks such as response prediction and remaining useful life estimation [133, 143]. In contrast, applying MC dropout in Bayesian RNNs is challenging, as dropout cannot be applied in RNNs except for the forward connections [28, 186, 187]. To overcome this limitation, Gal [28] proposed a dropout variant that randomly masks (zeros) rows of each weight matrix across all time steps, which is identical to dropping the same network units and randomly dropping inputs, outputs, and recurrent connections in an RNN at each time step. This method enables the formulation of Bayesian RNNs using MC dropout and can be extended to variants of RNNs such as gated recurrent units (GRUs) and long-short term memories (LSTMs) by applying different dropout masks for different gates [28]. The predictive distribution can be estimated by performing dropout at test time for UT. Owing to its computational efficiency, this method has also been applied in structural dynamics for UT [134].

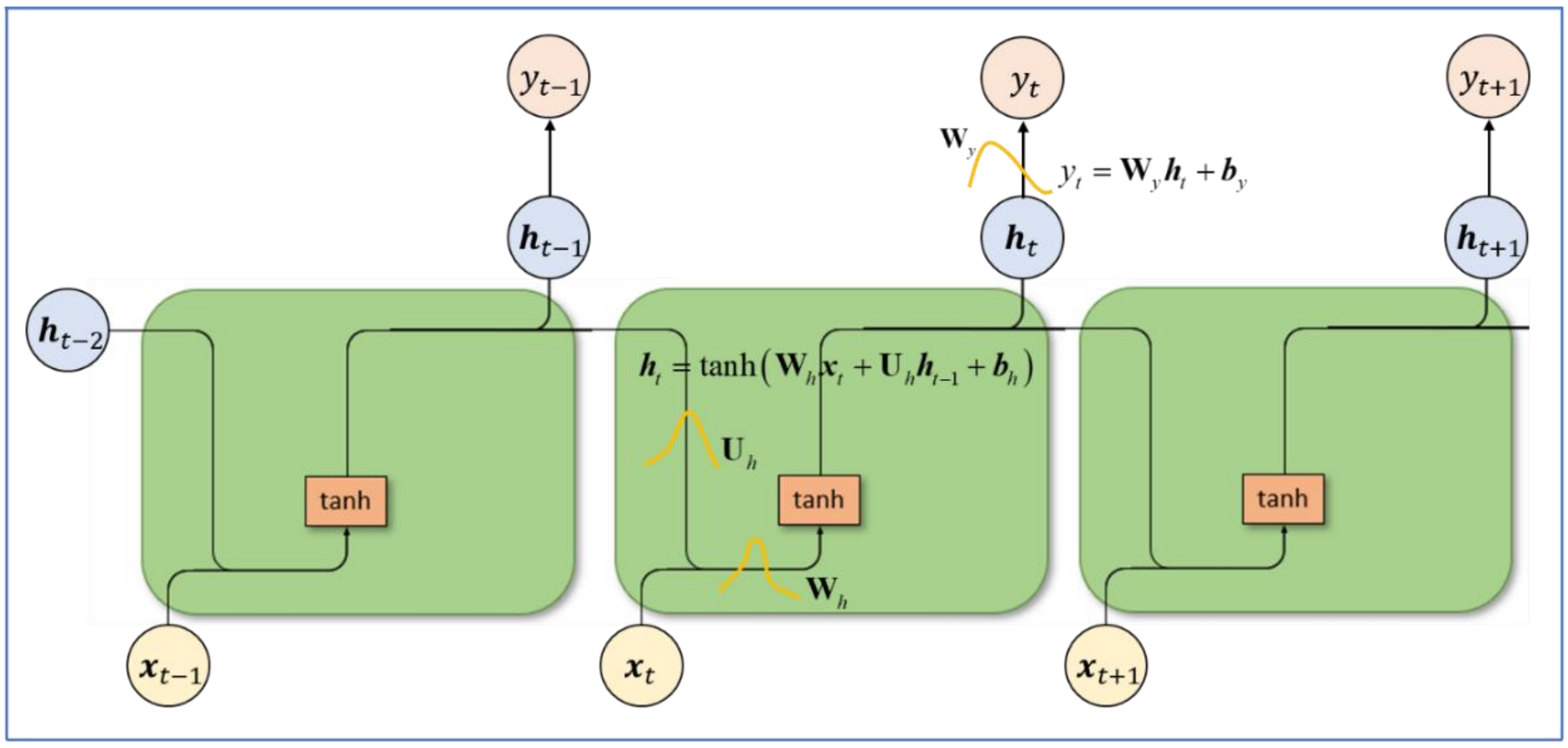

Figure 17. The architecture of the Bayesian recurrent neural network proposed in [28].

#### *4.3.4.3 Variational autoencoders*

AEs are NN architectures designed to copy the inputs to the outputs in an unsupervised manner, are composed of two types of NNs known as the encoder and the decoder. The encoder transforms data from a higher-dimensional space into compressed and meaningful representations, referred to as codes, within a lower-dimensional latent space, while the decoder aims to reconstruct the input based on these codes [89]. Therefore, AEs are effective feature extractors with the latent space capturing the most valuable information from the data space and ignoring redundant noise and ambiguity, but they are prone to overfitting due to the non-regularized latent space [89]. Variational autoencoders (VAEs) [155] (Fig. 18) are motivated to address this issue by modeling the codes $z$ as uncertain variables that follow some distribution (normally Gaussian), namely $z \sim p(z)$. The encoder and decoder can then be viewed as two conditional distributions represented by two NNs that maps from the input space to the latent space and from the latent space to the output space, denoted by $q_\phi(z|x)$ and $p_\theta(x|z)$, respectively. The loss function can then be formulated using VI to model the ELBO:

$$\mathcal{L}_{VAE}(\theta,\phi) = \sum_{x_i \in \mathcal{D}_{tr}} \mathbb{E}_{q_\phi(z|x_i)}\left[\log p_\theta(x_i|z)\right] - D_{KL}\left(q_\phi(z|x_i) \| p_\theta(z)\right) \tag{41}$$

Training the VAE requires the gradient of the loss function with respect to both the variational parameters $\phi$ and the generative parameters $\theta$, but difficulty arises from the high variance and impracticality of traditional Monte Carlo gradient estimator in approximating $\nabla_\phi \mathcal{L}_{VAE}(\theta,\phi)$ [155]. To address this issue, Kingma and Welling [155] proposed a reparameterization trick by introducing an auxiliary variable, which yields a differentiable estimator of the ELBO with lower variance, as detailed in Section 4.3.1 during the introduction of Bayes by Backprop.

It is worth noting here VAEs are different from typical BNNs since the NNs in VAEs are deterministic and only the latent variables are uncertain. Nevertheless, VAEs are a remarkable combination of NNs with Bayesian inference, and the reparameterization trick has been widely applied in modern approximating techniques for BNNs including both Bayes by Backprop and MC dropout. Consequently, VAEs are included in this section with a great emphasis. In terms of UQ, VAEs account for uncertainty from two aspects, namely the input uncertainty and output uncertainty [17]. Specifically, VAEs embed each sample through a variational posterior $q_\phi(z|x)$ instead of deterministic codes, where the mean represents the extracted feature and variance serves as an estimate of uncertainty. Similarly, the reconstructed output is governed by a generative model $p_\theta(x|z)$ for uncertainty estimation. As a result, VAEs are powerful models in learning meaningful representations and eliminating redundant information from the model input, thus have been widely used in structural dynamics as an automatic feature extractor for tasks such as optimal senser placement [188] and damage diagnosis [123, 124, 189].

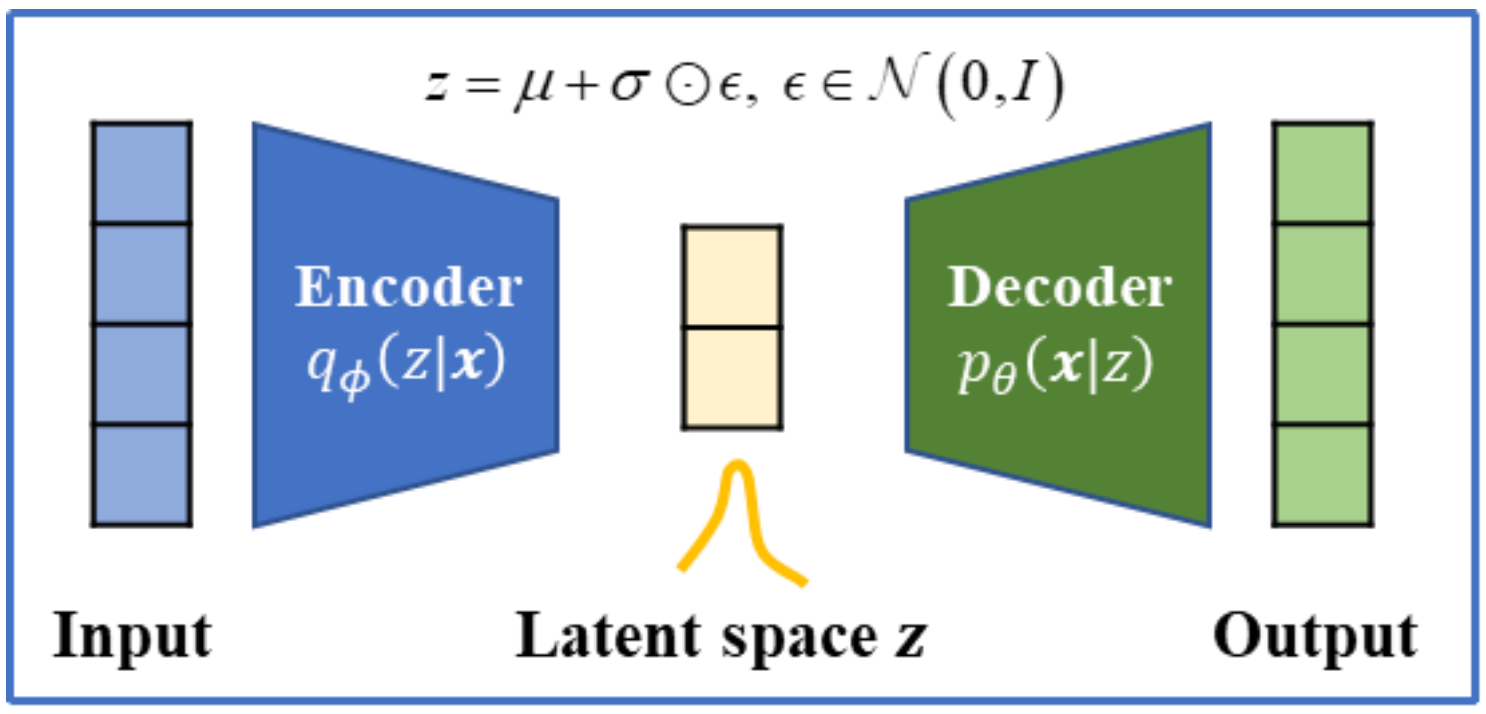

Figure 18. The architecture of a variational autoencoder that addresses uncertainty in latent variables.

***4.3.3.4 Bayesian deep transfer learning***

In the context of ML, an implicit assumption is that the training data and testing data originate from the same input space with identical marginal distributions. However, this assumption is usually unsatisfied in the context of structural dynamics, as datasets from different structures or different operating conditions do not conform to the same distribution [190], which could significantly exacerbate the lack of well-annotated training data when using ML models in structural dynamics. Transfer learning (TL) has been widely recognized as a potential candidate to alleviate this issue, which tries to train a ML model on a known domain (source domain) to solve the problems on the target domain [191]. Conditioned on the dominance of DNNs in contemporary ML, current TL methodologies are usually applied to DL models, a subset referred as deep transfer learning (DTL), which can be divided into four categories [192], as illustrated in Fig. 19. The uncertainties in TL are primarily dominated by two key aspects: (i) the methodology for transferring information, and (ii) the selection of relevant information to transfer between the source and target domains. Consequently, the Bayesian framework naturally lends itself potential in DTL applications due to its unique advantages for UT, especially in scenarios where expert knowledge is highly limited [193, 194].

Generally, Bayesian TL has the following three paradigms [193]:

**(1) Shared parameters:** This approach assumes shared or partially shared parameters in the likelihood specification for source and target data, whose posterior distribution conditioned on source domain data is used as the prior in the target domain for the analysis of likelihood and posterior [193]. This approach can be used in instance-based, mapping-based, and network-based DTL.

**(2) Hierarchical model:** This approach posits that the parameters of source and target data come from a jointly specified or identical prior distribution, while the Bayesian framework can be implemented by assigning hyperpriors for the parameters of this prior distribution to address uncertainties in model specification.

**(3) Shared latent space:** This approach aims to learn a latent space shared by data from both the source and target domains. Subsequently, data from both domains are projected onto this latent space with task-specific projection matrices to train the DL model. This method is typically used in mapping-based DTL, providing the flexibility of operating without labels in the target domain, while Bayesian method can be used in the process of determining the projection matrices for UT.

These Bayesian DTL methods enhance the capabilities of DTL techniques in addressing both aleatoric and epistemic uncertainties stemming from knowledge transfer, rendering DTL more reliable when utilized to alleviate the insufficiency of well-annotated data for training DNNs in the target domain. A comprehensive literature survey on state-of-the-art Bayesian DTL methods is provided in [194], underscoring the potential of Bayesian DTL in improving the generalizability of DL models. However, despite these promising aspects, there is a scarcity of research applying Bayesian DTL in structural dynamics, representing a notable research gap for future studies.

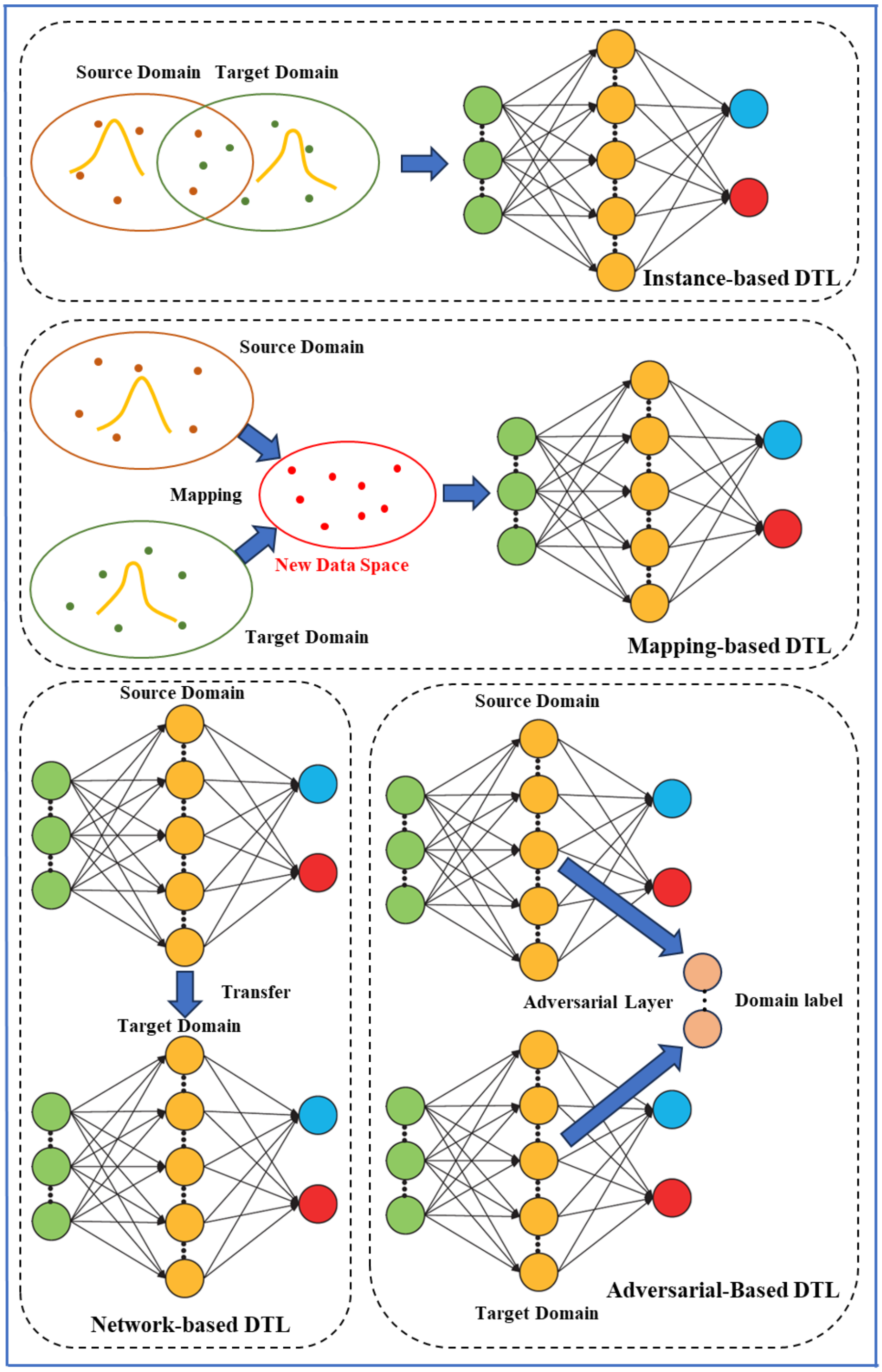

Figure 19. Categories of deep transfer learning

## 5 Non-probabilistic Machine Learning Techniques Considering Uncertainties

Non-probabilistic convex models represent a class of UQ models utilized in scenarios where information is limited or when focusing on uncertainty bounds. While less common than probabilistic models, non-probabilistic convex models have found extensive application in various engineering fields, including structural dynamics. These methods are particularly suitable for situations where only boundary information is crucial in UQ and propagation of uncertainty. They require minimal data and exhibit superior computational efficiency compared to probabilistic approaches. However, their rough approximation of uncertainties may limit their ability to capture the intricate structure and characteristics of the data effectively. Common non-probabilistic convex models include the traditional interval model, multidimensional ellipsoid model, and multidimensional parallelepiped model. These models face limitations in terms of compactness for arbitrarily distributed samples and the inability to characterize samples with multi-modal distributional features. The integration of emerging ML techniques offers new avenues to address these limitations. Therefore, this section is structured from an ML perspective, encompassing unsupervised and supervised non-probabilistic methods for UQ. Table 4 summarizes the literature on applications of non-probabilistic ML methods for UT in the field of structural dynamics.

Table 4. Selection from the literature on applications of non-probabilistic ML methods for UT in the field of structural dynamics

| UT method | | Advantages | Disadvantages | Ref. |
| --- | --- | --- | --- | --- |
| Non-probabilistic methods | PCA | simplicity; efficiency | not suitable for nonlinear data | [195] |
| | K-means | capability of quantifying clustered data | requiring the number of components specified | [196] |
| | Fuzzy clustering | adaptively determining the number of clusters | relatively complex | [197] |
| | | | | [198] |

| | | | |
|---|---|---|---|
| | | | [199] |
| DINN | ensuring enclosure of prediction with sharp bounds | unstable training process | [200] |
| FNN | efficiency; automatic differentiation | unsuitable for small variations | [201] |
| | | | [202] |
| LUBE | robust to the problem of interval expansion | sensitive to choices of hyperparameters | [203] |
| Interval predictor model | clear mathematical formulation | strong requirements based on semidefinite programming and sum of squares | [204, 205] |
| Sliced normal distribution | flexible; multimodal distribution | complex computation | [206] |

### *5.1 Unsupervised learning for non-probabilistic uncertainty analysis*

#### *5.1.1 Principal component analysis (PCA)*

The traditional interval model has been extensively used to quantify structural uncertain parameters and responses. Since the traditional interval model does not consider the correlations between uncertain variables, it may result in an overconservative estimation. In this light, improved convex models have been presented by Elishakoff et al. [207]. Cao et al. [195] proposed the non-probabilistic polygonal convex set (PCS) model for uncertainty quantification, which aims to generate a more compact convex model compared with interval model. As Fig. (20a) shows, a traditional interval model is firstly constructed. Then, principal component analysis in terms of the samples is conducted to establish a new coordinate system based on the principal directions of PCA. In this coordinate system, a new interval model termed PCA interval model can be constructed. Finally, the PCS model is obtained by the intersection of the PCA interval model and the traditional interval model as given in Fig. (20b).

Compared with the traditional interval model, the PCS model is more compact, indicating that the uncertainty is reduced. Therefore, the PCA could improve the effectiveness of the uncertainty quantification. Nevertheless, the PCS model exhibits two limitations: (1) its applicability is primarily suited for samples distributed within an inclined cube, despite its relatively compact nature; (2) the model's boundaries are more intricate compared to the traditional interval model, necessitating increased computational resources for uncertainty propagation.

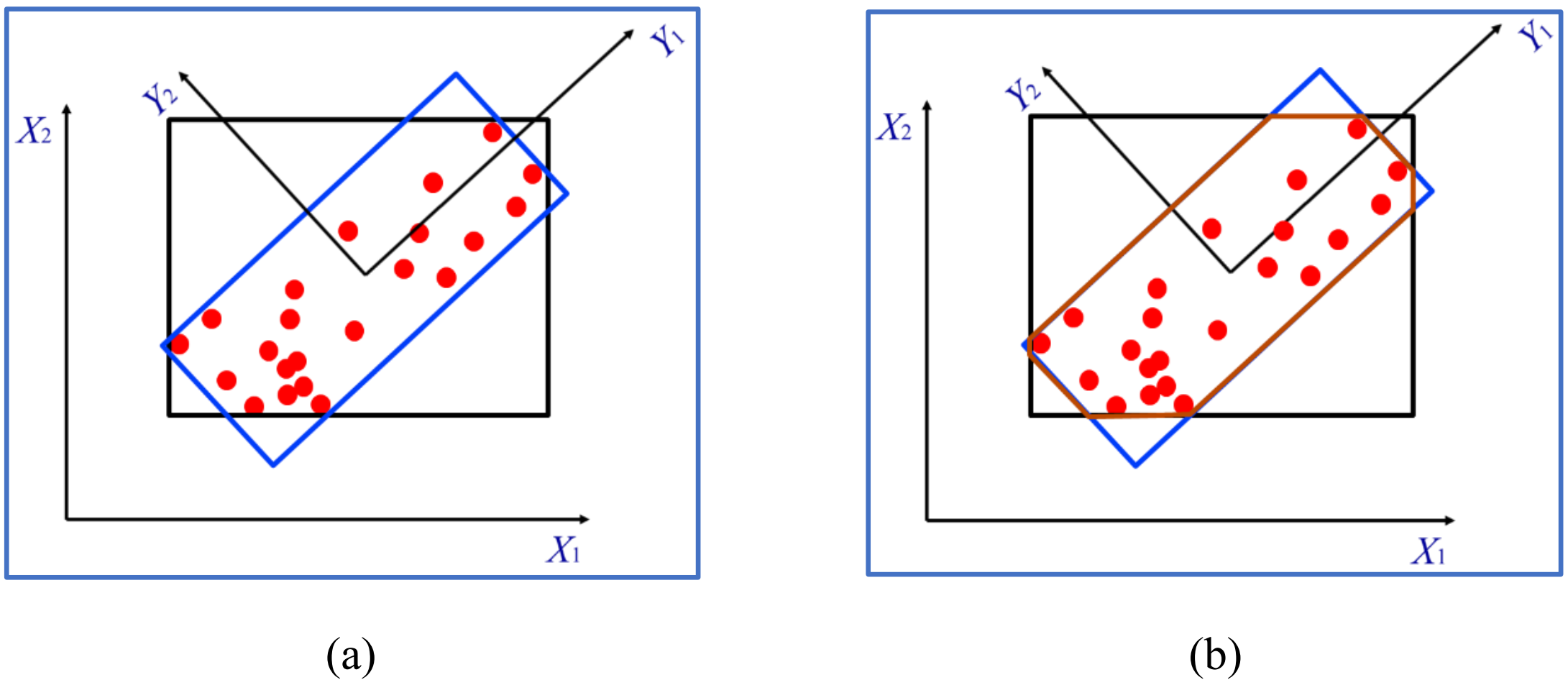

Figure 20. Non-probabilistic polygonal convex set model: (a) The two-dimensional PCA interval model; (b) The two-dimensional PCS model (reproduced from [195]).

### *5.1.2 Non-probabilistic uncertainty modeling based on clustering techniques*

The PCS model aims to obtain a compact model when the traditional interval model is less informative. However, it cannot deal with multimodal samples or inconstant samples. In order to quantify inconstant uncertainty, clustering techniques have been introduced in uncertainty modeling.

##### *5.1.2.1 Multimodal ellipsoid model (MEM) based on GMM*

In terms of the multimodal samples with different centers, Liu et al. [208] proposed the multimodal ellipsoid model (MEM) for non-probabilistic uncertainty quantification. Considering that the Gaussian kernel coincides with the ellipsoid model, hence, the Gaussian mixture clustering was employed in the MEM. According to the clustering results, a traditional ellipsoid sub-model can be constructed for each cluster and the final uncertainty model is the union of all the sub-models. Fig. 21 shows that the MEM can well characterize the multimodal variables. Furthermore, the non-intersecting and intersecting conditions are automatically determined by the GMM algorithm according to the distributional features of the samples. The union of all the sub-models forms the final non-probabilistic convex model.

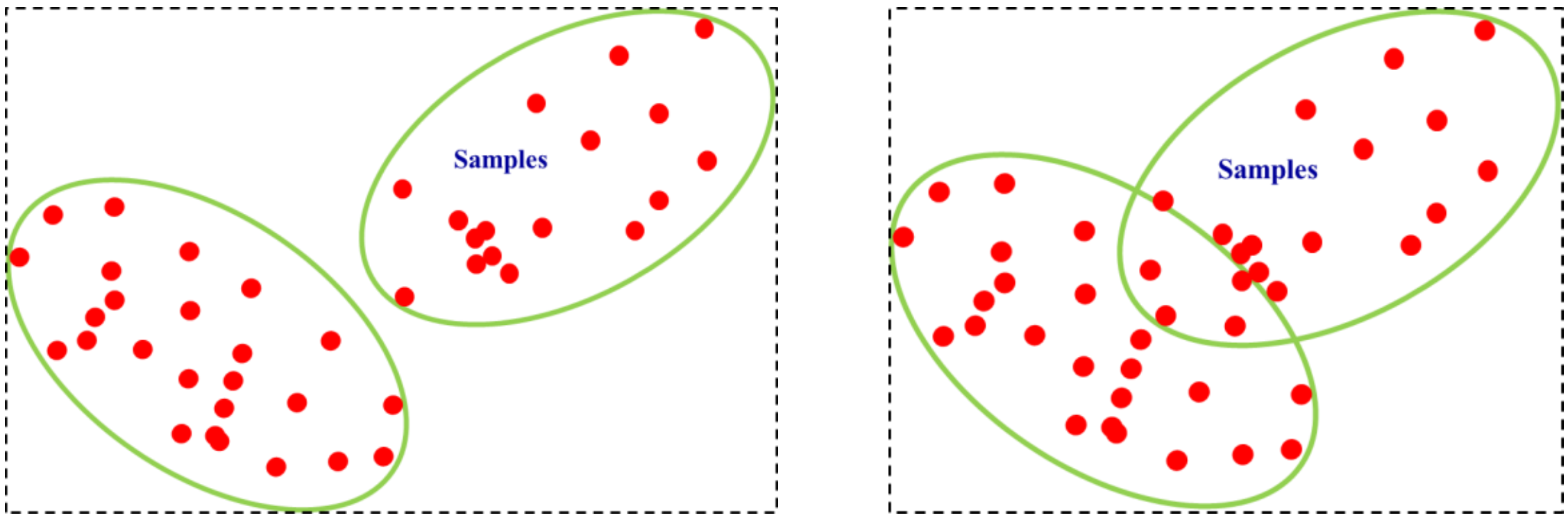

(a) (b)

Figure 21. The two-dimensional multimodal ellipsoid model: (a) Non-intersecting MEM; (b) Intersecting MEM (reproduced from [208]).

##### *5.1.2.2 Sub-parallelepiped model based on K-means clustering*

Wang et al. [196] proposed the sub-parallelepiped modeling for inconstant uncertainty in multi-parallelepiped modeling. As Fig. 22 shows, the two-dimensional samples of an uncertain variable have two cluster centers. The two blue parallelograms can compactly enclose the samples and hence is a suitable quantification of the uncertain variable. If we use the red parallelogram to quantify the variable, the uncertainty domain will be much larger than the two blue parallelograms, which may result in overestimation in subsequent computations. In the modeling process of the sub-parallelepiped model, the K-means clustering method is adopted to divide the available experimental samples into several non-empty sample subsets. Subsequently, the parallelepiped construction procedures are executed to generate a sub-parallelepiped model for each sample subset. Based on the union operation, the final parallelepiped model to quantify the inconstant uncertainty is obtained.

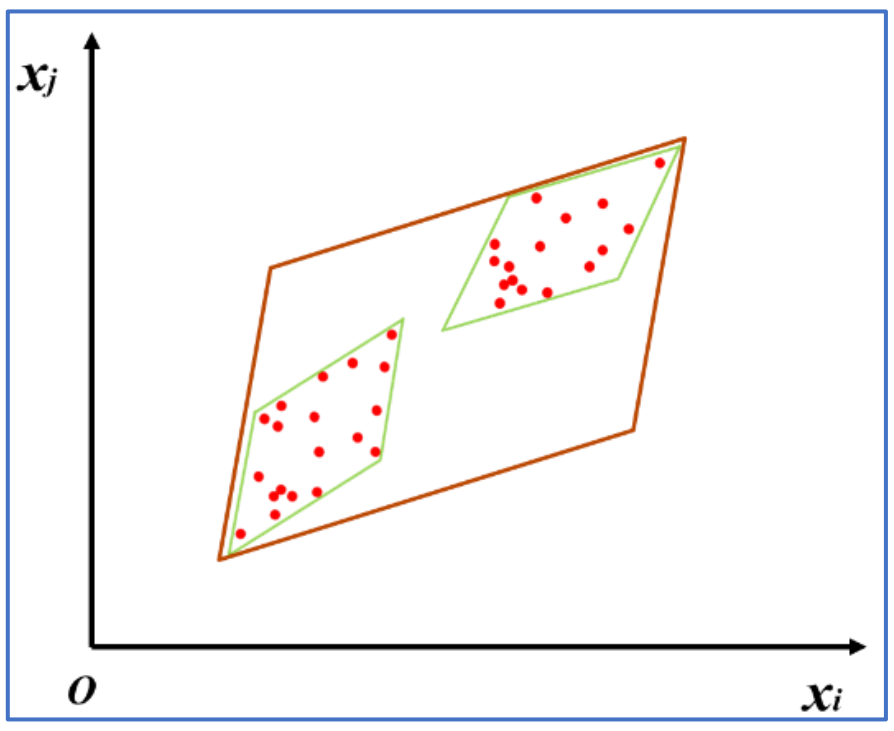

Figure 22. The two-dimensional sub-parallelepiped model (reproduced from [196]).

In the work of [196], the interaction between variables may be discarded. The work of Faes and Moens [209] considered multivariate dependent interval finite element analysis, where bivariate dependence between intervals is considered based on copula method.

#### *5.1.2.3 Fuzzy equivalence relation-based clustering method (FERCM)*

In addition to GMM and K-means clustering, fuzzy clustering is another widely adopted clustering technique. Wang et al. [197] investigated the dispersed samples using fuzzy equivalence relation-based clustering method. Assume there are *n* samples to be clustered, then a fuzzy equivalence relation matrix $\boldsymbol{R}=\left(r_{ij}\right)_{n\times n}$ can be constructed to characterize the similarity level between any two samples $x_i, x_j$. If $r_{ij}$ is equal to or larger than the given similarity threshold $\lambda$, the two samples $x_i, x_j$ will be clustered to the same group. Distinct from abovementioned methods, the FERCM need not to specify the number of clusters in advance. It realizes automatic determination of cluster numbers by defining the $\lambda$-cut matrix of the fuzzy equivalence matrix. Readers can refer to [197] for more details. In Fig. 23, the two-dimensional samples are separated into two clusters and each cluster is quantified by an ellipsoid. The final domain is the union of the ellipsoids.

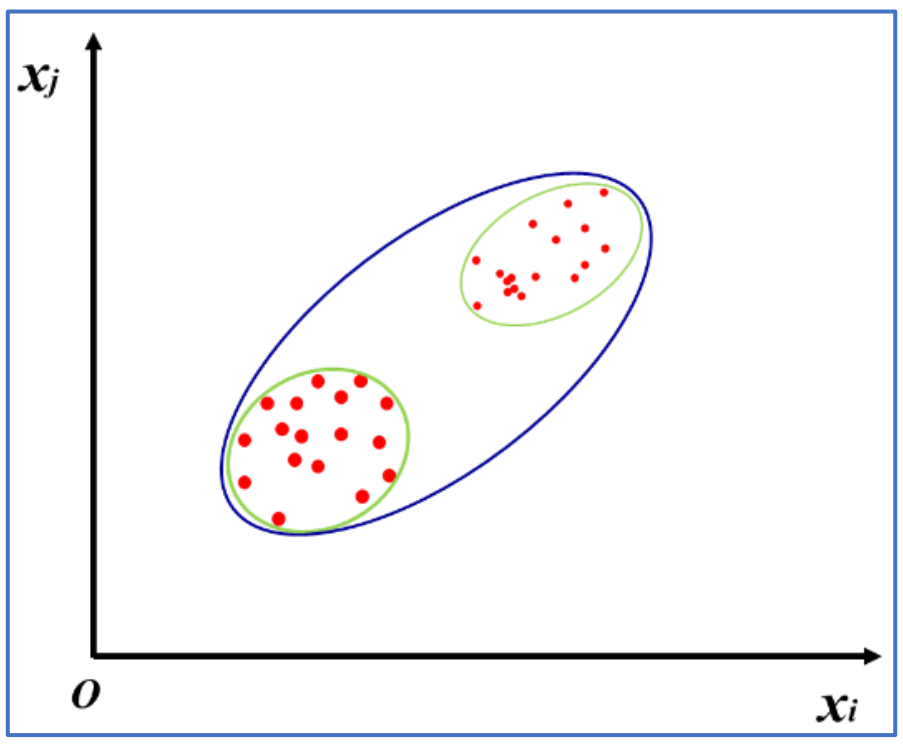

Figure 23. The two-dimensional clustered ellipsoid model (reproduced from [197]).

### *5.2 Supervised learning for non-probabilistic uncertainty analysis*

#### *5.2.1 Deep interval neural network (DINN)*

Deep neural networks [89] possess strong abilities in both regression and classification tasks, and various neural network models have been used in structural engineering. Despite their exceptional prediction capabilities, existed DNNs do not have a mechanism to quantify and propagate non-probabilistic uncertainty through neural networks. In this light, Betancourt et al. [200, 210] developed a deep interval neural network (DINN) which directly incorporates the interval arithmetic (IA) into the feedforward neural network and therefore the DINN can propagate interval uncertainty through the network. The DINN is a supervised learning algorithm in a regression setting which seeks to learn an interval predictive model $F:\left[\underline{X},\overline{X}\right]\rightarrow\left[\underline{Y},\overline{Y}\right]$. As Fig. 24 shows, the basic structure of the DINN is similar to deterministic NNs, whereas the inputs $\left[\underline{X},\overline{X}\right]$, outputs $\left[\underline{Y},\overline{Y}\right]$ and weights $\left[\underline{W},\overline{W}\right]$ of the DINN are intervals, which is the essential difference between deterministic and interval neural networks. The propagation from interval inputs to interval outputs are realized through IA. The DINN mainly has the following contributions: (1) it quantifies the aleatoric and epistemic uncertainty in the input by using interval analysis; (2) it incorporates the IA into a feedforward neural network; (3) it can quantify the parameter uncertainty of the neural network itself as part of the algorithm. Despite above contributions, the interval overestimation as an intrinsic feature of IA, is a severe issue in both forward uncertainty propagation and back-propagation parameters updating through the network. Each computation starting from the second layer of the network using IA will confront with interval expansion due to the interval dependency problem. Although the authors adopted some tricks in training to alleviate this issue, it cannot be completely avoided under the framework of IA. To address this issue and enhance training stability, the authors implemented several strategies during the training of the DINN. First, it is required that the interval functions in the DINN exhibit Lipschitz continuity. Second, the algorithms are designed to minimize interval dependencies. Third, due to the nature of interval matrix multiplications, there is a need to avoid large weights that could result in exploding gradients.

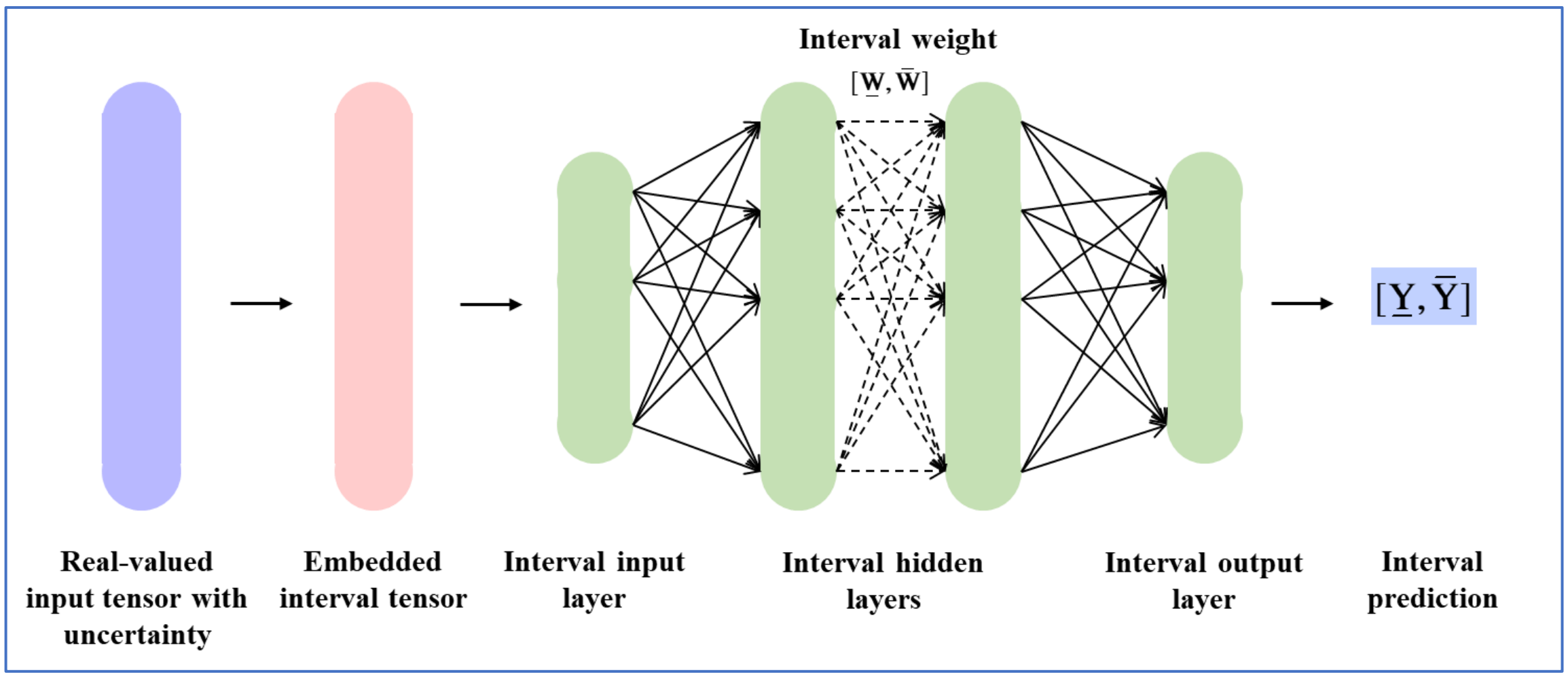

Figure 24. DINN architecture (in which the input, parameters and output are intervals) reproduced from ([200]).

### 5.2.2 *Feedforward neural network-based uncertainty propagation methods*

Feedforward neural network (FNN) is a classical supervised learning model. It has been directly used in non-probabilistic uncertainty analysis as surrogate model for its strong fitting capability and efficient automatic differentiation mechanism. MC simulation is a universal and powerful tool for propagating uncertainties. In traditional non-probabilistic uncertainty propagation methods, the MC results are usually taken as true results for comparison. However, the high precision of the MC method relies on a large number of samples, which is infeasible for computationally expensive structural models. Deng et al. [211] employed a radial basis neural network to replace the original structural model function mapping the structural parameters to modal characteristics. Then, MC simulation was conducted based on the constructed neural network. The efficiency of the proposed method outperformed the traditional MC method. Apart from surrogate-oriented applications, neural networks have been utilized in other non-probabilistic uncertainty propagation methods for their efficient differentiation. The interval perturbation method (IPM) [212] is an efficient method to calculate the output intervals given input intervals. For example, assume the $n$-dimensional structural interval parameters are $\boldsymbol{x}^I = [\underline{\boldsymbol{x}}, \overline{\boldsymbol{x}}]$, then the predicted structural outputs $Y^I$ can be estimated using the first-order Taylor expansion as follows:

$$\begin{aligned} \overline{Y}(\underline{x}, \overline{x}) &= Y(x^c) + \sum_{i=1}^{n} \left| \frac{\partial Y(x)}{\partial x_i^c} \right| x_i^w \\ \underline{Y}(\underline{x}, \overline{x}) &= Y(x^c) - \sum_{i=1}^{n} \left| \frac{\partial Y(x)}{\partial x_i^c} \right| x_i^w \end{aligned} \tag{42}$$

where $x^c$ and $x^w$ are the midpoint and radius of $x^I$, respectively; $x_i^c$ and $x_i^w$ are the *i*-th components of $x^c$ and $x^w$, respectively.

Since the IPM requires the first order derivatives of its outputs, it is suitable for a case whose numerical differentiation is easy to compute, but most practical problems are non-differentiable systems. Considering that neural networks have a strong approximation capability and automatic differentiation scheme, the structural output function $Y(x)$ can be replaced by an FNN: $F(x)$ [201]. In the FNN differentiation method, an FNN is constructed to approximate the structural output of interest, and the first order derivative $\partial F(x)/\partial x$ can be derived from the FNN according to backpropagation algorithm. In this approach, the FNN serves as a deterministic surrogate model and automatic differentiation calculator. This non-intrusive uncertainty propagation method is very efficient. However, there are several shortcomings that may hinder its practical application. Firstly, a large number of neurons are needed. As stated in Ref. [201], achieving precise approximations of the derivatives of a structural function demands significantly more neurons than those required for approximating the function itself, potentially restricting the method's approximation capacity and efficiency. Secondly, perturbation-based methods are primarily suitable for small-interval problems. For problems with large uncertainties, the truncation error in the first-order expansion can lead to substantial estimation errors in the interval bounds. When variable intervals are extensive and/or the function exhibits high nonlinearity, the first-order Taylor expansion may introduce considerable errors.

### *5.2.3 Lower upper bound estimation framework*

Lower upper bound estimation (LUBE) is proposed by Khosravi et al. [213] for directly generating prediction intervals (PIs). Fig. 25 illustrates the structure of LUBE interval prediction using a neural network. The network will produce the lower and upper bounds enclosing the actual output. In order to construct the optimal PIs, a PI-based cost function is required. Khosravi et al. [214] defined the measures for the quantitative assessment of PIs. By definition, future observations are expected to lie within the bounds of the PIs with a prescribed probability called the confidence level $\left((1-\alpha)\times 100\%\right)$. Assume that there are $n$ targets (true values) $y_1, y_2, \cdots, y_n$, and the outputs of the LUBE network are the PIs comprising the lower bound $\hat{y}_{Li}$ and upper bound $\hat{y}_{Ui}$ of $y_i\left(i=1,\cdots,n\right)$. It is expected that the coverage probability of PIs will asymptotically approach the nominal level of confidence $\left((1-\alpha)\times 100\%\right)$. On this basis, the PI coverage probability (PICP) is defined as [213-215]:

$$PICP = \frac{1}{n}\sum_{i=1}^{n} c_i \tag{43}$$

where $c_i\left(i=1,\cdots,n\right)$ is determined by:

$$c_i = \begin{cases} 0, & y_i \notin \left[\hat{y}_{Li}, \hat{y}_{Ui}\right] \\ 1, & y_i \in \left[\hat{y}_{Li}, \hat{y}_{Ui}\right] \end{cases} \tag{44}$$

The PICP describes how accurate the output intervals are. However, the PICP is not enough to quantify the quality of the PIs since a very wide PI can always enclose its target value. Such a wide PI provides little useful information. Based on this fact, another measure is required for quantifying the width of PIs. The mean prediction interval width (MPIW) is defined as follows [214, 215]:

$$MPIW = \frac{1}{n}\sum_{i=1}^{n}\left(\hat{y}_{Ui} - \hat{y}_{Li}\right) \tag{45}$$

The normalized MPIW (NMPIW) is given as below [214, 215]:

$$NMPIW = \frac{MPIW}{R} \tag{46}$$

where $R$ is the range of the underlying target.

From a practical standpoint, a PI capable of enclosing the target value and having a small interval width is preferred, which means the maximization of PICP and minimization of MPIW and NMPIW. However, these two goals are conflicting as reducing the width of PIs often results in a decrease in PICP and vice versa. To this end, a synthesized loss function called combinational coverage width-based criterion (CWC) is proposed for evaluation of the PIs [213]:

$$CWC = NMPIW\left(1 + \gamma\left(PICP\right)e^{-\eta(PICP-\mu)}\right) \tag{47}$$

where $\alpha$ and $\mu$ are hyperparameters, $\mu$ denotes the nominal confidence level associated with the PIs ($1-\alpha$), and $\gamma\left(PICP\right)$ is given by:

$$\gamma\left(PICP\right) = \begin{cases} 0, & PICP \geq \mu \\ 1, & PICP < \mu \end{cases} \tag{48}$$

CWC requires that the PICP is larger than the nominal confidence level, otherwise it will be heavily penalized with an exponential term. Once the confidence level is achieved, the width measure plays a major role in the loss.

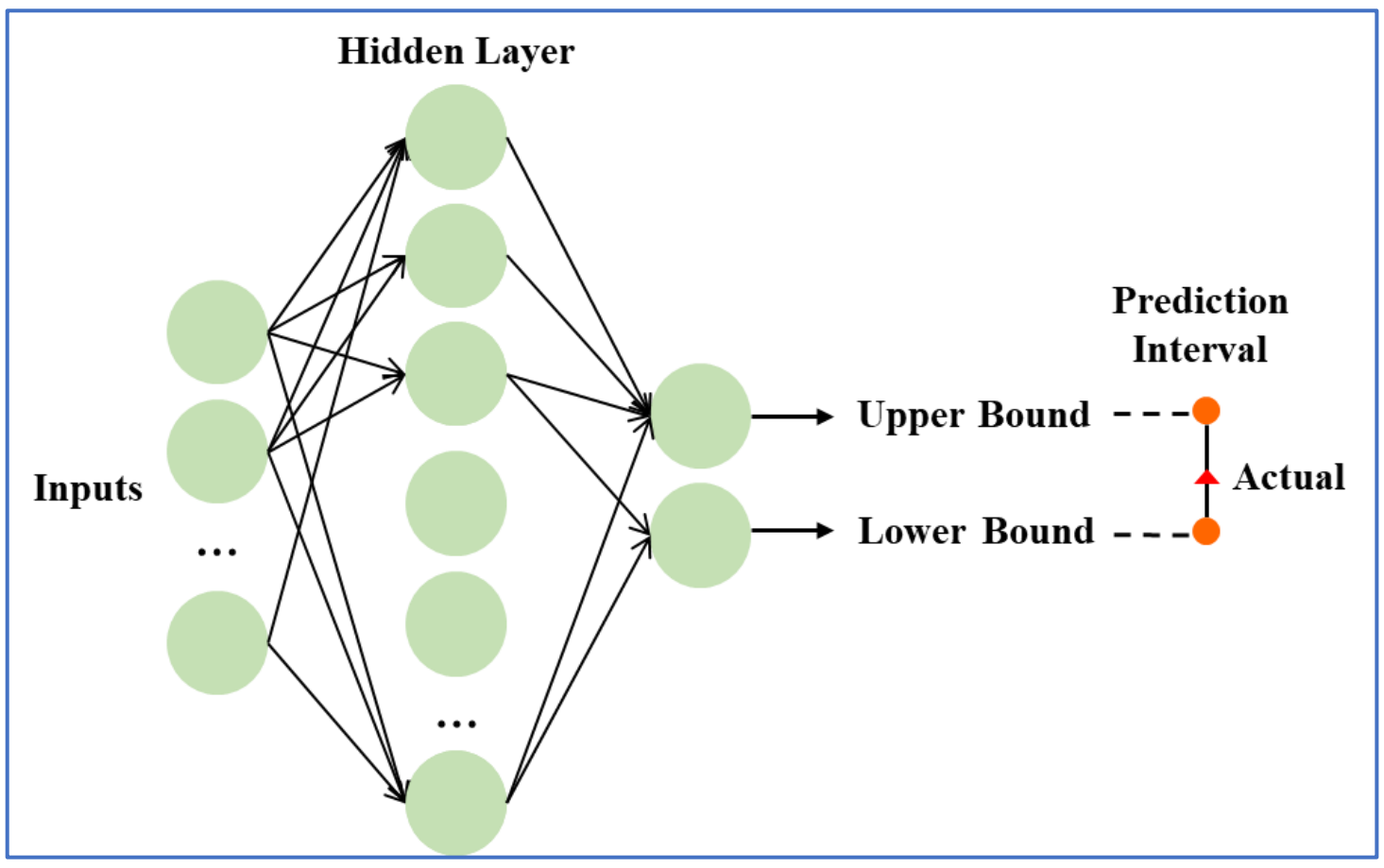

Figure 25. The structure of LUBE interval prediction using neural networks (reproduced from [213]).

# 6 Applications of Uncertainty-Aware Machine Learning to Structural Dynamics

In this section, some concrete examples concerning the application of uncertainty-aware ML models in structural dynamics are summarized and discussed, aiming to highlight current developments and trends while underscoring the importance of UT in this high-stakes, label-scarce field.

As illustrated in Fig. 1, structural dynamics includes forward and inverse problems according to the information provided and the objective of the analysis [2]. Forward problems aim to estimate the responses or characteristics of a given system with specified inputs, whereas inverse problems focus on inferring properties of the underlying system from measured or simulated dynamic responses. Each of these problem types comprises a variety of tasks that correspond to different ML paradigms. For example, structural damage identification, a representative inverse problem, may be formulated as a classification, regression, or anomaly detection task, each requiring distinct mechanisms and assumptions to formulate the predictive distribution and estimate uncertainty, as illustrated in Sections 2.3.1 and 2.3.2. Therefore, in order to provide a comprehensive and well-organized overview, this section adopts a task-specific structure. Specifically, for probabilistic methods, key tasks in structural dynamics are organized into three representative forward problems, including stochastic dynamic response prediction, sensitivity analysis, and reliability analysis, and two major inverse problems, including stochastic model updating and structural damage identification, while representative applications of uncertainty-aware PML models to these tasks are summarized accordingly. In contrast, applications of non-probabilistic methods are discussed individually due to their relatively limited use. The summary of these applications highlights each method's effectiveness in improving the performance of the employed ML model and its approach to interpreting and utilizing predictive uncertainty.

Additionally, among the ML models with UT applied in structural dynamics, regression models have received the most attention [26, 183]. This is primarily due to the prevalence of regression-based tasks such as model updating and reliability analysis, which align well with approaches combining domain knowledge in structural dynamics with ML techniques. Furthermore, predictive uncertainty in regression tasks can be intuitively quantified using the variance of the predictive distribution, which can also be decomposed into aleatoric and epistemic components using Eq. (2) to provide more insights about the reducible part of the predictive uncertainty. While several prior studies have comprehensively reviewed the use of PML models for UT in regression tasks within structural dynamics [26, 183], this survey expands the scope to include regression, classification, clustering, and anomaly detection tasks in structural dynamics, encompassing both probabilistic and non-probabilistic approaches. Additionally, this work emphasizes the application of Bayesian methods, especially BDL, owing to their distinct advantages in representing and estimating both aleatoric and epistemic uncertainties in DNN-based methods.

## *6.1 Applications of PML methods to structural dynamics*

### *6.1.1 Forward problems*

Forward problems typically involve regression tasks aiming to assess the response or functional condition of the investigated structural system [24], where PML models are often utilized as surrogate models to replace traditional high-fidelity finite element models (FEMs) for computationally efficient predictions without compromising the accuracy and robustness [24, 216]. In terms of UT, Bayesian approaches have become the current mainstream, as uncertainties arising from the choice and parameters of the ML model to describe the dynamic system are non-trivial, which are primarily epistemic uncertainty stemming from the lack of knowledge about the dynamic system and the relationship between the input and output. Based on these Bayesian PML methods, the posterior predictive distribution can be formulated to explicitly quantify the uncertainty in forward problems through its variance.

#### *6.1.1.1 Dynamic response prediction*

Predicting the response of a dynamical system with some given input would be the most intuitive application of ML techniques in the field of structural dynamics, which serves as the foundation for further operation and risk management of the monitored structure [24]. However, this task intrinsically involves considerable uncertainties arising from various sources such as the measurement noise, the random nature of stochastic vibration, the high nonlinearity of complex dynamical systems, the environmental variabilities, the inadequate choice of ML model parameters, etc. [133-135], which underscores the significance of UT in response prediction for more accurate and reliable results. In current literatures, BDL models and GPs are viewed as promising uncertainty-aware surrogate models due to their Bayesian nature and nonlinear mapping capability [108, 109, 133].

Conditioned on the uncertainty resulting from stochastic ground motion and structural nonlinearity, Wang et al. [136] proposed a Bayesian CNN trained by Bayes by Backprop for seismic response prediction, which, investigated using a numerical building structure, exhibits notable performance and robustness against input noise in predicting both acceleration and displacement responses. Based on the Gaussian assumption of model predictions, Kim et al. [135] proposed a new loss function to account for data uncertainty, while model uncertainty is captured via a Bayesian CNN with MC dropout. Concerned with the interpretability of the black-box model and prediction uncertainty in response prediction, Liu et al. [121] proposed a Physics-guided deep Markov model (PgDMM) by utilizing the physical property of a dynamical system to construct state space models, while transition and emission functions of the system is represented by a VAE to capture uncertainty. This method exhibits considerable generalization and prediction capabilities due to a more structured and physically interpretable uncertainty estimation framework. Combining the advantages of CNN in feature extraction and LSTM in modeling sequential data, Li et al. [133] proposed a BDL framework trained using Bayes by Backprop, which treats a set of selected network parameters as uncertain variables to estimate the uncertainty in vehicle-induced bridge response prediction. This method explicitly quantifies the confidence of the BDL to its predictions and exhibits robustness against various sources of randomness including variations in training dataset size, vehicle speed, and model input noise.

In summary, response prediction is a regression task with substantial uncertainties arising from both the characteristics of the dynamic system and the modeling process of the ML methods. Bayesian PML emerges as a common choice for addressing these uncertainties, which formulates the posterior predictive distribution $p\left(y \mid x, \mathcal{D}_{tr}\right)$ by integrating prespecified priors of model parameters with training data, with its expected value representing the most probable prediction and its variance explicitly quantifying the uncertainty in the response predictions.

#### *6.1.1.2 Sensitivity analysis*

Sensitivity analysis for dynamics systems aims to determine the contributions of individual system parameters to the investigated characteristics of the dynamic system, which is also a regression task and plays an essential role in the design and risk assessment of dynamical systems [112]. ML commonly serves as surrogate models to alleviate the computational burden resulting from repetitively running FEMs to acquire the investigated characteristics [111]. Since sensitivity analysis typically involves multiple system parameters with intricate correlations and coupled effects on the investigated characteristics, the choice of ML methods to model the input-output relationship in sensitivity analysis could entail significant epistemic uncertainties arising from both model architecture and model parameters, which makes GP a popular choice for UT in this task due to its nonparametric nature [111, 112]. In addition, there are also works dedicated to exploring non-Bayesian PML methods such as RFs and boosting techniques [61, 217] to address uncertainties associated with the inputted system parameters.

Conditioned on the flexibility of GPs, Wan et al. [112] proposed a variance-based global sensitivity analysis framework without assuming the form of input distributions, which offers a global and quantitative assessment of input importance in uncertainty of dynamic responses. Considering the uncertainty of factors that affect the fire resistance of RC slabs such as the fire scenario material strength, and convection conditions, Zhang et al. [61] used ensemble models including random forests, gradient boosting decision tree, and XGBoost for UT in sensitivity analysis of slab fire resistance, where the XGBoost-based method exhibits the best performance due to its superior UT capability. Xiong et al. [202] proposed an RBF-NN based on an improved mind evolutionary algorithm for global sensitivity analysis of spacecraft thermal design, which accounts for uncertainty through a predictive distribution characterized by the RBF cumulative distribution functions. This method exhibits superiority in both accuracy and efficiency compared with two traditional sensitivity analysis methods.

Despite the success of uncertainty-aware ML models in sensitivity analysis, most of current works rely on shallow ML models that may struggle to accurately capture the complex mappings from system parameters to investigated characteristics, especially in highly nonlinear systems. This underscores the potential of BDL for UT in sensitivity analysis of dynamic systems, which warrants further investigation.

#### *6.1.1.3 Reliability analysis*

Reliability analysis aims to find the probability of a dynamic system that performs its intended functions under specified operating conditions during a specified period, which is inherently a probabilistic problem and can be generalized as follows [13]:

$$p_f(t) = P\Big(G\big[x(t)\big] \le 0\Big) = \int_{G[x(t)]\le 0} f\big[x(t)\big]dx(t) \tag{49}$$

where $p_f(t)$ is the failure probability of the dynamic system at time $t$; $f\big[x(t)\big]$ is the joint PDF of the vector of basic uncertain variables, $x(t)$, and $G\big[x(t)\big]$ is some performance measure. Similar to other forward problems, PML models are also normally used as surrogate models for some performance measures in reliability analysis, such as limit state function, performance function, and reliability index [13]. In addition to the aleatoric uncertainties arising from the noise in the inputs, reliability analysis usually involves substantial epistemic uncertainties in the modeling process, including estimating the joint PDF, selecting the performance measure, and determining the ML method for surrogate modeling. As a result,

Bayesian methods are favored in PML-based reliability analysis owing to their advantages in addressing epistemic uncertainty, with prevalent methods including DPMM, BN, GP, RVM, and BDL.

To address the uncertainty in estimating the underlying PDF of the load- and resistance-related uncertain variables, Chen and Ni [94] proposed a DPMM-based method for bridge reliability analysis, which utilizes the nonparametric nature of DPMM to address the epistemic uncertainty associated with selecting the probabilistic model for the heterogeneous monitoring data used in reliability analysis. Considering both the efficiency and interpretability, Lu and Zhang [99] introduced a reliability analysis method for power distribution system. This method combines dynamic BN to capture uncertainty stemming from time-varying physical parameters of the power distribution system, and GP as a nonparametric surrogate model to account for uncertainty related to the choice of ML architecture in mapping inputted system parameters to the failure probability. Owing to its systematic UT framework, this method provides a dependable assessment of reliability for a dynamic system undergoing degradation, but it does not fully utilize the UQ capability of the Bayesian techniques to express the confidence in the predicted failure probability. Conditioned on the superior predictive performance of DL methods compared to traditional ML, there is a trend toward employing BDL in reliability analysis to utilize both its prediction and uncertainty estimation capabilities. Dang et al. [138] proposed a BDL-based surrogate model of the limit state function for reliability analysis, which yields a distribution of desired reliability indexes to account for both aleatoric and epistemic uncertainty, providing robust results and throughout understandings of structures' static and dynamic responses. He et al. [139] proposed a BDL framework that integrates a Bayesian MLP with a comprehensive evaluation metric for reliability analysis of railway ties, which explicitly quantifies the predictive uncertainty through prediction interval estimation. This method exhibits a better performance in deteriorate rate prediction than some traditional methods, owing to the combination of DNNs for nonlinear mapping with Bayesian inference for UT.

Compared to response prediction and sensitivity analysis, reliability analysis could entail greater uncertainty due to the complex time-variant characteristics of the dynamic system that could affect its present and future functional condition, which requires PML methods with both excellent predictive performance and comprehensive uncertainty estimation for reliable results, thereby increasing the focus on BDL methods in state-of-the-art research.

### *6.1.2 Inverse problems*

Inverse problems aim to refine a mathematical model of an existing structure using measured or simulated structural responses [2]. Broadly, there are two categories of approaches to addressing inverse problems that can employ PML models, namely the "model-driven" approaches and "data-driven" approaches [3]. In model-driven approaches, structural dynamic analysis relies on numerical models, typically finite element models, of the investigated structures, where PML models usually serve as surrogate models to learn the mapping from structural parameters to dynamic responses, with the goal of replacing computationally intensive iterative finite element analyses using efficient ML-based predictions. These methods are suitable when sufficient prior knowledge about the structures is available to build numerical models, and generally require fewer response measurements. In contrast, data-driven approaches directly use PML models as the mathematical model to map response measurements to the investigated system parameters or system condition. These methods do not need to construct and update numerical models, but typically require more extensive response measurements to train reliable ML models. Compared to forward problems, inverse problems involve a broader range of ML tasks, including anomaly detection, clustering, classification, and regression. However, the performance of ML tools in these inverse problems, especially for large-scale civil infrastructures, is highly dependent on the quality and quantity of empirical data available for training. In cases without sufficient and representative training data, it becomes challenging to construct reliable ML models. For instance, in structural damage detection, if adequate data from known healthy conditions are not provided, the baseline distribution may fail to capture the full diversity of normal structural behavior, potentially resulting in false negatives or false positives. A potential approach to alleviate this issue is employing finite element models to generate training data for different structural states, combined with hierarchical Bayesian modeling to estimate global model parameters and transfer their associated uncertainties from the baseline state to other states [218]. However, this approach is still limited by challenges in accurately constructing the finite element model and realistically representing real-world structural states.

These challenges underscore the importance of uncertainty estimation in ML-based inverse problems. Without proper UT, such problems become highly susceptible to ill-posedness and ill-conditioning due to limited information on inputs and incomplete knowledge of the structural system [219]. Additionally, predictive uncertainty estimates can play a proactive role in guiding the decision-making process. When a model's prediction is associated with high uncertainty, it indicates that the result is unreliable, and additional data should be collected to refine the model and reduce epistemic uncertainty, thereby leading to more robust and trustworthy predictions.

Uncertainty in inverse problems can arise from various sources, including inherent noise in response measurements, the architecture of the employed ML model, the effect of EOVs on the system, and parameters of the selected ML model. Consequently, Bayesian methods play a crucial role in UT for inverse problems, with the approach to formulate the predictive probability and quantify the associated uncertainty following the framework provided in Eqs.(1)-(3). On the other hand, non-Bayesian methods also hold a position in some fundamental tasks such as structural damage detection.

#### *6.1.2.1 Stochastic model updating*

Model updating refers to the task of establishing a reference model, which is typically a FEM, that reflects the current state of the investigated structure based on the measured responses, which is performed based on continuously updating the structural parameters, such as mass and stiffness, to match the model predictions and real response measurements [2]. ML techniques are normally used as surrogate models to learn the mapping from structural parameters to the dynamic responses, with the purpose of replacing time-consuming iterative FE analyses in performing the updating procedure [140]. Conditioned on uncertainties stemming from various sources such as material variability and measurement noise, a distribution or an interval of structural parameters is usually derived as the model updating result for better robustness and generalization, which facilitates the application of uncertainty-aware ML in model updating [220]. Zhang et al. [140] proposed a probabilistic model updating method based BDL with approximate Gaussian inference, which enables UQ of the estimated structural parameters efficiently and exhibits robustness against data scarcity. To quantify structural parameter uncertainty from responses variability, Yan et al. [221, 222] conducted a series of studies on model updating based on GP with TMCMC sampling, which demonstrate the robustness and computational efficiency of GPs in model updating due to its nonparametric nature and UQ capability. These works illustrate that special attention has been paid to UQ for model updating in current literature, and thus uncertainty-aware ML models have gained increasing prevalence with promising potential for this task. To improve the efficiency of UQ, Ref. [223] applied transfer learning realized by domain adaptation to bridge the gap between the biased numerical model and the real structure and to guide the Bayesian model updating process. Yu and Liu [224] introduces a novel physics-guided generative adversarial network (PG-GAN) for probabilistic structural parameter identification, incorporating physical awareness and a physics-based loss function to guide the training of generative models.. Considering the robustness of the learning process, Yin and Zhu [127] proposed a tailor-made algorithm for efficiently designing the appropriate architecture of Bayesian neural network and applied the method to model updating. Wang et al. [225] utilized Bayesian convolutional neural network to establish the relationship between the feature map of frequency response functions and model parameters, whereby realizing direct model updating without modal identification and modal matching. To address the optimization challenge and curse of dimensionality, Mo and Yan [226] proposed a novel solving framework called StocIPNet for stochastic inverse problems and applied it to high-dimensional stochastic model updating, where the distribution parameters of the physical random vector are embedded as learnable weights and biases and the inversion process is efficiently achieved using gradient-based optimization.

#### *6.1.2.2 Structural damage identification*

Structural damage identification entails the entire process of implementing a damage diagnosis strategy for aerospace, civil or mechanical engineering structures, which mainly involves dynamic response measurement, damage-sensitive feature extraction, and statistical analysis of these features to determine the current state of system health [3]. Generally, there are two major categories of damage identification methods, namely the model-driven methods and the data-driven methods. The former require a physics-based or law-based FEM of the monitored structure, while damage identification can subsequently be achieved by comparing the response measurements with the simulated data using the FEM [3]. To ensure that the FEM can accurately represent the investigated structure, a system identification or model updating process is usually involved in model-driven damage identification methods, wherein ML can be used as surrogate models [6] with similar UT approaches as illustrated above [53, 113]. On the other hand, data-driven methods directly establish a mapping from the structural features extracted from raw response measurements to the relevant diagnostic class labels or quantities through ML models. Additionally, as acquiring labels corresponding to damage scenarios for real-world engineering structures is often challenging or even inaccessible, there are also model-data-codriven methods (such as digital twin-based methods) for structural damage identification, which utilize the FEM to simulate well-annotated data for different damage scenarios to training the ML model [3]. To distinguish from the previous two sections, this section will focus on the applications of uncertainty-aware PML models in data-driven structural damage identification.

Generally, structural damage identification is the inverse problem that encompasses a wide range, which can be categorized into the following five levels according to Rytter's hierarchy [227], with each hierarchy requiring all lower-level information available [3], thereby involving higher level of uncertainty to be recognized and addressed.

- **Damage detection**: Damage detection is a binary classification task aiming to provide a qualitative indication of the presence of structural damage, which is the most fundamental but practical damage identification task as it can be achieved using unsupervised PML methods such as anomaly detection and clustering. The application of uncertainty-aware PML typically concentrates on aleatoric uncertainty arising from measurement noise, EOVs, manufacturing tolerance, etc., by employing methods such as PD-based anomaly detection [42-46, 74], EVT-based anomaly detection [48, 50-52], and GMM-based clustering [55, 56]. Additionally, there are also works addressing uncertainties in damage detection from a Bayesian perspective. For example, Rogers et al. [95] proposed a DPMM clustering-based method for damage detection under the effect of EOVs, which exhibits better performance than traditional MSD-based method due to correctly addressing the epistemic uncertainties stemming from the architecture and parameters of the statistical model used to represent the normal condition of the monitored structure.
- **Damage localization**: Damage localization aims to provide information about the probable position of damage, which is usually achieved by partitioning the monitored structure into prespecified subregions. This task can utilize both unsupervised and supervised PML models [49, 228]. Compared to damage detection, Bayesian PML are more prevalent in damage localization to address the higher level of epistemic uncertainty inherent in spatial information of damage. Eltouny and Liang [49] proposed an unsupervised damage localization method by integrating EVT and Bayesian optimization to address uncertainties from both data noise and parameters in the extreme joint PDF, which exhibits robustness to signal distortion and improved the hyperparameter tuning due to properly addressing uncertainties. Based on semantic damage segmentation (SDC), Liang and Sajedi [228] introduced a BDL framework that utilizes deep Bayesian U-Nets based on MC dropout to address uncertainties in damage localization. This method not only exhibits superior robustness with enhanced global and mean class accuracies compared to traditional SDC methods but also quantifies the predictive uncertainty through the softmax class variance of different predictions.

- **Damage classification**: damage classification aims to give information about the type of structural damage, which requires supervised PML methods with well-annotated training data. Consequently, beyond this hierarchy, PML applications are normally limited to laboratory or small-scale structures or dynamic systems due to the difficulty in recognizing and defining different damage scenarios for large-scale structures. Popular uncertainty-aware PML methods for damage classification include HMMs [79], Adaboost [69], etc., while BDL has also received attention recently due to its advantages in UT. Sajedi and Liang [142] proposed a Bayesian CNN based on MC dropout for damage detection, localization, and classification, where uncertainty can be quantified through variations in softmax probability and entropy of the dropout model predictions. Conditioned on the feature extraction and UT capabilities of VAEs, Martin et al. [124] proposed a bearing fault classification method using a VAE to automatically extract low-dimensional damage-sensitive features, which outperforms some other dimensionality reduction methods in both accuracy and robustness.

- **Damage assessment**: Assessment of damage refers to estimating the extent of damage, which also involves uncertainty arising from the choice of indices to quantify the damage extent, in addition to uncertainties in the above tasks. As a result, BDL is also employed in damage assessment for uncertainty representation and quantification. Utilizing the gap along the bearing surface boundary of miter gates as a damage quantification index, Hoskere et al. [145] proposed a Bayesian MLP based on MC dropout for damage assessment of miter gates, which results in robust predictions of the damage extent and provides an indication of the NN's confidence in the damage extent through the predictive variance. Liu et al. [229] proposed a distributed damage diagnosis method by utilizing a VAE for extracting statistical features from the long-gauge strain transmissibility and a regressor for mapping the features to elemental stiffness reduction.
- **Remaining useful life (RUL) prediction**: RUL prediction aims to estimate the safety of the structure, which entails additional uncertainty resulting from the time-varying characteristics of the monitored structure compared to damage assessment. Therefore, the application of BNNs in RUL prediction can be combined with NN architectures designed for sequential data. Caceres et al. [143] proposed a Bayesian RNN trained using Bayes by Backprop with the Flipout method for RUL prognostics, which outperforms some deterministic DL models as well as the MC dropout-based Bayesian RNN, especially when dealing with more complex scenarios involving multiple operating conditions and fault modes, due to considering both epistemic and aleatoric uncertainties.

In summary, data-driven structural damage identification comprises a hierarchy of five tasks, with PML directly used to map dynamic responses to the structural conditions. As the complexity of damage identification tasks increases, higher levels of uncertainty arise due to the sparsity of labeled training data and the requirement of all lower-level information. Consequently, uncertainty-aware PML with complex structures, such as BDL methods, are primarily utilized in high-level damage identification tasks such as damage classification, damage assessment, and RUL prediction.

### *6.2 Applications of non-probabilistic ML methods to structural dynamics*

Compared with PML-based methods, non-probabilistic ML methods have limited applications on UT and structural response prediction. The first category of applications is UQ. In non-probabilistic problems, finding out the most compact convex model enclosing the sample points is the core issue of UQ. Conventional convex models such as interval model, multidimensional ellipsoid model and multidimensional parallelepiped model cannot effectively quantify clustered samples, resulting in overestimation of uncertainties. Unsupervised methods are accordingly adopted for UQ in case of clustered samples [196, 197, 208], effectively reducing the epistemic uncertainty.

The second category of applications is interval response prediction. Computing the output intervals with given input intervals remains non-trivial for non-probabilistic problems due to interval expansion. In [210], the DINN was devised to approximate the relationship between input intervals and output intervals, and further applied to predict the concrete compressive strength with interval-valued inputs. LUBE is another effective interval prediction model as it avoids interval arithmetic and directly produces output intervals. Liu et al. [203] proposed a warning interval construction method under the framework of LUBE, successfully carrying out early warning for the abnormal state of engineering structures. In [201], FNN was combined with perturbation methods for its exceptional differentiation mechanism and utilized to predict the interval bounds of structural responses.

### *6.3 Systematic summary of the applications*

To provide a more intuitive and systematic overview of the applications of uncertainty-aware ML models in structural dynamics, Table 5 lists some representative works related to the applied scenarios of the reviewed ML methods with UT in this work. BNNs are predominantly used in regression tasks in structural dynamics, such as response prediction, reliability analysis, and damage assessment, to explicitly quantify the predictive uncertainty through the variance of the predictive distribution. However, their applications are still limited by the computational burden stemming from their complex model architectures with substantial parameters, highlighting the demand for further improvements in the efficiency and accuracy of approximation techniques for BNNs.

Table 5. Selection from the literature on applications of uncertainty-aware ML methods in the field of structural dynamics.

| Uncertainty-aware ML method | | Application in structural dynamics | Ref. |
|---|---|---|---|
| Non-Bayesian | PD | sensitivity analysis | [41] |
| | | damage identification | [42-46, 74] |
| | EVT | reliability analysis | [47] |
| | | damage identification | [48-52] |
| | Mixture models | damage identification | [53-56] |
| | HMM | damage identification | [57-59] |
| | RF | sensitivity analysis | [60, 61] |
| | | load identification | [62] |

| | | | | |
|---|---|---|---|---|
| | | | damage identification | [63, 64] |
| | | Boosting | sensitivity analysis | [61] |
| | | | reliability analysis | [65] |
| | | | damage identification | [66-69] |
| | | Ensemble of NNs | response prediction | [70] |
| | | | damage identification | [71-73] |
| Bayesian | Non-NN methods | DPMM | reliability analysis | [94] |
| | | | damage identification | [95] |
| | | BN | reliability analysis | [99-103] |
| | | | damage identification | [230] |
| | | RVM | reliability analysis | [105] |
| | | | load identification | [106] |
| | | | damage identification | [107] |
| | | GP (DGP) | response prediction | [108, 109] |
| | | | sensitivity analysis | [110-112] |
| | | | model updating | [231] |
| | | | damage identification | [114, 115] |
| | NN-based methods | Shallow BNN | model updating | [127] |
| | | | damage identification | [25, 128, 129] |
| | | BBL | response prediction | [131] |
| | | | system identification | [130] |
| | | BDL | response prediction | [121, 122, 132-137] |
| | | | reliability analysis | [123, 138, 139, 232] |
| | | | model updating | [140] |
| | | | damage identification | [124-126, 141-147, 228] |
| Non-probabilistic | | PCA | uncertainty modeling for uncertain variables | [195] |
| | | K-means | uncertainty modeling for uncertain variables | [196] |
| | | Fuzzy clustering | uncertainty modeling for uncertain variables | [197] |
| | | | response prediction | [198] |
| | | | damage identification | [199] |
| | | DINN | reliability analysis | [200] |
| | | FNN | response prediction | [201] |
| | | | sensitivity analysis | [202] |

| | | | |
|---|---|---|---|
| | LUBE | damage identification | [203] |

# 7 Open Challenges and Future Outlook

Despite the exhaustive efforts and impressive achievements analyzed in previous sections, estimating uncertainty for ML models in structural dynamics is still an active research topic with some open issues and research gaps remaining to be addressed in the future. This section outlines some challenges that remain related to the desirable properties that any uncertainty-aware ML model should bear in mind, along with some potential directions for future work.

## *7.1 Challenges and potential future works of PML*

For PML methods, current research gaps in their applications to structural dynamics mainly arise from challenges related to the reliability, efficiency, and interpretability of their uncertainty estimation, which involve the following aspects:

- **Uncertainty quality:** While incorporating uncertainty estimation aims to evaluate the reliability of PML model predictions, the quality or reliability of the predictive uncertainty itself is often overlooked and underexplored in PML-based methods for structural dynamics. Generally, evaluating the quality of predictive uncertainty is more challenging than estimating or quantifying it, as the ground truth of uncertainty estimates is typically unavailable [36]. Commonly used methods concerning uncertainty quality include uncertainty calibration and generalization [28, 36]. Uncertainty calibration focuses on the discrepancy between subjective uncertainty estimates and (empirical) long-run frequencies, which can be measured by proper scoring rules such as log predictive probability and the Brier score [36]. It is worth noting that calibration is a separate issue from accuracy, implying that a PML model's predictions may be accurate yet with miscalibrated uncertainty, and vice versa. On the other hand, uncertainty generalization concerns generalization of the predictive uncertainty to domain shift, which can be estimated through the predictive uncertainty on OOD samples, as PML model with robust uncertainty estimates should output high predictive uncertainty for inputs drawn from a dataset different from the training data. While UT for PML-based methods in structural dynamics has attracted increasing attention in recent research works, few of them have evaluated the quality of the estimated predictive uncertainty, indicating a research gap that requires further exploration. Table 6 provides a qualitative comparison of the estimated predictive uncertainty among some commonly used PML methods, aiming to guide the selection of proper uncertainty-aware PML methods for reliable prediction as well as uncertainty estimation in structural dynamic problems.

Table 6. Comparison of the uncertainty quality of different PML methods (reproduced from [26]).

| Compared PML methods with UT | GP | BNN | | | Ensemble of NN |
|---|---|---|---|---|---|
| | | MCMC | VI | MC dropout | |
| Quality of uncertainty calibration | High | High-medium | Medium | Medium-low | High |
| Performance on detecting OOD samples | High | Low | Low | Low | Medium |
| Computational efficiency (training) | Low | Low | Medium-low | High | High |

- **Uncertainty decomposition:** Although the current mainstream of UT for PML methods focuses on the predictive uncertainty, explicitly distinguishing the aleatoric and epistemic components of predictive uncertainty could be more significant in practical applications. On one hand, modeling aleatoric uncertainty reveals the irreducible part of predictive uncertainty without assorting to expensive Monte Carlo simulations. On the other hand, epistemic uncertainty is crucial in safety-critical problems as it indicates whether more training data is required for more reliable model predictions. For regression tasks, the variance decomposition approach described in Section 2.3.1 provides a well-established framework for separating aleatoric and epistemic components of predictive uncertainty, and has been widely applied in structural dynamics for UT [26, 183]. However, for classification and unsupervised learning tasks, uncertainty decomposition remains largely underexplored despite the presence of many relevant problems in structural dynamics, highlighting a significant research gap that warrants further investigation. A promising direction for classification tasks is to quantify epistemic uncertainty using mutual information (MI) between the predicted class label $y^*$ and the model parameters $\theta$ [37]. MI can be derived using the difference between the predictive uncertainty, measured by the entropy of the expected predictive distribution $\mathcal{H}\left[\mathbb{E}_{p(\theta|\mathcal{D})}\left[p\left(y^*|x^*,\mathcal{D}\right)\right]\right]$, and the expected aleatoric uncertainty, measured by the expected entropy over an ensemble of predictions conditioned on sampled parameters $\mathbb{E}_{p(\theta|\mathcal{D})}\left[\mathcal{H}\left[p\left(y^*|x^*,\mathcal{D}\right)\right]\right]$. This approach enables the separation of uncertainty as follows, where $\mathcal{I}\left(y^*,\theta|x^*,\mathcal{D}\right)$ denotes the MI:

$$\underbrace{\mathcal{I}\left(y^*,\theta|x^*,\mathcal{D}\right)}_{\text{epistemic uncertainty}}=\underbrace{\mathcal{H}\left[\mathbb{E}_{p(\theta|\mathcal{D})}\left[p\left(y^*|x^*,\mathcal{D}\right)\right]\right]}_{\text{Predictive uncertainty}}-\underbrace{\mathbb{E}_{p(\theta|\mathcal{D})}\left[\mathcal{H}\left[p\left(y^*|x^*,\mathcal{D}\right)\right]\right]}_{\text{aleatoric uncertainty}} \tag{50}$$

- **Scalability and computational efficiency of PML methods:** Real-world structural dynamic problems usually involve large amounts of high-dimensional heterogeneous data, posing challenges for practical UT methods that must possess scalability to high dimensions and efficiency with limited computational resources. Generally, Bayesian PML methods are regarded as more systematic candidates for UT, but they typically incur a higher computational burden with lower scalability due to introducing prior and posterior distributions of model parameters [17, 26]. In contrast, Non-Bayesian methods are more efficient and scalable but have limited capability in dealing with epistemic uncertainty. As a result, more efficient UT approaches, especially Bayesian approaches, still require further investigation for large-scale structural dynamic problems. A promising direction for improving the efficiency of Bayesian approaches that has been applied to structural dynamics is Hamiltonian neural network (HNN)-based Hamiltonian Monte Carlo (HMC) sampling methods [233, 234]. These methods employ HNNs to predict gradients of the Hamiltonian system, rather than manually computing them as in traditional HMC sampling, thereby reducing the computational cost in drawing samples for posterior estimation and uncertainty quantification.

- **Toward UT with PML in the context of small data:** The challenges related to the scarcity of well-annotated training data have been widely recognized and emphasized in PML-based methods for structural dynamics [3, 26, 235, 236], as it can lead to substantial epistemic uncertainty in model predictions thereby complicating subsequent decision-making. Potential solutions to these issues include model-data co-driven approaches that employ high-fidelity FEMs to generate labeled data to train ML models for dynamic analysis [218, 237], TL-based methods [148, 223, 238-240] that leverage knowledge transfer among different dynamic systems and ML models, and physics-informed ML (PIML) techniques [183, 236, 241] that embed physics-informed loss functions or prior knowledge into the training process to avoid solving the true label [26, 242]. While these methods can significantly alleviate the limitations caused by insufficient domain knowledge or training data, UT for such approaches is often more complex and introduces additional considerations. For example, in TL-based methods, the uncertainty associated with transferring knowledge from one domain or system to another must be carefully modeled and addressed. Similarly, in PIML-based methods, integrating physical knowledge into the ML model, whether through the loss function or model architecture, introduces additional uncertainty during the integration process, which is significant yet difficult to capture. These challenges highlight the need for more comprehensive and systematic UT frameworks tailored to these approaches, in order to enhance the reliability and applicability of PML models in structural dynamics.

- **Online estimation of uncertainty with PML:** Typically, structural dynamic problems involve a long-term process that can span the entire service life of the monitored structure. Therefore, it is desirable for PML-based methods to make predictions and estimate uncertainty in an online manner as time progresses [95]. However, as only one sample is observed by the PML model at each step, online learning is more vulnerable to uncertainty arising from noisy data and insufficient knowledge [243]. Moreover, online learning requires the model to estimate the predictive distribution with the acquisition of each new sample, resulting in higher computational costs compared to offline methods due to the repeated inference. To this end, there exists a research gap in developing reliable and efficient UT approaches suitable for online PML methods, warranting further investigation.
- **Feature extraction for structural dynamic analysis with PML:** A fundamental challenge in applying ML to structural dynamics lies in the nature of the dynamic response data used for training, which can be represented in the time, frequency, or time-frequency domain. Each representation requires distinct preprocessing techniques and ML models to effectively preserve and learn the essential physical characteristics relevant to the underlying structural behavior. For example, time-domain responses are inherently sequential and require ML models capable of capturing their temporal dependencies. In contrast, frequency-domain data are often complex-valued and must be appropriately processed, e.g., through bandwidth selection, to ensure the model captures the modal characteristics. Time-frequency representations, while valuable for analyzing nonstationary signals, introduce additional challenges such as resolution trade-offs and high dimensionality. These complexities highlight the need for well-designed ML model architectures and domain-specific feature engineering strategies that can handle structured, sequential data while maintaining interpretability. A promising direction for addressing these challenges is the development of uncertainty-assisted model selection frameworks, such as cross-validation and Bayesian model selection guided by predictive uncertainty, to facilitate the establishment of appropriate and reliable ML models for different data types. Additionally, multi-domain data fusion techniques, such as time-frequency feature fusion networks [244], deep convolutional neural network-based fusion [245], and hybrid convolutional-recurrent architectures [246], also provide a powerful way to integrate complementary information from different feature domains, with the potential to enhance the robustness and generalizability of ML models in structural dynamics.

- **Beyond uncertainty from data noise and model parameters:** Although this work provides a comprehensive discussion of UT for ML models, it primarily focuses on uncertainty stemming from data noise (aleatoric uncertainty) and model parameters (epistemic uncertainty), as these are most commonly considered and addressed in structural dynamics applications. However, in practice, a range of additional factors can significantly contribute to the overall uncertainty in ML workflows and warrant careful consideration:

**Uncertainty of model architecture**: For a given training set, variations in model architecture, such as different numbers of layers, widths, and activation functions in NNs, can lead to different performance and uncertainty estimates. This introduces a form of epistemic uncertainty that is difficult to represent and estimate, as identifying the most appropriate architecture to explain the observed data is often infeasible in engineering applications, particularly in problems of structural dynamics involving complex systems with limited physical information. Current approaches to capture this uncertainty include ensemble models [247], cross validation [23], and Bayesian model selection [248]. However, these methods often rely on additional assumptions or explore only a limited set of architectures. Therefore, this remains an open area requiring further investigation.

**Uncertainty of prior and noise specification**: Apart from uncertainty in model parameters, choosing prior distributions to capture and represent these parameters, such as network weights, bias, and the noise variable in regression, is itself an uncertain modeling decision involving epistemic uncertainty. Promising future directions to address this uncertainty include hierarchical Bayesian modeling [249], which introduces hyperpriors to capture uncertainty in the selection of prior distributions, and physics-informed priors [250], which incorporate domain knowledge to facilitate prior specification.

**Uncertainty in the training process**: In real-world structural dynamics problems involving complex ML models, non-convex objective functions are common, making it challenging to reach the global optimum during training. In this scenario, factors related to the training process, including the choice of optimization algorithm, data splitting strategy, and regularization techniques, can introduce uncertainty and affect models' performance. To mitigate this uncertainty, some improved training strategies can be employed. For example, well-informed initialization, such as using TL techniques to provide a starting point for training in a new domain based on pretrained models, can guide optimization toward better solutions [251]. Additionally, advanced optimization algorithms such as Adam [252] and RMSprop [253] often provide better convergence in non-convex optimization with high-dimensional parameter spaces than traditional methods [254]. Reformulating training objectives with physics-informed loss or regularization terms as minimax problems can also help balance competing objectives and enhance convergence stability [255].

**Variance-bias tradeoff**: In ML, the variance-bias tradeoff describes the balance between a model's complexity, the accuracy of its predictions, and its generalizability to unseen data beyond the training set. As model complexity increases, the model becomes more flexible and better fits the training data, which means that it is effective to mitigate bias and avoid underfitting. However, overparameterized models are more sensitive to small variations in the training data, leading to increased variance and a greater risk of overfitting. Therefore, managing the tradeoff between variance and bias reflects a form of uncertainty that directly influences model performance. Some pioneering studies have explored solutions using physics-informed constraints, transfer learning, model pruning, and dictionary learning [256, 257]. Nevertheless, the variance-bias tradeoff in ML models for structural dynamics remains an open challenge that warrants further investigation, particularly in structural dynamics.

These forms of uncertainty underscore the need for more principled and systematic UT frameworks that extend beyond data noise and model parameter uncertainty, in order to further enhance the reliability and interpretability of ML applications in structural dynamics. While there has been notable progress in capturing some of these uncertainties, these efforts remain relatively limited compared to the extensive research devoted to addressing aleatoric uncertainty from data and epistemic uncertainty from model parameters. This gap highlights an important direction for future exploration.

### *7.2 Challenges and potential future works of non-probabilistic ML*

For non-probabilistic methods, their applications in structural dynamics are mainly hindered by the issues associated with the accuracy and efficiency, which involve the following two aspects:

- **Uncertainty quality:** The accuracy of uncertainty-aware ML methods, particularly within safety-critical domains, remains largely unexplored in a systematic manner. Among present methods, the DINN stands as an intrusive approach that integrates interval arithmetic into NNs, while the FNN-based method operates as a non-intrusive technique utilizing NNs as surrogate models. Despite offering novel perspectives for addressing uncertainty propagation and prediction challenges, neither method demonstrates superior accuracy compared to conventional approaches. The inherent issue of interval expansion within interval arithmetic significantly impacts the DINN, resulting in substantial overestimation and consequently low-quality uncertainty assessments. Similarly, the FNN-based method, relying on conventional Taylor expansion with low-order terms, may incur significant prediction errors, particularly in nonlinear and high-dimensional scenarios. Despite some applications, research on ML-based non-probabilistic uncertainty propagation remains limited. Moving forward, the development of additional ML methodologies and enhancement of accuracy stand as primary avenues for future investigation and advancement in this domain.

- **Scalability and computational efficiency:** The inaccuracy of the NN-based non-probabilistic methods directly constrain the scope of problem dimensions that can be effectively addressed. Present methods are predominantly suited for low-dimensional and small-scale problems, primarily due to the challenges stemming from interval expansion and truncation error within Taylor expansions as parameter dimensionality escalates. Furthermore, the DINN exhibits susceptibility to instability during the training phase and necessitates greater computational resources in comparison to deterministic models of similar scale. Consequently, it is imperative that future research prioritizes the exploration of scalability and computational efficiency as critical areas.

## 8 Conclusions

This paper provides a systematic overview that encompasses the applications of uncertainty-aware ML methods in the field of structural dynamics, with a particular emphasis on state-of-the-art UT methods for NNs. Specifically, this work addresses four salient aspects: (1) types of uncertainty (aleatoric and epistemic), sources of each type of uncertainty in ML, and potential causes of each type of uncertainty in structural dynamics; (2) types of UT methods for ML models (probabilistic and non-probabilistic methods); (3) applications of uncertainty-aware ML methods in forward and inverse problems of structural dynamics; and (4) current research gaps and future directions in this field.

Without delving into detailed introductions of each ML model, this work aims to provide a guidance to readers about the differences in representing and estimating uncertainty for each method by elaborating the formulation of their predictive uncertainties. Among the UT methods discussed, Bayesian approaches, especially BNNs, stand out as promising candidates for addressing uncertainties in complex real-world problems due to their distinctive perspective to capture epistemic uncertainties stemming from model parameters. Additionally, this work offers a brief review of the applications of the discussed uncertainty-aware ML methods in forward and inverse problems of structural dynamics, with a focus on the predominant types and causes of uncertainty, and the attainment of UT in each specific tasks. Significant emphasis is placed on the applications of BDL, as it leverages the superior learning capability of DNNs in large-scale complex problems, combined with Bayesian approaches to address uncertainties in DNNs' predictions, aiming to improve their reliability. Having these ML methods with UT and their successful applications in mind, this work also delineates some contemporary research gaps, primarily concerning issues related to the efficiency, scalability, and quality of estimated uncertainty in current methods. These gaps aim to provide researchers with deeper insights into the future directions of this field.

In essence, UT serves as a critical layer of safety assurance on top of ML models, facilitating rigorous and quantitative risk analysis and management of ML applications in high-stake scenarios such as structural dynamics. With the evolution of UT methods for ML models, they are expected to play a pivotal role in ensuring the safety, reliability, and trustworthiness of ML solutions for real-world engineering problems, enabling to risk minimization in decision-making processes. Consequently, the development of uncertainty-aware ML methods holds paramount significance in broadening the applicability of ML models both extensively and deeply. The accurate, systematic, and reliable quantification of predictive uncertainty holds tremendous potential to fundamentally tackle the safety assurance challenge that haunts ML development. With this in mind, this study systematically reviews the background, current advancements, open challenges, and future directions of UT for ML-based methods in structural dynamics, aiming to provide the research community with more comprehensive insights into this domain.

**Acknowledgements**

This research has been supported by the Science and Technology Development Fund, Macau SAR (File no.: 101/2021/A2, 0010/2021/AGJ, 001/SKL/2024), the Research Committee of University of Macau (File no.: MYRG2022-00096-IOTSC), Guangdong-Hong Kong-Macau Joint Laboratory Program (Project No.: 2020B1212030009).

**Declaration of generative AI and AI-assisted technologies in the writing process**

During the preparation of this work the authors used ChatGPT in order to improve the language. After using this tool, the authors reviewed and edited the content as needed and take full responsibility for the content of the publication.